\documentclass{article}

\usepackage[utf8]{inputenc} % allow utf-8 input
\usepackage[T1]{fontenc}    % use 8-bit T1 fonts
\usepackage{libertine}
\usepackage[libertine]{newtxmath}
\usepackage{inconsolata}
\usepackage{hyperref}       % hyperlinks
\usepackage{url}            % simple URL typesetting
\usepackage{booktabs}       % professional-quality tables
\usepackage{nicefrac}       % compact symbols for 1/2, etc.
\usepackage{microtype}  
\usepackage{bbm}
\usepackage{float}
\usepackage[symbol]{footmisc}
\usepackage{threeparttable}

\usepackage{makecell}

\newcommand{\showComments}{no}

\usepackage[accepted]{sysml2019}

\sysmltitlerunning{Accelerating Deep Learning by Focusing on the Biggest Losers}

\author{%
  David S.~Hippocampus\thanks{Use footnote for providing further information
    about author (webpage, alternative address)---\emph{not} for acknowledging
    funding agencies.} \\
  Department of Computer Science\\
  Cranberry-Lemon University\\
  Pittsburgh, PA 15213 \\
  \texttt{hippo@cs.cranberry-lemon.edu} \\
}

\usepackage{algorithm}
\usepackage{algorithmic}
\usepackage{subcaption}
\usepackage{siunitx,booktabs}
\usepackage{xcolor}
\usepackage{microtype}
\usepackage{graphicx}
\usepackage{booktabs} % for professional tables
\usepackage{enumitem}

\usepackage{comment}

\newcommand{\note}[2]{\ifthenelse{\equal{\showComments}{yes}}{\textcolor{#1}{#2}}{}}

\definecolor{pink}{rgb}{0.858, 0.188, 0.478}

\newcommand{\mk}[1]{\note{red}{MK: #1}}

\newcommand{\angela}[1]{\note{green}{Angela: #1}}
\newcommand{\greg}[1]{\note{cyan}{GG: #1}}
\newcommand{\babu}[1]{\note{pink}{Babu: #1}}
\newcommand{\gauri}[1]{\note{magenta}{Gauri: #1}}
\newcommand{\giulio}[1]{\note{magenta}{Giulio: #1}}
\newcommand{\daniel}[1]{\note{orange}{Dan: #1}}
\newcommand{\zack}[1]{\note{orange}{ZL: #1}}

\newcommand{\System}{Selective-Backprop}
\newcommand{\Kath}{Kath18}

\newcommand{\SB}{SB}
\newcommand{\Stale}{Stale-SB}

\newcommand{\Scan}{Traditional} %Uniform/NoFilter

\newcommand{\sourceURL}{\url{https://bit.ly/SelectiveBackpropAnon}}

\newlength\myindent
\newcommand\bindent{%
  \begingroup
  \setlength{\itemindent}{\myindent}
  \addtolength{\algorithmicindent}{\myindent}
}
\newcommand\eindent{\endgroup}

\newcommand{\MaxSpeedup}{3.5x}
\newcommand{\MaxStaleSpeedup}{5x}
\newcommand{\StaleOverSB}{26\%}
\newcommand{\StaleOverKath}{1.3--2.3x}
\newcommand{\SBOverKath}{1.02--1.8x}

\newcommand{\CurrentAsymmetry}{2x}
\newcommand{\FutureAsymmetry}{10x}

\newcommand{\squeeze}{\vspace{-2mm}}

\begin{document}
%%% FROM SYSML TEMPLATE %%%
\twocolumn[
\sysmltitle{Accelerating Deep Learning by Focusing on the Biggest Losers}

\sysmlsetsymbol{equal}{*}

\begin{sysmlauthorlist}
\sysmlauthor{Angela H. Jiang}{cmu}
\sysmlauthor{Daniel L.-K. Wong}{cmu}
\sysmlauthor{Giulio Zhou}{cmu}
\sysmlauthor{David G. Andersen}{cmu,goo}
\sysmlauthor{Jeffrey Dean}{goo}
\sysmlauthor{Gregory R. Ganger}{cmu}
\sysmlauthor{Gauri Joshi}{cmu}
\sysmlauthor{Michael Kaminksy}{brdg,cmu}
\sysmlauthor{Michael Kozuch}{intel}
\sysmlauthor{Zachary C. Lipton}{cmu}
\sysmlauthor{Padmanabhan Pillai}{intel}
\end{sysmlauthorlist}

\sysmlaffiliation{cmu}{Carnegie Mellon University}
\sysmlaffiliation{goo}{Google Brain}
\sysmlaffiliation{intel}{Intel Labs}
\sysmlaffiliation{brdg}{BrdgAI}

\sysmlcorrespondingauthor{Angela H. Jiang}{ahjiang@cs.cmu.edu}

\sysmlkeywords{Machine Learning, SysML}

\vskip 0.3in

%%% END FROM SYSML TEMPLATE %%%

%\printAffiliationsAndNotice{\sysmlEqualContribution} % otherwise use the standard text.

]
\printAffiliationsAndNotice{} 

\begin{abstract}

This paper introduces \System{}, a technique that accelerates the training of
deep neural networks (DNNs) by prioritizing examples with high loss at each iteration.
\System{} uses the output of a training example's forward pass 
to decide whether to use that example to compute gradients and update parameters, 
or to skip immediately to the next example. By reducing the number of computationally-expensive
backpropagation steps performed, \System{} accelerates training.  Evaluation on CIFAR10, CIFAR100, and SVHN,
across a variety of modern image models, shows that \System{} converges to target error rates up to
\MaxSpeedup{} faster than with standard SGD and between \SBOverKath{} faster 
% \zack{What is "a state-of-the-art sampling approach". If you are comparing to an existing method, state it explicitly.}
% \zack{I would say the name of the method here and save the citation for the body. Citations in the abstract are a bit stylistically weird. --- the abstract will appear in places where the bibliography does not.}
% \angela{The sampling approach was never given a name. I agree that a ref in the abstract is not ideal}
than a state-of-the-art importance sampling approach. Further acceleration of \StaleOverSB{} can
be achieved by using stale forward pass results for selection, thus also skipping forward
passes of low priority examples. The implementation of \System{} is open-source and
can be found at \sourceURL{}.

\begin{comment}
Currently, DNN training time is dominated by calculating and updating of small gradients 
\zack{Too vague. What does "relatively low contribution to the trained network" mean? The network is determed *only* by the gradients, even if they are all small!}
that have relatively low contribution to the trained network.
\System{} uses the output of a candidate training example's
forward pass 
to decide whether to use that example 
to compute gradients and update parameters, 
or to skip immediately to the next candidate.  
\zack{I don't think you need to say "automatically. Everything that an algorithm
does is automatic just say "adapts...."}
\System{} adapts selectivity to the state of the network, as training progresses, 
by deciding whether to backpropagate an example 
with a probability proportional to its loss.  
Evaluation on MNIST, CIFAR10, CIFAR100 and SVHN, 
across a variety of modern image models, 
shows that this mechanism converges to target error rates
\SpeedupRange{} faster than with standard SGD and \greg{range-2} faster 
\zack{What is "a state-of-the-art sampling approach". If you are comparing to an existing method, state it explicitly.}
than a state-of-the-art sampling approach.  
The code to \System{} is open-source and can be found at \sourceURL{}.
\end{comment}

\end{abstract}

% !TeX root = ./selective_backprop.tex

\section{Introduction}
\label{sec:intro}

%\zack{Might be too informal to use "backpropped" as a verb}
% \zack{Be careful about speaking too loosely about ``information'', also a term with formal technical meaning but used loosely here. What precisely does it mean to "impart information to the network"?}
% \zack{I would not have a figure floating before the introduction starts. The CV clowns do this sort of thing but NeurIPS reviewers would find it tacky. Top of page 2 is a great place for a solid figure 1.}\mk{Agreed}

% Training deep neural networks (DNNs) consumes a great deal of time and resources---the 
% majority of which is spent on calculating and gradients from small losses.
% Move FIG 1 from design.tex to ensure it appears on page 2

% \zack{This sentence is too loose semantically and a little loose grammatically.
% Let me know if you'd like me to take a crack at reworking.
% E.g. ``contain redundancy'' is not crisp. also there's a lack of agreement in *samples provide little marginal improvement in the model*. I know what you mean but it will trip out a reviewer's language model and perhaps expose the paper as non-native to ML.}

% Large datasets, such as those used to train DNNs, contain redundancy, 
% and thus certain training samples provide little or no marginal improvement in the model if a similar 
% example has already been considered~\cite{hinton07}.
While training neural networks (e.g., for classification), computational effort
is typically apportioned equally among training examples,
regardless of whether the examples are already scored with low loss or if
they are mis-predicted by the current state of the network~\cite{hinton07}.
In practice, however, not all examples are equally useful. As training progresses, the network begins to classify some examples accurately,
especially redundant examples that are well-represented in the dataset.
%score some examples well after several epochs of training.  (Also, large datasets
%often have redundancy in the form of similar examples.)
Training using such samples may provide little to no benefit; 
hence, limited computational resources
may be better spent training on examples 
that the network has not yet learned to predict correctly. 

\begin{figure}
    \centering
    \includegraphics[width=.9\linewidth]{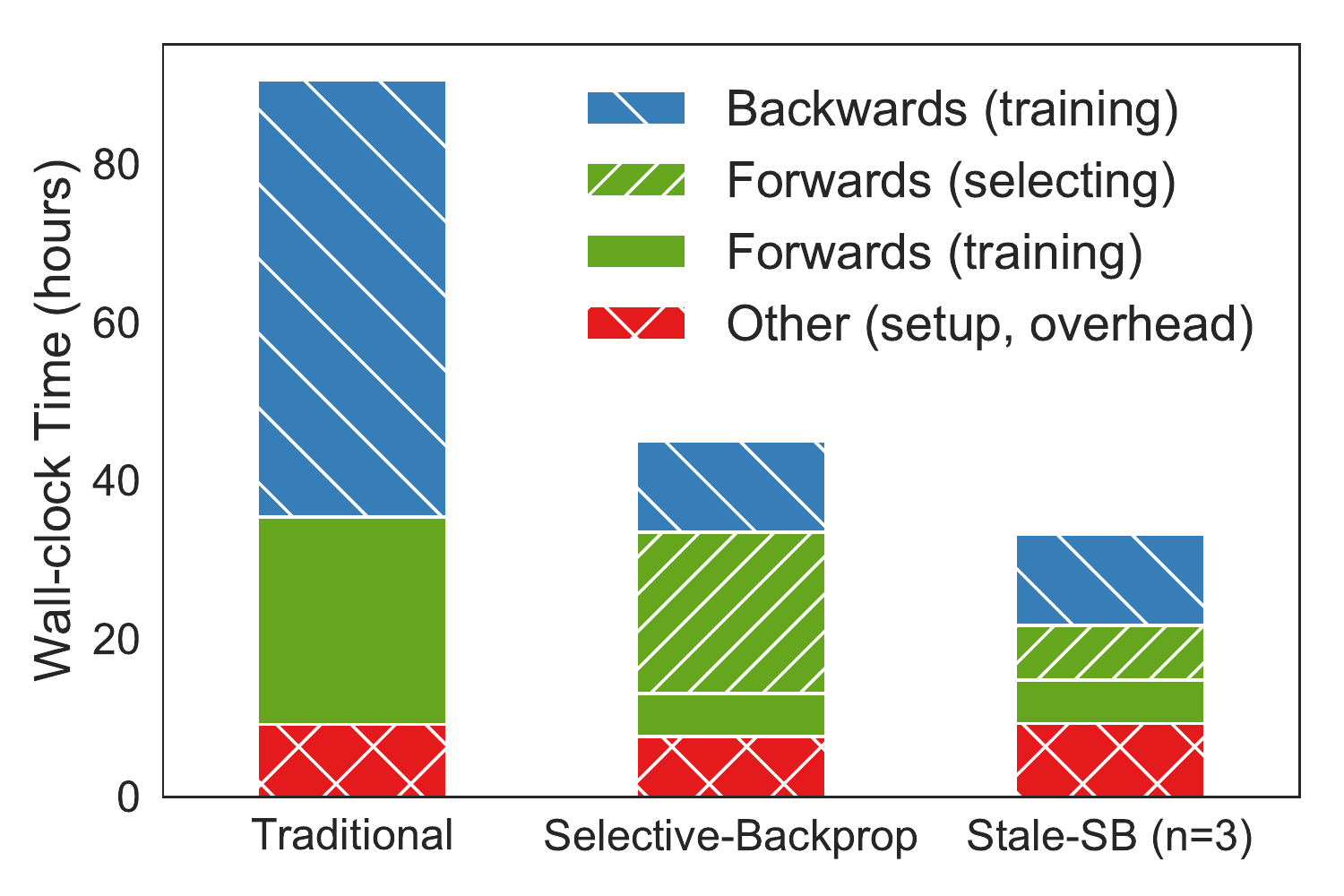}
    \caption{Comparison and breakdown of training time by \Scan{} training and proposed
    \System{} approaches, for training Wide-Resnet on SHVN until 1.72\% error rate is achieved 
    (1.2 times the final error of \Scan{}).  \SB{} accelerates training by reducing the number of 
    computationally expensive backward passes.  \Stale{} further accelerates training by sometimes 
    reusing losses calculated in the prior epoch for example selection.}
    \label{fig:front-page}
\end{figure}

\begin{figure*}
\begin{minipage}{\textwidth}
  \hspace*{\fill}%
        \begin{subfigure}[b]{0.45\textwidth}
          \centering
          \includegraphics[width=.7\linewidth]{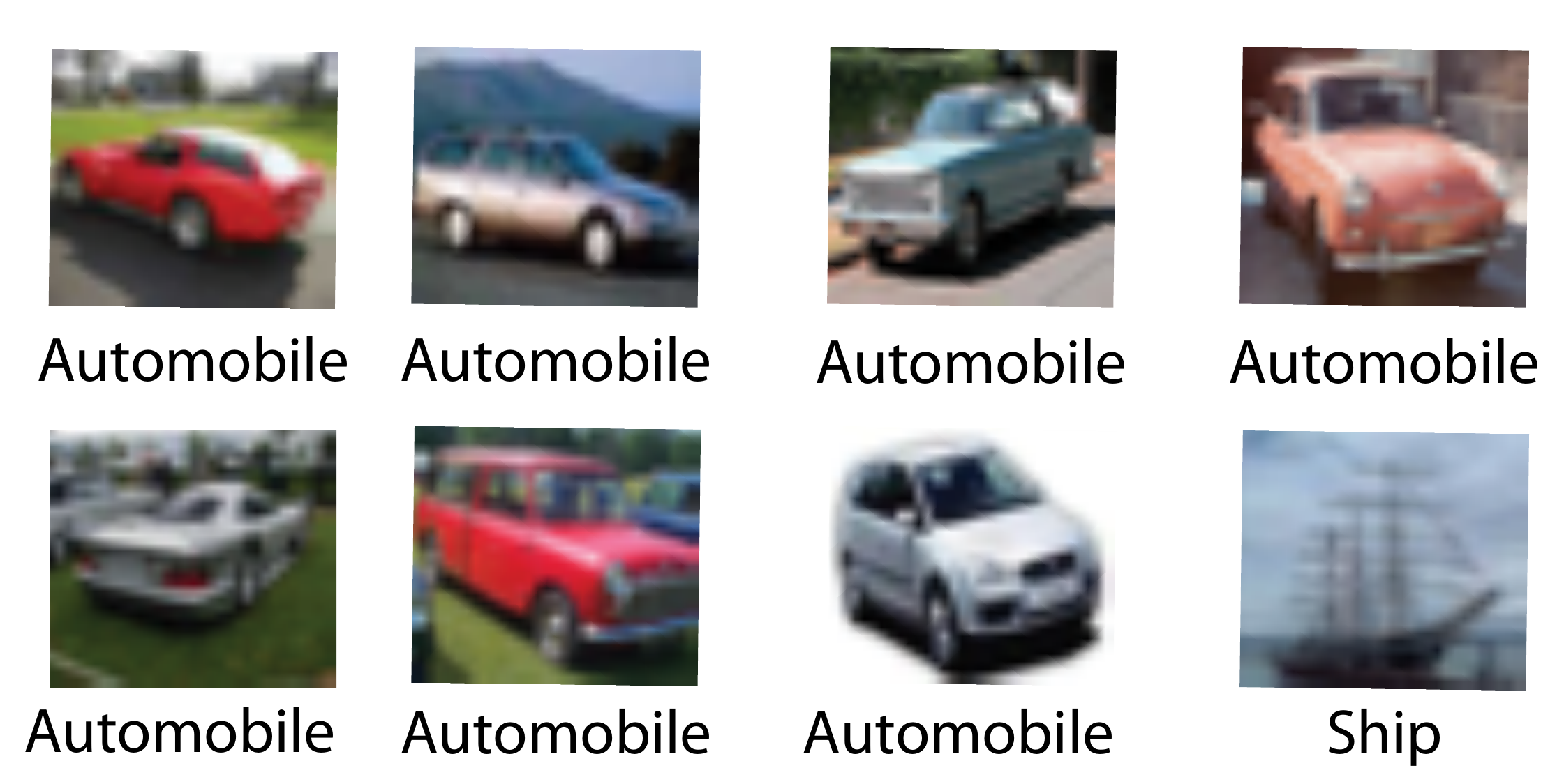}
          \caption{Examples chosen least frequently by \SB{}}
          \label{fig:easy}
        \end{subfigure}%
        \hfill%
        \begin{subfigure}[b]{0.45\textwidth}
          \centering
          \includegraphics[width=.7\linewidth]{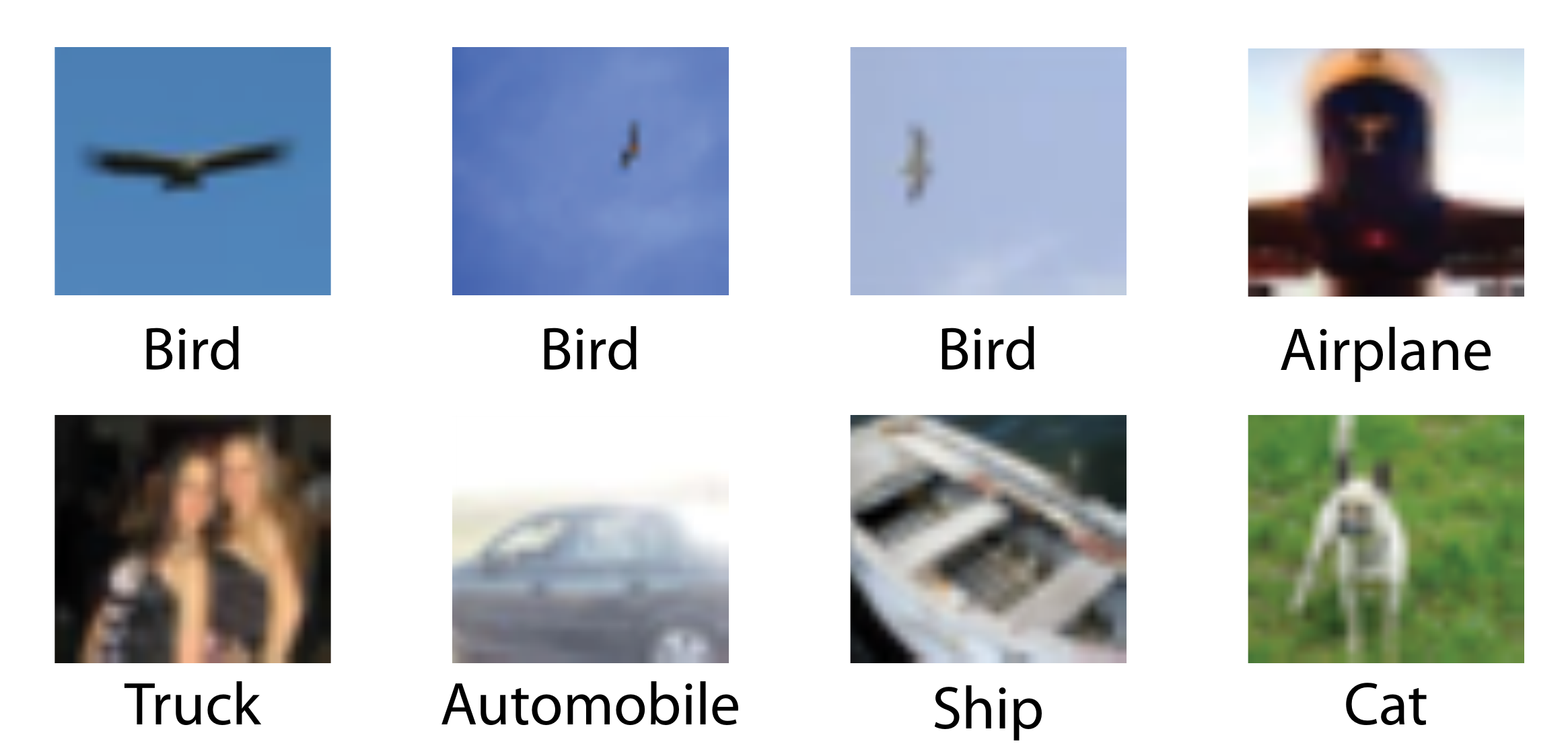}
          \caption{Examples chosen most frequently by \SB{}}
          \label{fig:hard}
        \end{subfigure}%
  \hspace*{\fill}%
\end{minipage}
\caption{Example images from CIFAR10}
\label{fig:example-images}
\end{figure*}

Figure~\ref{fig:example-images} illustrates the 
%Supporting our intuition that there is significant 
redundancy difference between ``easy'' examples and ``hard'' examples.
Figure~\ref{fig:easy} shows examples from CIFAR10 that consistently
produce low losses over the course of training. 
Compared with examples that generate high losses (Figure~\ref{fig:hard}),  the classes of low-loss examples are easily distinguishable.

Motivated by the hinge loss~\cite{rosasco04}, which provides zero loss %for examples 
% ZL: the examples don't exceed a threshold, the predictions do
% which exceed
% also the threshold isn't simply a "confidence" score.
% *confidently wrong* predictions still get loss
% a confidence threshold, 
whenever an example is correctly predicted by sufficient margin,
this paper introduces \emph{\System{}} (\SB{}), 
a simple and effective sampling technique 
for prioritizing high-loss training examples
% ZL: we're not really doing online learning here. 
% in an online fashion.
throughout training.
% 
%we explore the effects of prioritizing training on examples which are challenging for the network to classify. 
% 
%which may have relatively low marginal impact on the final model~\cite{hinton07}.
% 
% For example, when training on an image classification task, 
% we find that 60\% of training is spent on examples with cross-entropy losses 
% that are are less than 0.01\% of the maximum loss seen. 
% \zack{This claim needs to be made more specific. It's possible for no individual gradient update to have a big impact and yet for them together to fully account for the learned model. I think you have a more specific motivation in mind but it needs to be made explicit. Seems like the huice of your argument }
We suspect, and confirm experimentally, that examples with low loss correspond 
to gradients with small norm and thus contribute little to the gradient update.
% Motivated by this intuition, 
Thus, \System{} uses the loss calculated during the forward pass
as a computationally cheap proxy for the gradient norm,
enabling us to decide whether to apply an update
without having to actually compute the gradient.
\System{} prioritizes gradient updates
for examples for which a forward pass reveals high loss,
probabilistically skipping the backward pass for examples exhibiting low loss.

% \begin{figure*}[h!]
% %\squeeze{}
% % \hspace*{\fill}%
% % \begin{minipage}[t]{0.45\linewidth}
% \centering
%     \includegraphics[width=.35\linewidth, clip]{sysml2019/figs/sysml20/speedup_bar_svhn.png}
%     % \vspace{-2em}
%     \caption{\small{Speedup for SVHN with target error of }}
%     \label{fig:speedup}
% % \end{minipage}%
%     % \hfill%
% % \hspace*{\fill}%
% %\squeeze{}
% \end{figure*}

By reducing computation spent on low-loss examples, 
\System{} reaches a given target accuracy significantly faster. 
%without sacrificing final accuracy. 
Figure~\ref{fig:front-page} shows this effect for one experiment in our evaluation.
As seen in the first stacked bar (``Traditional''),
in which every example is fully trained on in each epoch, backpropagation generally consumes approximately twice 
% $\approx$\CurrentAsymmetry{} 
the time of forward propagation~\cite{convbenchmarks}.
In this experiment, \System{} (second stacked bar) reduces the number of backpropagations by $\approx$70\% and thereby cuts the overall training time in half. Across a range of models and datasets, our measurements show that \System{} speeds training to target errors by up to \MaxSpeedup{}.

Given \System{}'s reduction of backpropagations, over half of the remaining training time is spent on forward passes, most of which correspond to non-selected examples.
These forward passes do play an important role, though, because
the loss order of examples varies throughout training and varies more with \System{}.
A model might generate relatively low loss on a given example after some training,
but progressively higher losses on the same example if it is ignored for several epochs~\cite{hinton07}.
%, its loss may stagnate or increase.
\System{} 
% always 
evaluates sampling probabilities 
on the basis of an up-to-date forward pass,
% which ensures that 
ensuring
% our belief about the network's likely performance 
its assessment of the network's performance on the example is out of date.

\begin{comment}
Although \System{} does not reduce the number of forward passes performed,
we show that reducing the number of backwards passes 
can significantly reduce wall-clock time.
\end{comment}

The number of forward passes can be reduced, however, by allowing for some
staleness during selection.  One simple approach, for which we call the
corresponding \SB{} variant \Stale{} (third stacked bar), is to perform forward
passes to inform selection only every $n\textsuperscript{th}$ epoch.  In
intervening epochs, \Stale{} uses results of the most recent forward pass of an
example for selection, though it needs an up-to-date forward pass for the
training of selected examples.  For $n$=3, Figure~\ref{fig:front-page} shows
that \Stale{} avoids approximately half of all forward passes, reducing
training time by \StaleOverSB{} relative to \System{} with minimal loss in
final accuracy.  Although our experiments show \Stale{} captures most of the
potential reduction, we also discuss other approaches to reducing
selection-associated forward pass time.

\System{} requires minimal modifications to existing training protocols,
% \zack{this claim is too vague. performance in what sense? does it work empirically as well in terms of accuracy regardless of dataset size? we don't know that. we haven't tried it with a billion examples and I don't see a theorem that proves it will get same accuracy benefits with 1BN examples. Is this claim about computation? Make it crystal clear.}
% 
% ZL    redundant with scalable to arbitrarily large .... later in this paragraph
% is scalable to large datasets sizes, and applicable to 
% a range of 
applies broadly to DNN training, and 
works in tandem with data augmentation, cutout, dropout, and batch normalization.
Our experiments show that, 
% by adding a single filtering step to the training cycle, 
% and 
without changing initial hyperparameters, \System{} and \Stale{} can decrease training times 
needed to achieve target error rates.
Across a wide range of configuration options, including training time budgets, \System{} and \Stale{} provide most of the Pareto-optimal choices.
%while maintaining the final converged error rate. 
%\System{} computes the sample probability of each example independently and as needed, 
%enabling it to be run online, and to scale to arbitrarily large datasets
% \zack{what precisely does per-example history mean here? history of what? the reader who doesn't knwo the related work will not be sure}
%does not need per-example history,
%where filtering redundant examples is particularly beneficial. 
%\greg{Any evidence for the "scalable to arbitrarily large datasets" ?}
%\System{} works in tandem with data augmentation, cutout, dropout, and batch normalization.
Sensitivity analyses also show that \System{} is robust to label error and effective across a range of selectivities.

% Contributions
This paper makes three primary contributions: 
(1) The design and evaluation of \System{} and \Stale{},
% \zack{probably better to say importance sampling,
% or eve n better ``importance sampling technique for accelerating neural network training''. 
% after all you're not making a contribution to "sampling" *in general*,
% and also not to importance sampling *in general*}
practical and effective sampling techniques for deep learning;
%(2) Measurements showing that, by focusing on improving the
%accuracy of challenging examples, \System{} reduces final error rates
%on MNIST and CIFAR10 on several modern classification networks
%and maintains comparable error rates on CIFAR100 and SVHN.
(2) Measurements showing that, compared to traditional training,
    \System{} and \Stale{} reduce training times to target errors
    on CIFAR10, CIFAR100, and SVHN by up to \MaxSpeedup{} and \MaxStaleSpeedup{}, respectively;
and (3) Comparison to a state-of-the-art importance sampling
    approach introduced in~\cite{katharopoulos18}, showing that
    \System{} and \Stale{} reduce training times needed to achieve target accuracy 
    by \SBOverKath{} and \StaleOverKath{}.

\section{Related Work}
\label{sec:related}

% Approaches that decrease number of iterations needed
  % None of these approaches are also show that they improve final test accuracy
  % Only one is not history-based
  % Some works have done this for different domains (e.g., RL, object detection)

% Approaches that improve test accuracy
  % In the face of label error
  % Curriculum learning
  % Loss reweighting
  % These ideas can potentially also be integrated into SB to further improve accuracy

% Approaches that decrease the number of iterations without importance sampling

% In this work, we provide a technique which
%\System{}
%does both.

%\subsection{Importance Sampling}
%\label{sec:related:speedup}

% \zack{too widely scoped.
% Importance sampling is a general statistical technique.
% E.g. there are times when you use importance sampling that have nothing to do with reducing variance. Should reword. ``Importance sampling is often incorporated into neural network training for
% the purpose of reducing ...''}

Several papers propose to reduce variance and accelerate neural network training.
The key idea of these techniques
is to bias the selection of examples from the training set,
selecting some examples with higher probability than others.
A common approach is to use importance sampling, where the goal is to more frequently sample
rare examples that might correspond to large updates.
Classic importance sampling techniques 
weight the items inversely proportional to the probability that they are selected,
producing an unbiased estimator of the stochastic gradient.
Previous approaches use importance sampling 
to reduce the number of training steps 
to reach a target error rate in classification~\cite{gao15,johnson18, loshchilov15},
reinforcement learning~\cite{schaul16},
and object detection~\cite{lin17, shrivastava16}.
These works generate a distribution over a set of training examples 
and then train a model by sampling from that distribution with replacement.  
They 
% rely on 
maintain historical losses for each example, requiring at least 
one full pass on all data to construct a distribution, 
which they subsequently update over the course of multiple training epochs. 
% As a result, 
Consequently, these approaches must maintain additional state
proportional to the training set in size
% ZL: true but if it's just importance weights they are storing,
%       then this additional state is just one float per example?
%       One would need a loooot of data to be concerned about this
% and they 
and rely on hyperparameters to modulate the effects of stale history.
In contrast, the base \System{} approach does not require such state, 
though some optimization options do.  \babu{Do we want to
mention infinite data sets?  e.g.  Thus, \System{} can be used  
effectively with click-streams and other very large datasets that may 
be run through just once during training. }

% A recent paper most-related to ours is~\citet{katharopoulos18}, 
The approach most related to ours \citep{katharopoulos18} %,  which 
also removes the requirement to maintain history, 
% and thereby provides a fully-online approach to importance sampling for classification. 
providing a fully-online approach to importance sampling for classification. 
Similar to \System{}, it uses extra forward passes 
instead of relying on historical data, 
allowing it to scale more easily to large datasets,
and it also makes decisions based on an up-to-date state of the network. 
Their sampling approach, however, predetermines the number of examples selected
per batch, so example selection is dictated by the distribution of losses in a batch. 
It also relies on a variable starting condition.
We compare against this technique in Section~\ref{sec:eval}.

Importance sampling and curriculum learning
are common techniques for generalizing DNNs.  
Differing philosophies motivate these approaches: 
supplying easy or canonical examples 
early in training as in self-paced learning~\cite{bengio09, kumar10}, 
emphasizing rare or difficult examples to accelerate learning, 
or avoiding overfitting~\cite{alain15, jiang15, katharopoulos17},
targeting marginal examples that the network oscillates
between classifying correctly and incorrectly~\cite{chang17}, 
or taking a black-box, data-driven, approach~\cite{jiang18, ma17, ren18}. 
These works improve target accuracy on image classification tasks, 
and datasets with high label error~\cite{jiang18, ren18}. 
These techniques, however, do not target and analyze training
speedup~\cite{bengio09, chang17, katharopoulos17, kumar10}, 
often adding overhead to the training process by, e.g.,
training an additional DNN~\cite{katharopoulos17, jiang18, zhang19}
or performing extra training passes on a separate validation set~\cite{ren18}.
For instance, ~\citet{zhang19} requires running an additional DNN asynchronously
on separate hardware to speed up training.

\section{Loss-Based Sampling with \System{}}
\label{sec:design}

% \zack{The captions in these figures takes up so many lines and is so narrow, what's up with the formatting? (Figure 2\&3)}
\begin{figure*}[t]
%\squeeze{}
\hspace*{\fill}%
\begin{minipage}[t]{0.45\linewidth}
\centering
    \includegraphics[width=.7\linewidth, clip]{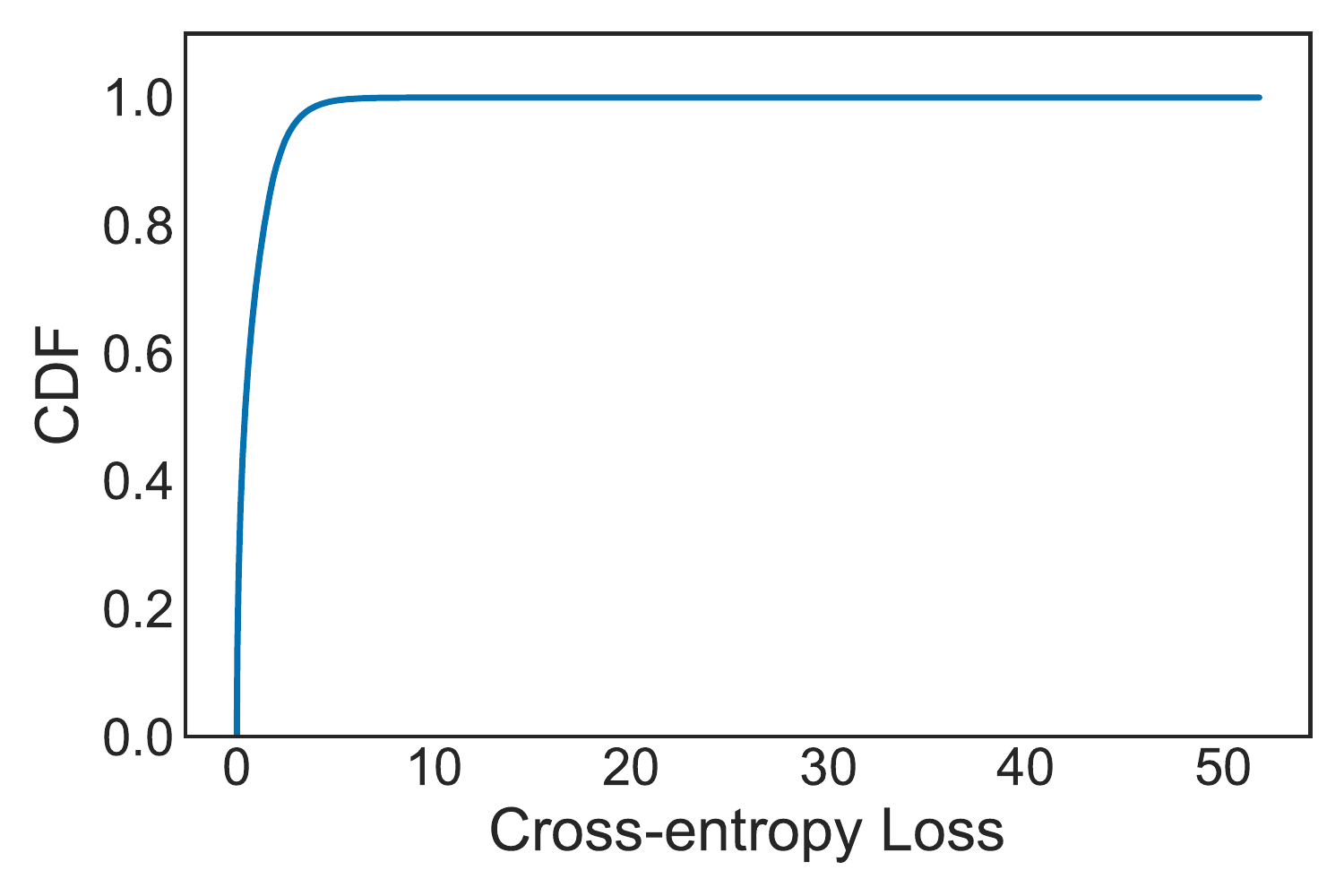}
    % \vspace{-2em}
    \caption{\small{Snapshot of a CDF of cross-entropy losses whentraining
    MobilenetV2 with CIFAR10}}
    \label{fig:sampling-percentile}
\end{minipage}%
    \hfill%
\begin{minipage}[t]{0.45\linewidth}
\centering
    \includegraphics[width=.7\linewidth, clip]{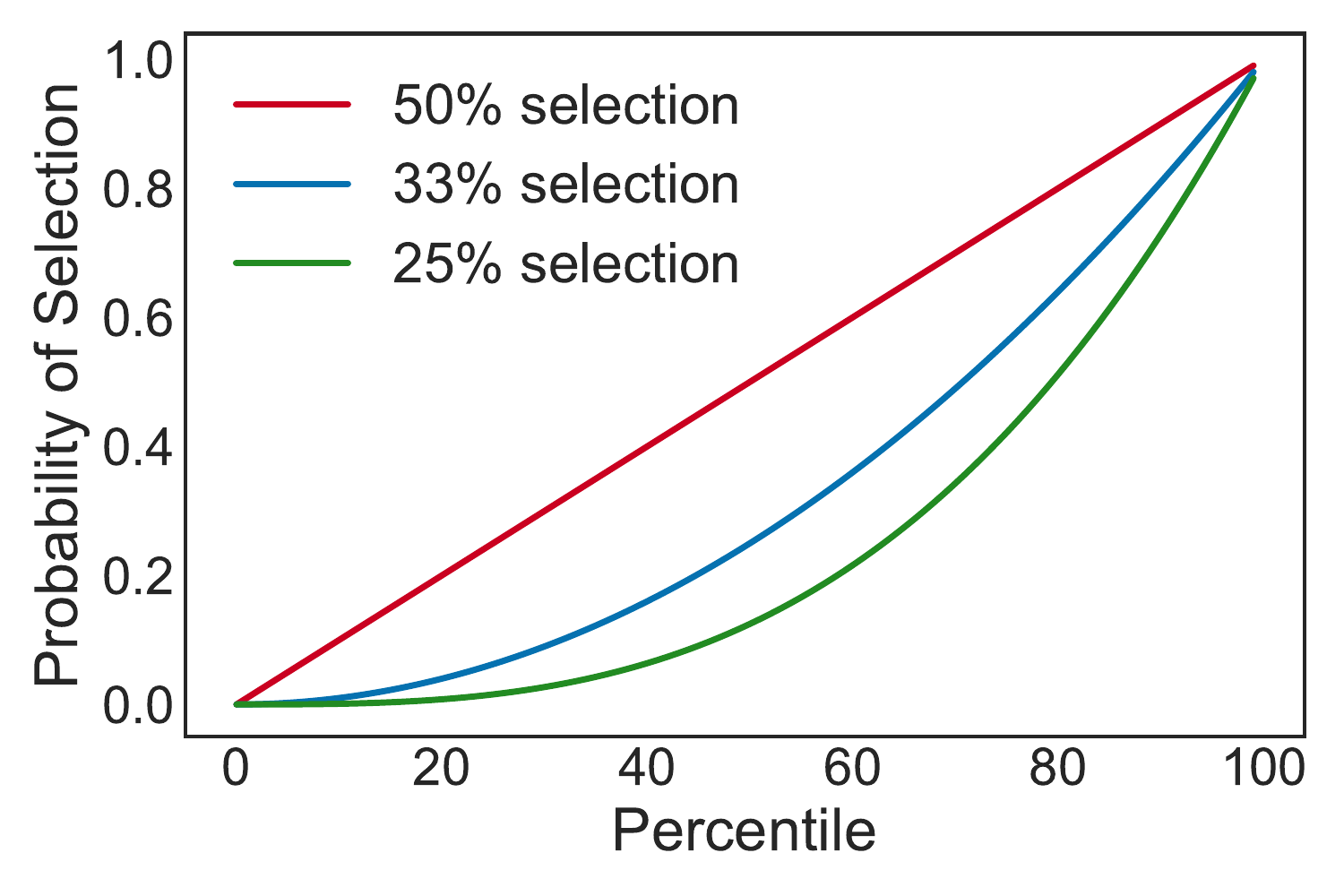}
    % \vspace{-2em}
    \caption{\small{\System{} calculates the probability of selection using $\mathcal{L}(d)$ as a percentile of the $R$ most recent losses
    % ZL:   I think the reader has gotten this point by now
    % \SB{} samples
    % high-loss examples 
    % with higher probability.
    }}
    \label{fig:sampling-probability}
\end{minipage} 
\hspace*{\fill}%
%\squeeze{}
\end{figure*}

\subsection{Background}

\System{} can be applied to standard mini-batch stochastic gradient descent 
(SGD), and is compatible with variants such as AdaGrad, RMSprop, and Adam that differ only in learning rate 
scheduling. The goal of SGD is to %to learn loss 
find parameters $\mathbf{w}^{\ast}$ that minimize the sum of the losses
$\mathcal{L}$ for a model $f(\mathbf{w})$ with $d$ parameters
over all points (indexed by $i$) in a dataset $\mathcal{D}$ consisting of $n$
examples ($\mathbf{x}_i$, $y_i$). 
$$
% \label{eq:sgd1}
  \mathbf{w}^{\ast} = \text{argmin}_{\mathbf{w}\in \mathbbm{R}^{d}}
  \sum_{i=1}^n
  \mathcal{L}(f_{\mathbf{w}}(\mathbf{x}_i), y_i)
$$
% \gauri{Notation problem with the above equation. You are summing over $i$ from $1$ to $N$ but $i$ does not appear in the summand. One option is to denote the model by $\Phi(\mathbf{w})$ and use $\xi_i$ instead of the second $\mathcal{D}$ to denote the $i^{th}$ training example. That is, $\mathcal{L}(\Phi(\mathbf{w}), \xi_i)$}
% 
SGD proceeds in a number of iterations,
at each step selecting a single example $i$ and updating the weights 
by subtracting the gradient of the loss multiplied by a step-size parameter $\eta$.
% 
% Minibatch SGD proceeds in a number of iterations,
% at each step
% At each step in SGD samples a mini-batch
% $\mathcal{M}_{t}$ with size $n$ from $\mathcal{D}$ and applies the update
% %
% \begin{equation}
$$
    % \label{eq:sgd2}
	\mathbf{w}_{t+1} = \mathbf{w}_t - \eta_t \nabla_{\mathbf{w}}\mathcal{L}(f_{\mathbf{w}_t}(x_i), y_i)
$$
% \end{equation}
% where $\eta_t$ is a scheduled learning rate.
In minibatch gradient descent, 
% Typically, one selects % approach is to select 
at each step, one selects a subset of examples  $\mathcal{M}_t$, 
often by sampling from $\mathcal{D}$ at random without replacement, 
traversing the full training set once per epoch, 
applying the update 
$$
\mathbf{w}_{t+1} = \mathbf{w}_t - \eta_t \sum_{x_i, y_i \in \mathcal{M}_t}\nabla_{\mathbf{w}}\mathcal{L}(f_{\mathbf{w}_t}(x_i), y_i)
$$
We refer to this approach as \Scan{}.
% and represent a batch generated by \Scan{} 
% with $\mathcal{M}^{\mathcal{U}}_t = \mathcal{U}(\mathcal{D})$. 

% 
% \gauri{I believe that standard method (intended by the original creators of SGD) is to sample uniformly at random WITH replacement. Most convergence analyses assume this sampling method. However in practice sampling without replacement is used for each traversal of the dataset.}

% 
% \begin{equation}
%      \label{eq:sgd3}
% 	\mathbf{w}^{\mathcal{P}}_{t+1} = 
% 	\mathbf{w}^{\mathcal{P}}_t - \eta_t  \nabla(\mathcal{L}(\Phi(\mathbf{w}^{\mathcal{P}}_t), \mathcal{M}^{\mathcal{P}}_t))
% \end{equation}
% 
% ZL:   This makes it look you are stating as a technical fact that \System{} accelerates training. 
%       This shoudl be stated instead as a goal
%  
% such that loss tends to decrease more quickly 
% compared to when the network is trained with 
% uniformly-sampled batches:
% \mk{Shouldn't the w on the right have a little superscript U?}

% \begin{equation}
%     \label{eq:sgd4}
%     \mathbb{E}[\mathcal{L}(\Phi(\mathbf{w}^{\mathcal{P}}_{t+1}), \mathcal{D}] <
%     \mathbb{E}[\mathcal{L}(\Phi(\mathbf{w}_{t+1}), \mathcal{D}]
% \end{equation}

\begin{figure*}[t]
%\squeeze{}
% \hspace*{\fill}%
\begin{subfigure}[b]{0.45\textwidth}
\centering
    \includegraphics[width=.8\linewidth, clip]{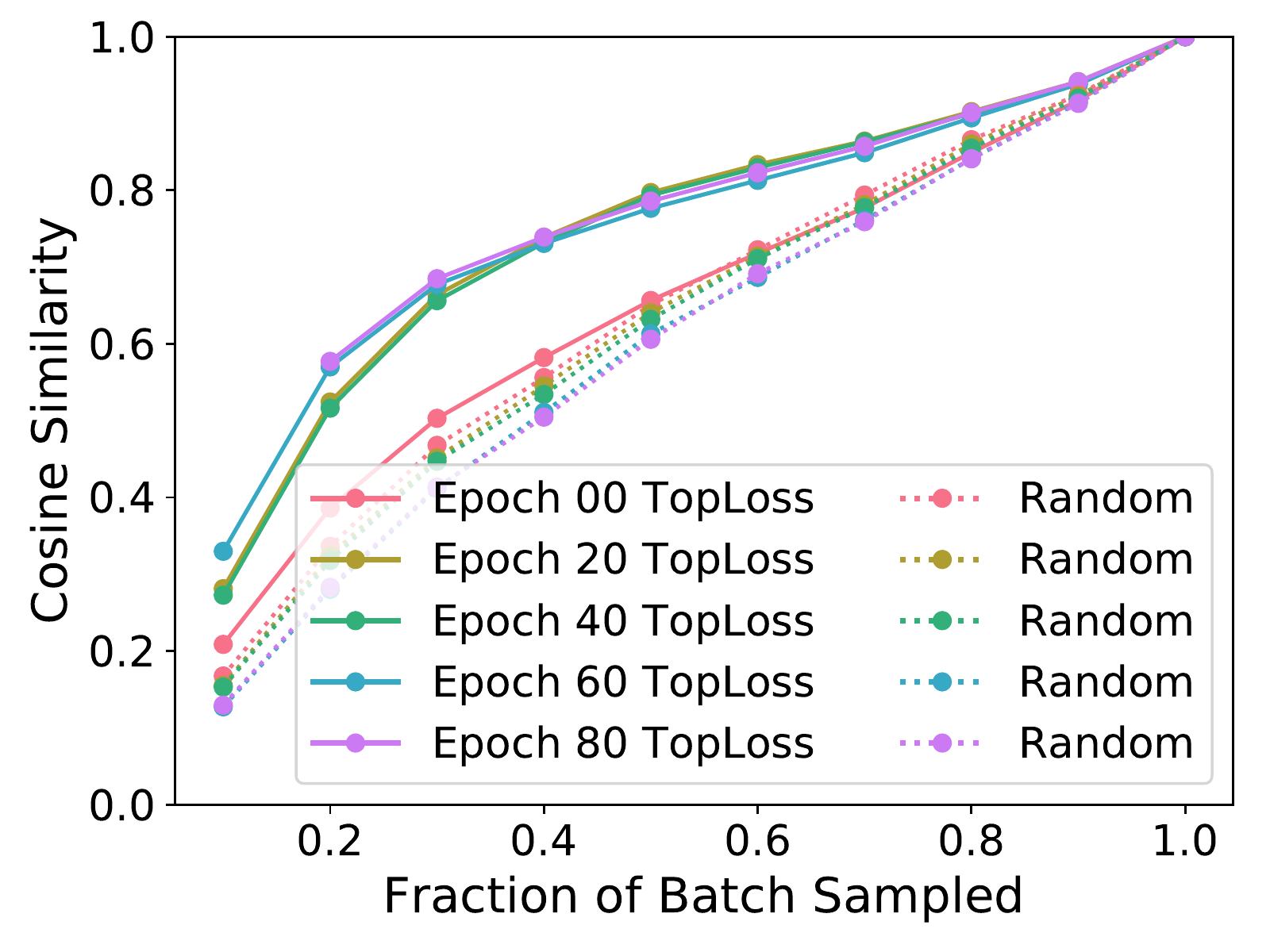}
    % \vspace{-2em}
    \caption{Cosine similarity}
    \label{fig:cosine-sim}
\end{subfigure} 
    \hfill%
\begin{subfigure}[b]{0.45\textwidth}
\centering
    \includegraphics[width=.8\linewidth, clip]{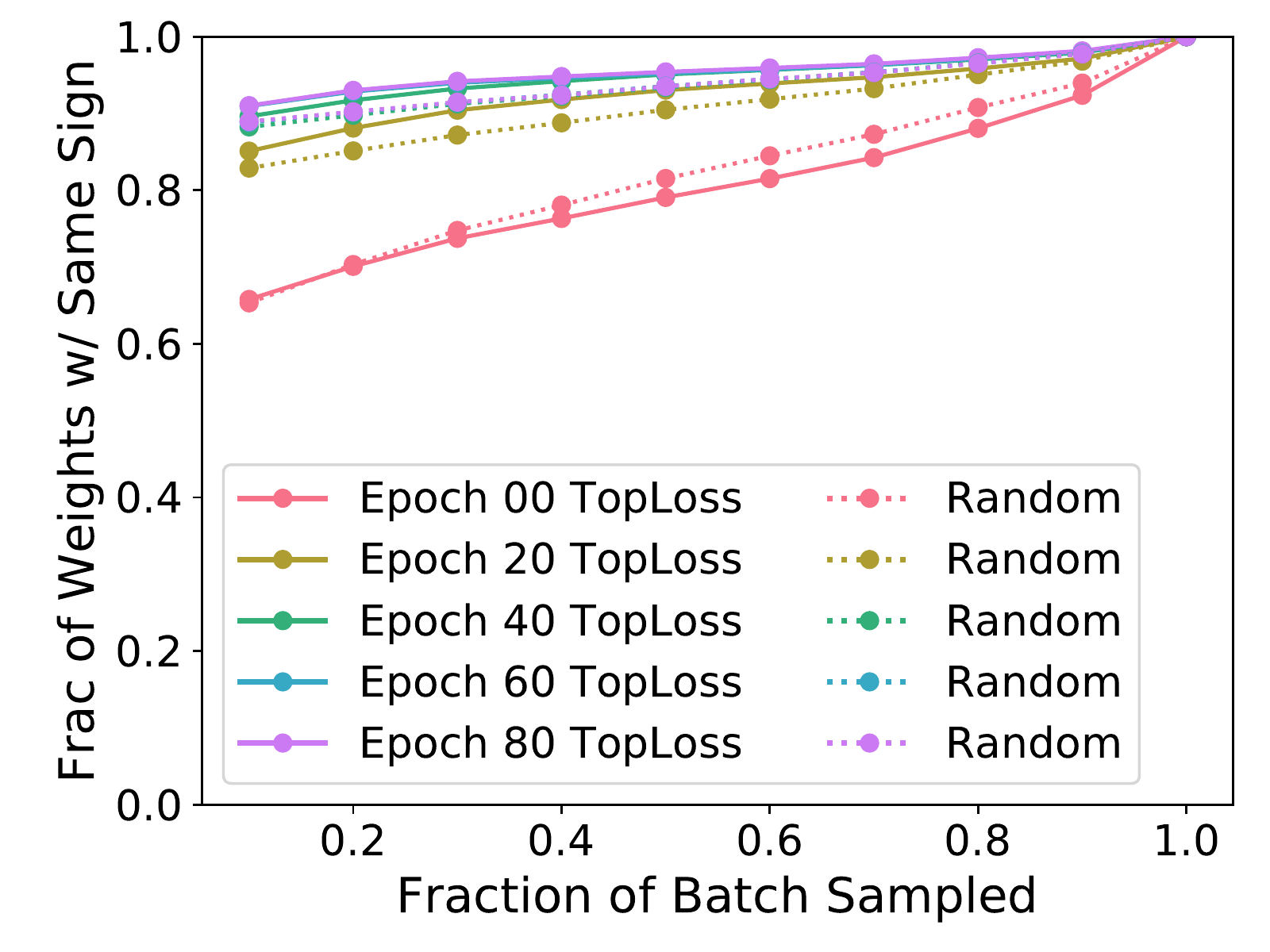}
    % \vspace{-2em}
    \caption{Fraction of gradient components with same sign}
    \label{fig:frac-same}
\end{subfigure}%
\caption{
% We show two measures of 
Similarity between gradients calculated 
on the original batch and subsampled batches when training MobilenetV2 on CIFAR10. 
After the first epoch,
high-loss examples 
% maintain higher 
are more similar than random
subsampling by both cosine similarity and fraction of weights with same sign.
}
\hspace*{\fill}%
%\squeeze{}
\end{figure*}

\begin{comment}
\begin{algorithm}
\begin{algorithmic}
\STATE \texttt{function Train}$(data, bSize, sSize)$
\bindent
\STATE $batch_{bp} \leftarrow []$
\FOR{$batch_{fp}\textit{ in data.getBatches(sSize)}$}
  \STATE $losses \leftarrow n.Forward(batch_{fp})$
  \FOR{\textit{exp, loss in zip(examples, losses)}}
      \STATE $prob \leftarrow sb.CalcProb(loss)$
      \STATE $chosen \leftarrow choose(prob)$
      \IF{$chosen$}
          \STATE $batch_{bp}.append(exp, loss)$
      \ENDIF
      \IF{$batch_{bp}.size == bSize}
          \STATE $n.Backward(batch_{bp})$
          \STATE $batch_{bp} \leftarrow []$
      \ENDIF
  \ENDFOR
\ENDFOR
\eindent
\end{algorithmic}
\captionof{algorithm}{\System{} training cycle.}
\label{alg:sb1}
\end{algorithm}
\end{comment}

%\begin{comment}

\begin{algorithm}
\begin{algorithmic}
\STATE $\it{function\; Train(data, bSize)}$
\bindent
\STATE $batch_{bp} \leftarrow []$
\FOR{$batch_{\it{fp}}\textit{ in data.getBatches(bSize)}$}
  \STATE $losses \leftarrow n.\it{Forward}(batch_{\it{fp}})$
  \FOR{\textit{exp, loss in zip(examples, losses)}}
      \STATE $prob \leftarrow sb.\it{CalcProb}(loss)$
      \STATE $chosen \leftarrow choose(prob)$
      \IF{$chosen$}
          \STATE $batch_{bp}.append(exp, loss)$
      \ENDIF
      \IF{$batch_{bp}.size == bSize$}
          \STATE $n.\it{Backward}(batch_{bp})$
          \STATE $batch_{bp} \leftarrow []$
      \ENDIF
  \ENDFOR
\ENDFOR
\eindent
\end{algorithmic}
\captionof{algorithm}{\System{} training cycle.}
\label{alg:sb1}
\end{algorithm}

\begin{algorithm}
  \begin{algorithmic}
    \STATE $hist \leftarrow deque(\it{histSize})$
    \STATE $\it{function\; CalculateProbability(loss)}$
        \bindent
        \STATE $hist.add(loss)$
        \STATE $perc \leftarrow percentile(hist, loss)$
        \STATE $prob \leftarrow perc^\beta$
        \RETURN $prob$
        \eindent
  \end{algorithmic}
\captionof{algorithm}{\System{}'s probability calculation.}
\label{alg:sb2}
\end{algorithm}
%\end{comment}

\subsection{\System{}}

%%%% Talk about algorithm

% \System{} (\SB{}) is a  mechanism for selecting $\mathcal{M}^{\mathcal{P}}_{t}$.  
% Our batch generation process is a function 
% of the current loss of each example 
% as determined by a forward pass of the network. 
% \begin{equation}
% \mathcal{P}_{t}(\mathcal{D}, \mathcal{L}(\mathbf{w}_{t}, \mathcal{D}))
% \end{equation}

\System{} also traverses the training set once per epoch, but, like other selection-based acceleration 
techniques, it generates batches using a non-uniform selection criteria
% $\mathcal{M}^{\mathcal{P}}_t = \mathcal{P}(\mathcal{D})$
designed to require fewer backward pass calculations to reach a given loss.
% \System{} (\SB{}) is a  mechanism for selecting $\mathcal{M}^{\mathcal{P}}_{t}$.  
% Our batch generation process is a function 
To construct batches, \System{} selects each example with a probability that is a function
of its current loss % example
as determined by a forward pass through the network:
$\mathcal{P}(\mathcal{L}(f_{\mathbf{w}}(\mathbf{x}_i), y_i)).$

In each epoch, \System{} randomly shuffles 
the training examples $\mathcal{D}$ and iterates over them in the standard fashion.  
However, for each example $i$,
after computing a forward pass to obtain its loss 
$\mathcal{L}(f_{\mathbf{w}}(\mathbf{x}_i), y_i)$,
% we use the output of a forward pass $\Phi(d_i)$ 
% after computing its forward pass loss 
% represent a batch generated by \Scan{} with 
% $\mathcal{M}^{\mathcal{U}}_t = \mathcal{U}(\mathcal{D})$. 
% to compute the corresponding 
% ZL:   training 
% loss $\mathcal{L}(d_i)$. 
% The key new aspect of training with \System{} 
% is that 
\System{} then decides whether to include the example for a gradient update
% the batch to be % $\mathcal{M}$
by selecting it with probability $\mathcal{P}(\mathcal{L})$ that is a function of the current loss. 
Selecting a sufficient number of examples for a full batch ($\mathcal{M}_t$) for a gradient update
typically requires forward pass calculations on more than  $\mathcal{M}_t$ examples. After collecting a full
batch, SB updates the network using gradients calculated based on this batch. 
% $\mathcal{M}^{\mathcal{P}}_{t}$ 
% which were accumulated during the preceding forward passes. 
Alg.~\ref{alg:sb1} details this algorithm.

%%%% Talk about probability

%
% ZL:   This is a bit too elementary and will make the reviewer suspicious 
%       that this is not from ML ppl. Compactifying here 
%       to keep the focus where the impact is
%       and not linger too much on basic details
%       A good NeurIPS reviewer will expect information density...
% 
% With this formulation, we can express both 
% selective and non-selective update strategies.  
% For example, standard non-selective training is achieved
% by setting $P_{d_i}=1$ for all $d_i$, 
% which is how we implement \Scan{} for comparisons in Section~\ref{sec:eval}.  
Setting the selection probability to $1$  for all examples expresses standard minibatch SGD.
%In \SB{}, we wish to implement a heuristic whereby examples with higher loss 
For \System{}, we develop an intuitive heuristic whereby examples with higher loss
are more frequently included in updates (Figure~\ref{fig:hard}), 
while those with the lower losses (Figure~\ref{fig:easy}) are included less frequently.
% are more likely to be suppressed.  
Our experiments show that suppressing gradient updates for 
low-loss examples has surprisingly little impact on the updates. 
% \angela{Zack -- what are your thoughts on the addition of 4a and the text here?}
% \babu{Yikes!  With this explanation, figure 4a shows random sampling is just as good as SB!!! 
% I would prefer dropping it and leaving just the cosine similarity.}
For example, we empirically find that the sign of over 80\% of gradient weights is maintained, 
even when subsampling only 10\% of the data with the highest losses (Fig.~\ref{fig:frac-same}). 
% Since a large amount of the information of weights are stored in
% the sign of the gradients~\cite{bernstein18}, 
Since recent research has demonstrated that the sign of the gradient alone
is sufficient for efficient learning \citep{bernstein18},
this bodes well for our method.
% we have reason to believe that subsampling
% could converge similarly compared to no sampling. 
Moreover, gradients calculated with only the highest loss examples maintain higher cosine similarity to those calculated with all examples as compared to 
randomly subsampling examples in a batch (Fig.~\ref{fig:cosine-sim}).
% with a more pronounced advantage of \System{} over random subsampling.

% By both metrics, after the first epoch, 
% higher similarity is maintained 
% by subsampling high-loss examples compared to random subsampling.

\gauri{This description of the Selective backprop algorithm needs to be written more formally. Describe concretely how the empirical CDF is constructed and how it is updated. Overall I think it would be better to give a step-wise description of the Selective Backprop algorithm. Currently, the description is interspersed with intuition about why the proposed idea will work well. }
\angela{I agree with Gauri here. SB's description should be more explicit}

Alg.~\ref{alg:sb2} details our heuristic for setting  $\mathcal{P}(\mathcal{L})$. We set
$\mathcal{P}(\mathcal{L})$ to be a monotonically increasing function of the CDF of
losses across the example set. In Figure~\ref{fig:sampling-percentile}, we show
an example of historical losses snapshotted during training. 
% ZL:   never say "of course" or "obviously" 
%       if it were truly obvious we wouldn't need to say it
% Of  course, 
Because recomputing the complete CDF after each update is not practical, 
we approximate the current CDF using a running tally 
of the losses of the last $R$ examples, denoted by $\text{CDF}_{R}$:
% 
%For each example, we use the $\Phi(d_i)$ to compute a probability of training,
%which is a function of $\mathcal{L}(d)$ as a percentile of the $R$ most recent
%example losses.
% 
\begin{equation}
  \label{eq:prob}
     \mathcal{P}\big(\mathcal{L}(f_{\mathbf{w}}(\mathbf{x}_i), y_i)\big)  = \big[\text{CDF}_{R}\big(\mathcal{L}(f_{\mathbf{w}}(\mathbf{x}_i), y_i)\big)\big]^{\beta},
\end{equation}

where $\beta>0$ is a constant that determines \System{}'s level of selectivity
and thus allows us to modulate the bias towards high-loss examples where larger
values produce greater selectivity (Figure~\ref{fig:sampling-probability}).  We
include a sensitivity analysis of $\beta$ in Section~\ref{sec:eval}.

%\subsection{Optimizations to reduce time spent in the forward pass.}
\subsection{Reducing selection overhead}

\System{} accelerates training by reducing the number of backward passes needed
to reach given levels of loss.  In our experiments (Section~\ref{sec:eval}), we
find that after reducing backward passes with \System{}, the largest remaining
fraction of training time is the full (original) complement of forward passes
used to select the \System{} batches. We distinguish these forward passes
from the forward passes used for training by referring to them as ``selection passes''
in the rest of the paper.
%show that \System{} accelerates training without reducing forward passes or
%performing them asynchronously.
This section describes four approaches to reducing the time spent in selection passes
of \System{} training, thereby further reducing overall training time.
%However, there are a number of potential optimizations to further accelerate
%training by reducing the cost of forward passes. 

\textbf{Re-using previous losses.} \System{}'s selection pass uses the
latest model parameters to compute up-to-date losses for all training examples
considered.  We define and evaluate a \System{} variant called \Stale{} that executes
selection passes every $n\textsuperscript{th}$ epoch. The subsequent $(n-1)$ epochs
reuse the losses computed by the previous selection pass to create the
backprop batch. The losses are reused in the following epoch(s), but only for
batch formation.
%(Normal training with selected batches is still performed, including forward and backward passes.) 
Intuitively,  if an example is deemed important in a given
selection pass, it will also have a high probability of being selected in the next $(n-1)$ epochs.
\Stale{} with $n=1$ is \System{}.  We evaluate \Stale{} in
Section~\ref{sec:eval} and find that, typically, it reduces selection pass cost significantly without impacting final accuracy. 
%finding that it achieves most of the potential benefit from reducing selection times.\babu{not exactly clear what this means}

\textbf{Using predicted losses.} Rather than simply re-using the loss from an
earlier epoch for selecting examples, one could construct a \System{} variant
that predicts the losses of examples using historical loss values.  To make the
problem easier, instead of predicting the loss directly, one could predict
whether the loss is high enough to cross the threshold for selection.  
%(e.g., make the random
%draw from the uniform distribution and then predict whether or not the
%probability would be greater than the value drawn).
We evaluated various prediction approaches, including tracking a exponentially
weighted average of historical losses and using historical losses to train 
a Gaussian process predictor. None outperformed the simpler \Stale{} approach.

\textbf{Pipelining loss computation.} Given multiple computation engines, one
could construct a \System{} variant that selects examples for batch $N+1$
%(computes losses for examples and selects from among them)
on a separate engine while training with batch $N$ is ongoing.
Such an approach would require using losses computed from stale versions
of the model, but could mask nearly all training delays for selection
and could do so without the ``history size''
concerns that would arise for the above approaches when training with giant or
continuous datasets.  Running a separate model for loss computation would,
however, introduce a new overhead of occasionally syncing the selection model
to reflect changes to the training model.\footnote{Using modern deep learning
frameworks such as PyTorch, we found that 
copying a new model and moving it to a second
device can take up to a minute.} This introduces a new trade-off between
frequency of syncing the selection model and the amount of staleness introduced
to the selection process.

Although one could use equivalent GPUs for such pipelining, it is unclear that
this would be better than data-parallel training. 
Rather, we think the natural application of the pipelining approach would
be in combination with the inference accelerators discussed next---that is,
a low-cost inference accelerator could be used to compute losses for example
selection, and then a powerful compute engine could be used for training on
batches of selected examples.
%We leave the implementation and evaluation of an asynchronous \SB{} to future
%work.

%We also show that we are able to achieve speedups without a second device, by
%performing the additional inferences serially.

\begin{comment}
\textbf{backward pass typically takes 2x the time of the forward pass.} In
propagation, we calculate the DNN loss gradient of each weight by
repeatedly applying the chain rule. During the forward pass, we calculate the
loss by performing one convolution per layer to get the output.  During the
backward pass, to calculate the gradient of the loss respect to the weights,
we need to repeatedly apply the chain rule on each layer.  Therefore, for each
layer, we calculate two gradients: one w.r.t. the input data and one w.r.t
layer weights. For convolutional neural networks performing image
classification, most (>90\% for ResNet18) of the computation is spent
in convolutional layers. Therefore the asymmetry between the time spent during
the forward and backward convolutional calculations often manifests in an
asymmetry in total computation time. We confirm that this asymmetry holds across
multiple GPUs and batch sizes (Figure.~\ref{fig:asymmetry}).
\end{comment}

\textbf{Inference accelerators.} For general-purpose hardware and training
frameworks, the cost of the backward pass is approximately twice  the forward pass cost.
This is because during the forward pass, we calculate one convolution per
convolutional layer, whereas in the backward pass we perform two: one convolution
to calculate the gradients w.r.t the input data and another w.r.t to the layer
weights \cite{nun2018}.  But, a variety of inference acceleration
approaches, such as reduced precision or quantization, may enable specialized hardware accelerators to run forward passes $\approx$10x faster than a
backward pass on a modern GPU~\cite{jouppi17}.  Since \SB{} selects examples by
running a forward pass, it can use such accelerators. Although
aggressive forward-pass acceleration can affect the outcome of training, use of inference acceleration for \System{}'s selections
may not have the same negative consequences. We leave exploration of this
approach to reducing selection time to future work, but include it in this list
for completeness.

\section{Implementation Details}
\label{sec:implementation}

%In order to validate our hypotheses discussed in Section~\ref{sec:design}, we
We built prototypes of \SB{} for PyTorch 0.4.1 and Keras
2.2.4; the Section~\ref{sec:eval} evaluation is based on the former.

To add \SB{} into existing training code, we introduce a mathematically simple
probabilistic filtering step to training, which down-selects examples
used for updates.  Filtering starts by calculating the loss for each
example using a forward pass. \SB{} adds the loss to $\text{CDF}_{R}$
(implemented using a bounded queue), and calculates what percentile of losses
it represents. Using this percentile, \SB{} calculates the selection
probability $\mathcal{P}$, and probabilistically adds this example to
the minibatch for training.  We measure the overhead introduced by
\SB{}'s selection step, excluding time spent in selection passes,
to $\approx$3\% of overall training time.

\System{}'s lightweight filtering step is simple
to implement and requires few changes to existing training code.
In traditional setups, data is formed into minibatches for training. In \SB{},
data is formed into selection minibatches and fed into \SB{}'s
filtering mechanism. \SB{} performs forward passes of selection minibatches (``selection passes''), forms a training minibatch of selected data examples, and passes it to the original training code. Therefore, the
training code can be agnostic to \SB{}'s filtering mechanism, allowing \SB{} to
work in tandem with any training optimizer (e.g., SGD, Adam) and
common optimizations such as batch normalization, data augmentation, dropout,
and cutout.

\textbf{Future implementation optimizations.} Our \SB{} implementation minimizes 
changes to existing code, and some obvious potential optimizations are not currently incorporated. For instance, in our implementation,
two forward passes are performed for each selected example: one for
selection and one for training. Unless selection passes are accelerated using
reduced precision or quantization, which is not the case in our implementation, a more optimized \SB{} implementation could cache the activations obtained from the selection passes to avoid doing extra
forward passes for training, and thus eliminate the time spent in ``Forwards (training)'' for \SB{} shown in Figure \ref{fig:front-page}. Another  optimization would use a minibatch size for selection that is larger than that of training, to reduce the number of selection passes needed to populate a training minibatch.

% !TeX root = ./selective_backprop.tex

\section{Evaluation}
\label{sec:eval}

We evaluate \System{}'s effect on training with modern image classification
models using CIFAR10, CIFAR100, and SVHN. 
The results show that, compared to
traditional training and a state-of-the-art importance sampling approach~\cite{katharopoulos18},
\SB{} reduces wall-clock time needed to reach a range of target error rates by up to \MaxSpeedup{}
(Section~\ref{sec:eval:speedup1}). We show that by reducing the
time spent in example selection, one can further accelerate training by on average
\StaleOverSB{} (Section~\ref{sec:eval:speedup2}). Additional analyses show the
importance of individual \SB{} characteristics, including selection of high-loss examples
and robustness to label error (Section~\ref{sec:eval:sweep}).
Throughout the evaluation, we also show that the speedup achieved by a \SB{}
depends the learning rate schedule, sampling selectivity, and target error.
Section~\ref{sec:eval:pareto} shows that, across a sweep of configurations,
the majority of Pareto-optimal trade-off points come from \System{} and \Stale{}.
We also provide a sensitivity analysis for \SB{} in Section~\ref{sec:eval:sweep}
and the results from two additional learning rate schedules in the appendix.

\vspace{-5px}
\subsection{Experimental Setup}
\label{sec:eval:setup}

We train Wide Resnet, ResNet18, DenseNet, and MobileNetV2 
\cite{zagoruyko2016, he16, huang17, sandler18} using \SB{},
\Scan{} as described in Section~\ref{sec:design}, and our implementation of \Kath{}
\cite{katharopoulos18} with variable starting, no bias reweighting, and using
loss as the importance score criteria. We tune the selectivity of \SB{} and
\Kath{} individually for each dataset. In our evaluation, we present
the results of training Wide Resnet. To train Wide Resnet, we use the training setup specified 
by \cite{devries2017}, which includes standard optimizations including cutout,
batch normalization and data augmentation. We observe similar trends when
using ResNet18, DenseNet, and MobileNetV2, and present those results in 
the appendix. In each case, we do not retune existing hyperparameters. We report
results using the default batch size of 128 but confirm that the trends remain
when using batch size 64.

\textbf{CIFAR10.} The CIFAR10/100 datasets \cite{krizhevsky09} contain 50,000
training images and 10,000 test images, divided into 10 and 100 classes,
respectively.  Each example is a 32x32 pixel image.  We use $\textit{batch\_size}=128$
and cutout of length 16. We use an SGD optimizer with $decay=0.0005$.
We train with two learning rate schedules. In the first schedule,
we start with $lr=0.1$ and decay by 5x at 60, 120, and 160 epochs. In the second
schedule, we decay at 48, 96, and 128 epochs. We use 33\% selectivity for \SB{},
\Stale{}, and \Kath{}. Training ends after 12 hours.

\textbf{CIFAR100.} We train on CIFAR100 using the setup specified
by~\citet{devries2017}. We use $\textit{batch\_size}=128$.
and cutout of length 8.  We train with two learning rate schedules. First,
we start with $lr=0.1$ and decay by 10x at 60 and 120. In the second
learning rate schedule, we decay at 48 and 96 epochs. We use 50\% selectivity
for \SB{}, \Stale{}, and \Kath{}. Training ends after 12 hours.

\textbf{SVHN.} SVHN has 604,388 training examples and 26,032 testing examples
of digits taken from Street View images \cite{netzer11}. For the first
schedule, we initialize the learning rate to 0.1 and decay to 0.01 and 0.001 at
epochs 60 and 120. For the second, we decay at 48 and 96. We use
$\textit{batch\_size}=128$ and cutout of length 20.  We use 25\% selectivity
for \SB{} and \Stale{}, and 33\% selectivity for \Kath{}.\babu{why the
different selectivity?} Training ends after 96 hours.

\textbf{Hardware.} We train CIFAR10 and CIFAR100 on servers equipped with 
16-core Intel Xeon CPUs, 64 GB RAM and NVIDIA TitanX GPUs. We train SVHN on
servers with four 16-core AMD Opteron CPUs, 128 GB RAM, and NVIDIA
Tesla K20c GPUs.  

\greg{I commented this out (add back if desired): This is typical for modern GPUs, as noted by~\cite{convbenchmarks}
and confirmed in experimental results provided in the appendix.
It seems to be referring to some remark that probably got cut (?).  I'm guessing it's the "backprop is 2X forward", but in any case I think we can just leave it out.}

\subsection{\System{} speeds up training}
\label{sec:eval:speedup1}

\begin{figure*}
\begin{minipage}{\textwidth}
        \centering
        \begin{subfigure}[b]{0.33\textwidth}
          \includegraphics[width=\linewidth]{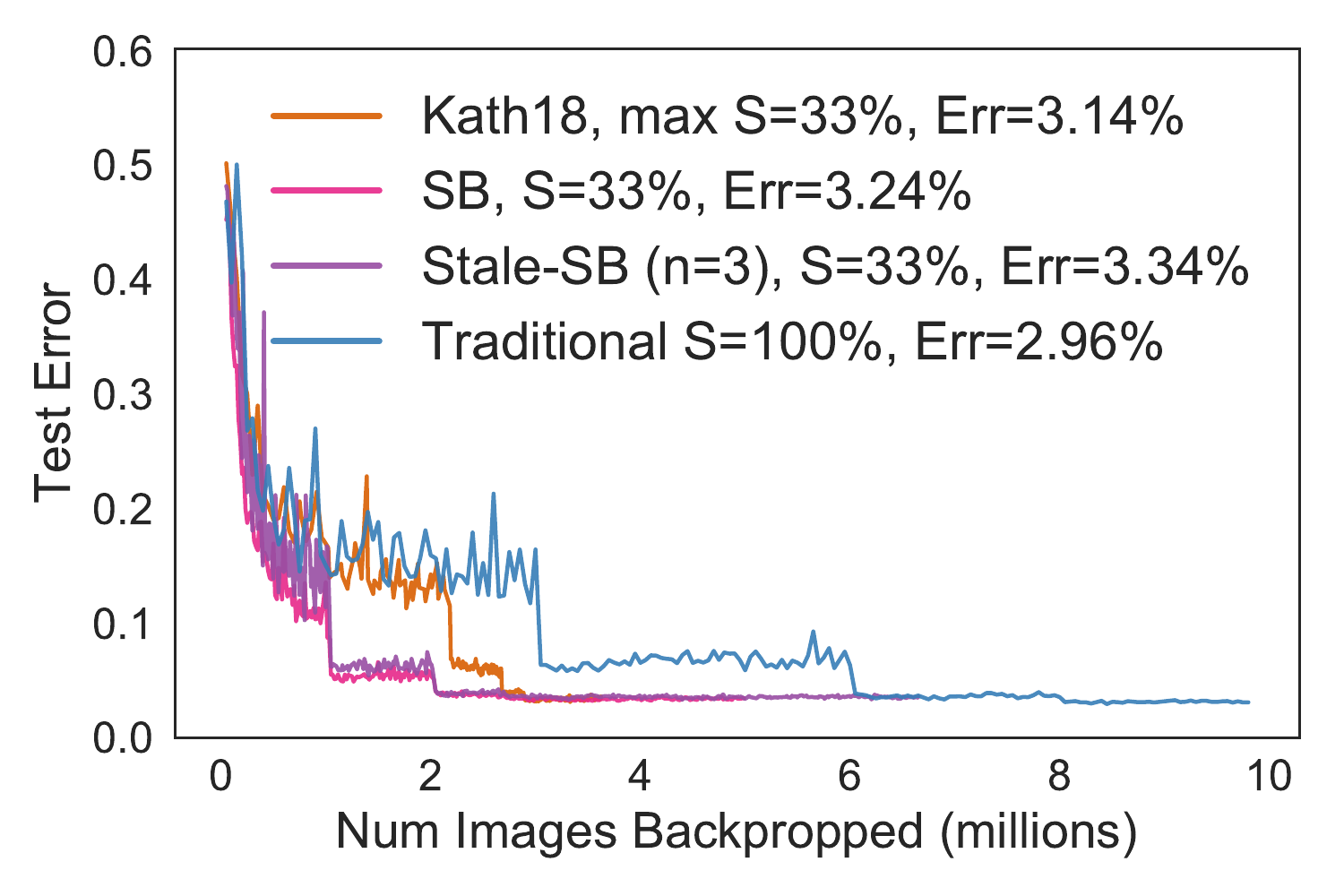}
          \caption{CIFAR10}
          \label{fig:strategy-backprops-cifar10}
        \end{subfigure}%
        \begin{subfigure}[b]{0.33\textwidth}
          \includegraphics[width=\linewidth]{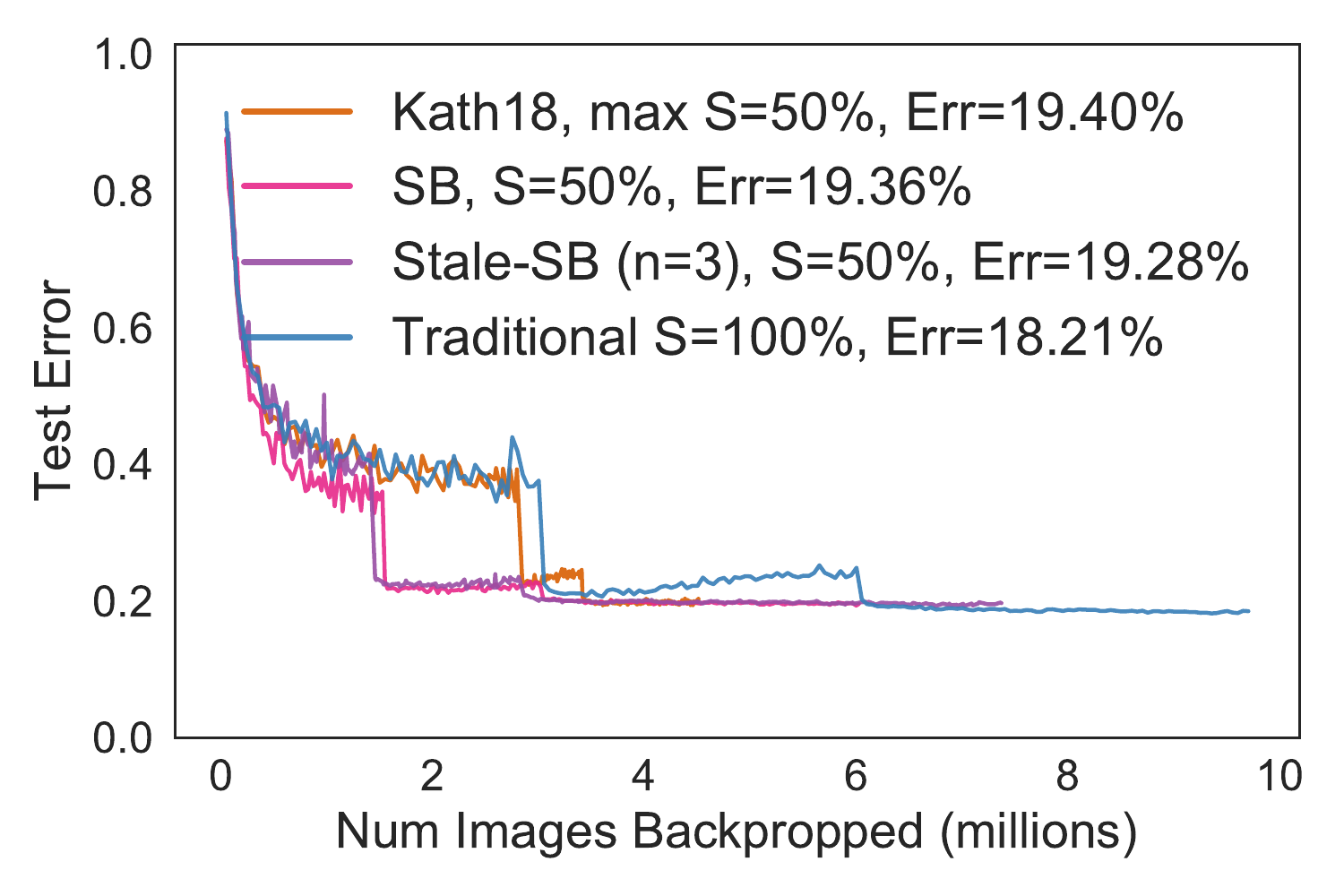}
          \caption{CIFAR100}
          \label{fig:strategy-backprops-cifar100}
        \end{subfigure}%
        \begin{subfigure}[b]{0.33\textwidth}
          \includegraphics[width=\linewidth]{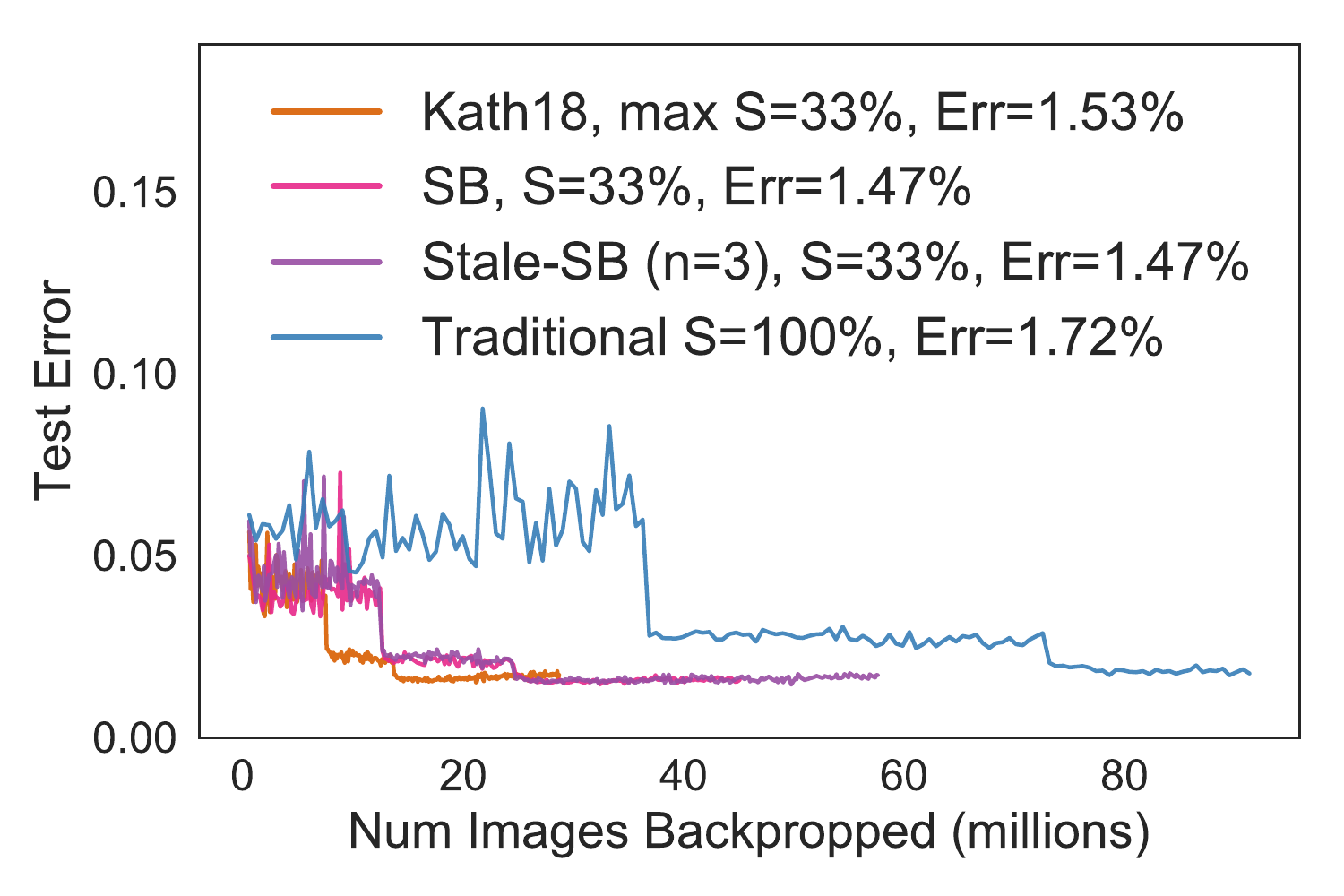}
          \caption{SVHN}
          \label{fig:strategy-backprops-svhn}
        \end{subfigure}%
        \caption{\SB{} reduces training iterations to target error. $S$ is the selectivity used, and $Err$ is the final test error reached.}
        \label{fig:strategy-backprops}
\end{minipage}
% \squeeze{}
% \squeeze{}
\end{figure*}

\textbf{\SB{} reduces training iterations to target error.} \SB{} probabilistically
skips the backward passes of examples with low loss in order to learn more per example. 
Figure~\ref{fig:strategy-backprops} shows that \SB{} reaches final or non-final target
error rates with fewer training iterations (updates to the network). This can be seen
by comparing the x-axis points at which lines for each approach reach a particular
y-axis value. Note that we plot different y-axes for different datasets.
Although the savings depends on the specific target test error rate
chosen, one way to visualize the overall speedup across different target accuracy
is by comparing the area under the three curves. \SB{} reaches nearly every test
error value with significantly fewer training iterations.

\begin{figure*}
\begin{minipage}{\textwidth}
        \centering
        \begin{subfigure}[b]{0.33\textwidth}
          \includegraphics[width=\linewidth]{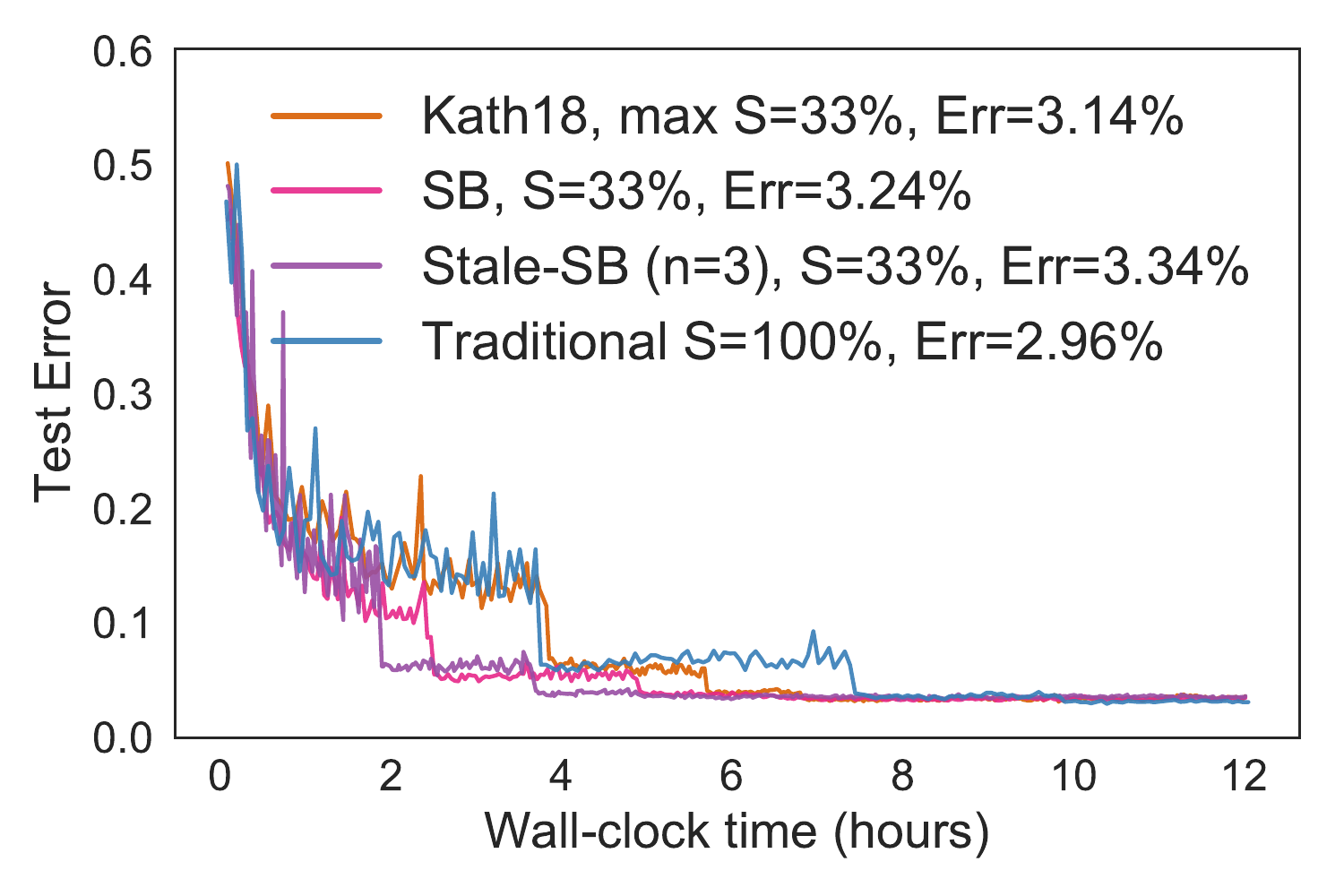}
          \caption{CIFAR10}
          \label{fig:strategy-seconds-cifar10}
        \end{subfigure}%
        \begin{subfigure}[b]{0.33\textwidth}
          \includegraphics[width=\linewidth]{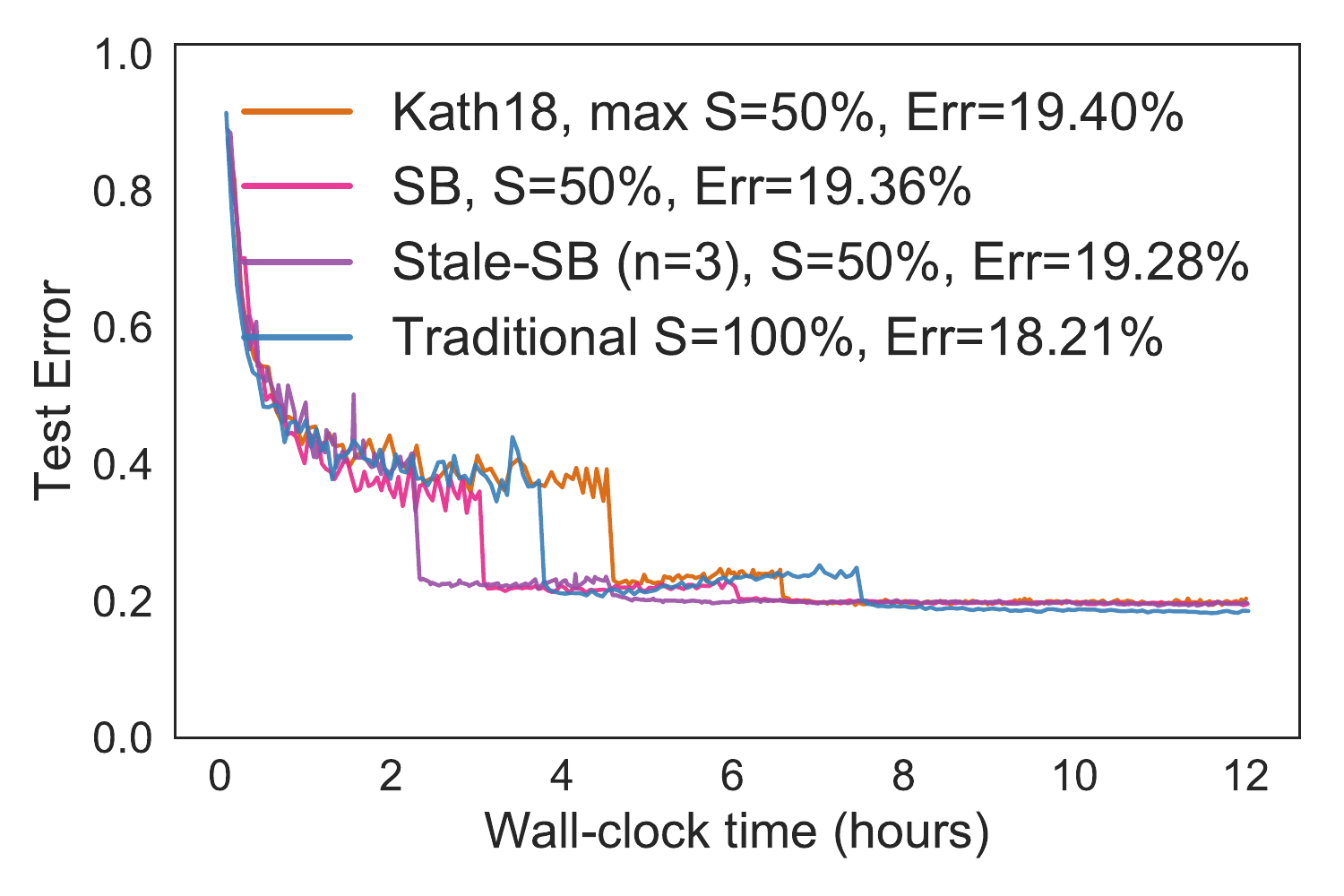}
          \caption{CIFAR100}
          \label{fig:strategy-seconds-cifar100}
        \end{subfigure}%
        \begin{subfigure}[b]{0.33\textwidth}
          \includegraphics[width=\linewidth]{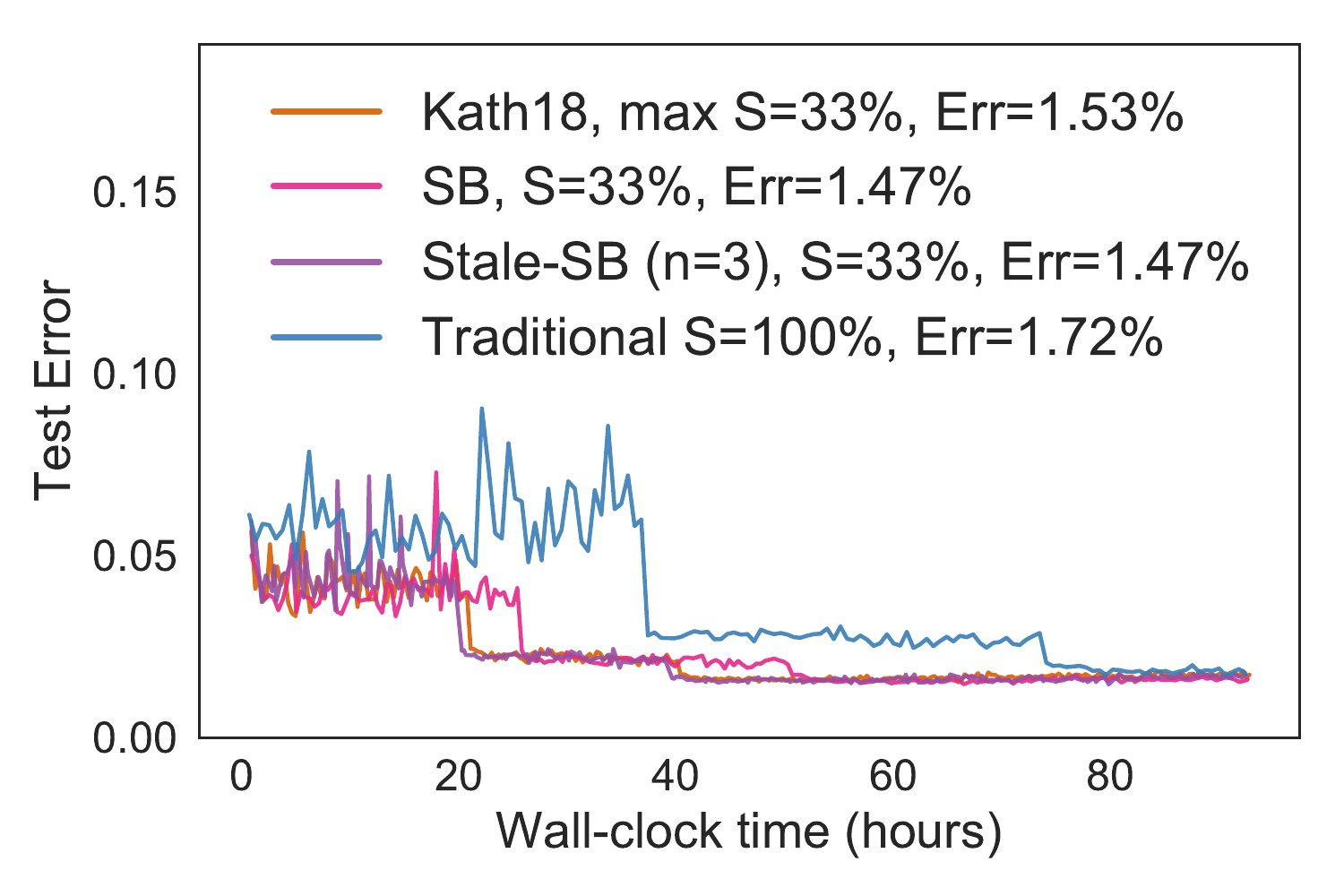}
          \caption{SVHN}
          \label{fig:strategy-seconds-svhn}
        \end{subfigure}%
        \caption{\SB{} reduces wall-clock time to target error.  $S$ is the selectivity used, and $Err$ is the final test error reached.}
        \label{fig:strategy-seconds}
\end{minipage}
% \squeeze{}
% \squeeze{}
\end{figure*}

\textbf{\SB{} reduces wall-clock time to target error.} Figure~\ref{fig:strategy-seconds}
shows error rate as a function of wall-clock time. \SB{} speeds up time
to target error rates by reducing backward passes, without
optimizations to reduce selection time discussed in Section~\ref{sec:eval:speedup2}.
Table~\ref{table:speedup} shows that for CIFAR10, \SB{} reaches within 10\%, 20\%, and 40\% of 
\Scan{}'s final error rate 1.2--1.5x faster. 
For SVHN, \SB{} provides a 3.4--5x speedup to reach 1.8\%, 2.1\%, and 2.4\%
error.

Intuitively, \SB{} is most effective on datasets with many redundant examples
or examples that are easy to learn. CIFAR100 is a more challenging dataset
for sampling as there are fewer examples per class and therefore likely
less redundancy. Despite this, \SB{} reaches within 20\% %(21.85\% error)
and 40\% %(25.49\% error)
of the \Scan{}'s final error rate 20\% faster. However, it sacrifices a small amount
of final accuracy for these speedups and does not reach within 10\% of \Scan{}'s final
error rate in the allotted training time. 
\Kath{} also accelerates training over \Scan{} by 
0.8--3.4x. Similarly to \SB{}, it is most effective on SVHN and least effective on 
CIFAR100, even leading to a small slowdown to certain target error rates.
\SB{} provides a speedup over \Kath{} of \SBOverKath{}.

\begin{figure*}
%\squeeze{}
\hspace*{\fill}%
\begin{minipage}{0.33\linewidth}
    \includegraphics[width=\linewidth, clip]{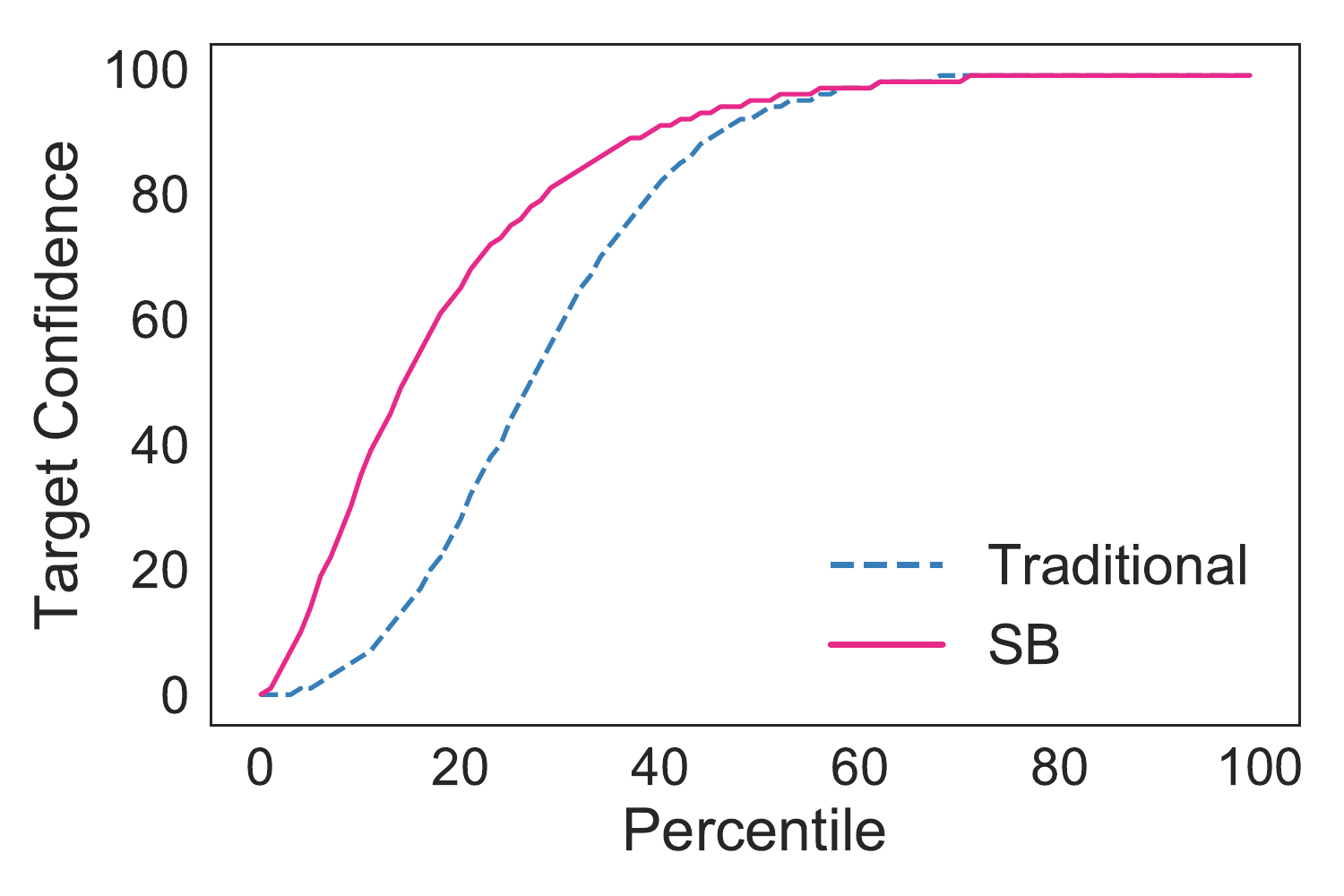}
    \vspace{-2em}
    \caption{\small{\SB{} has higher confidence in harder examples with almost no cost of
    confidence in easy examples.}}
    \label{fig:confidences-wideresnet}
\end{minipage}%
    \hfill%
\begin{minipage}{0.33\linewidth}
    \includegraphics[width=\linewidth, clip]{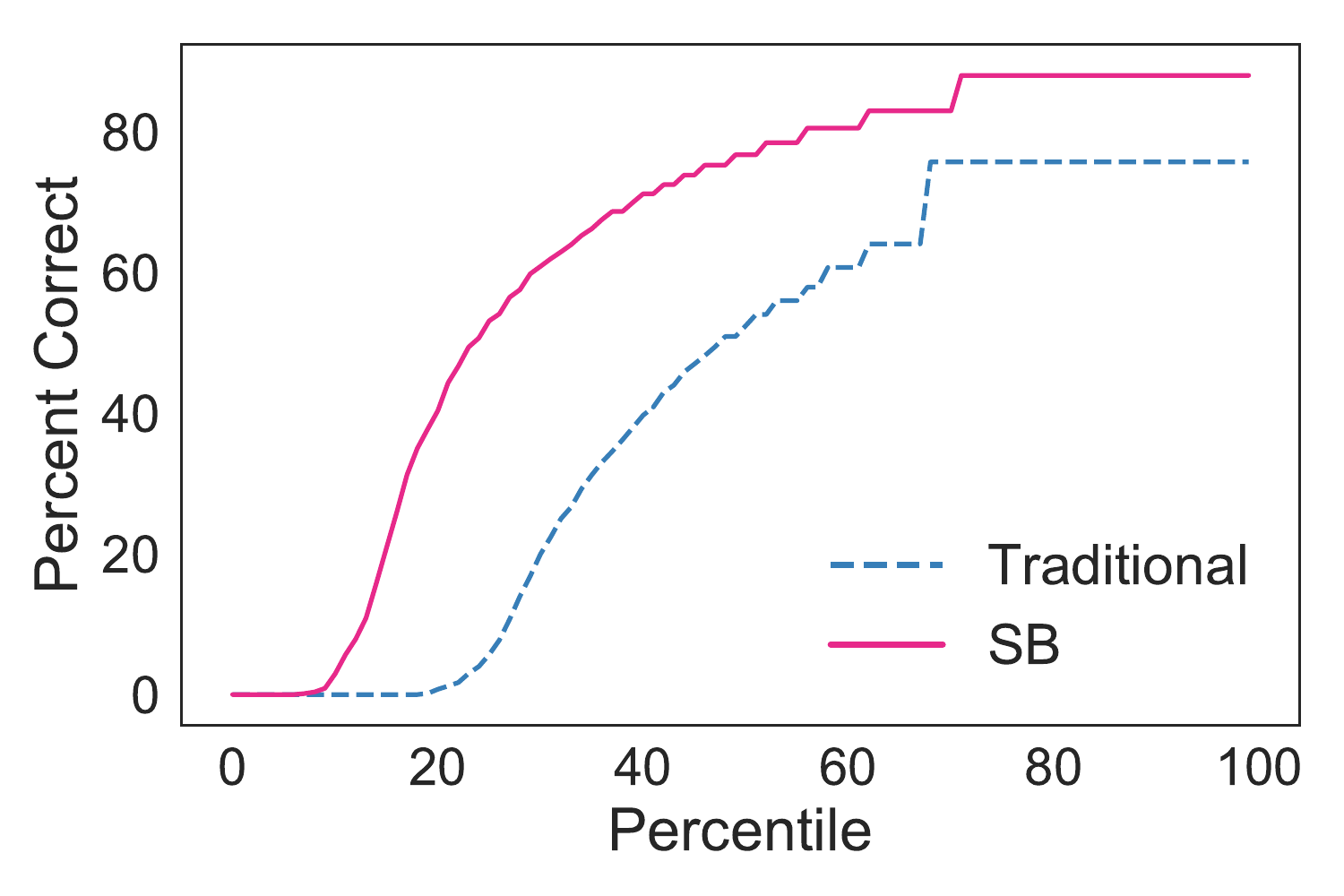}
    \vspace{-2em}
    \caption{\small{\SB{} increases accuracy of harder examples, without
    sacrificing accuracy of easy examples.}}
    \label{fig:accuracy-wideresnet}
\end{minipage} 
\hspace*{\fill}%
%\squeeze{}
\end{figure*}

\textbf{\System{} performs better on challenging examples.} \SB{} converges
faster than \Scan{} by outperforming \Scan{} on challenging examples.
Figure~\ref{fig:confidences-wideresnet} shows an inverse CDF of the network's
confidence in each ground truth label of the test set; the data represents a
snapshot in time after training \SB{} or \Scan{} for ten epochs.  For each
percentile, we plot the target confidence on the y-axis (e.g., the 20\textsuperscript{th}
percentile of target confidences for \SB{} is 55\%).
The network's classification is the class with the Top-1 confidence (using
argmax). Therefore, we cannot infer the classification accuracy of an example
solely from its target confidence (if the target confidence is $\leq 50\%$).  In
Figure~\ref{fig:accuracy-wideresnet}, we also plot the accuracy of each
percentile of examples.  Generally, examples at lower percentiles are harder
for the network to accurately classify. Using \SB{}, the network has higher
confidence and accuracy in these lower percentiles.  For instance, among the
examples at the 20th percentile of target confidences, 29\% of these examples
are classified correctly using \SB{} while only 3\% are classified correctly by
\Scan{}. While this comes at the cost of confidence in higher percentile
examples, test accuracy is not sacrificed.  In fact, \SB{} is able to
generalize better across all examples of all difficulty levels.

\begin{table*}
  \centering
  \small
  \begin{threeparttable}
  \begin{tabular}{llllll}
  \textbf{Dataset} & \textbf{Strategy} & \thead{Final error \\ of Traditional} & %
  \thead{Speedup to \\ final error $\times$ 1.1} & \thead{Speedup to \\ final error $\times$ 1.2} %
  & \thead{Speedup to \\ final error $\times$ 1.4} \\
 \toprule
 CIFAR10  & \SB{}    & 2.96\%  & 1.4x & 1.2x & 1.5x \\ % & 2.9X \\ %33%
 CIFAR10  & \Stale{} & 2.96\%  & -- & 1.5x & 2.0x \\ % & 2.0X \\ %25%
 CIFAR10  & \Kath{}  & 2.96\%  & 1.4x & 1.1x & 1.3x \\ % & 1.3X \\ %33%
 \midrule
 CIFAR100 & \SB{}    & 18.21\%  & 1.2x & 1.2x & 1.2x \\ % & 1.2X \\ %50%
 CIFAR100 & \Stale{} & 18.21\%  & 1.5x & 1.0x & 1.6x \\ % & 1.6X \\ %50%
 CIFAR100 & \Kath{}  & 18.21\%  & 1.1x & 0.8x & 0.8x \\ % & 0.8X \\ %25%
 \midrule
 SVHN     & \SB{}    & 1.72\%   & 3.4x & 3.4x & 3.5x \\ % & 4.9X \\ %20% 
 SVHN     & \Stale{} & 1.72\%   & 4.3x & 4.9x & 5.0x \\ % & 2.5X \\ %20%
 SVHN     & \Kath{}  & 1.72\%   & 1.9x & 2.8x & 3.4x \\ % & 1.8X \\ %33%
\end{tabular}
\end{threeparttable}
\caption{Speedup achieved by \SB{} and \Kath{} over \Scan{}}
% \vspace{-10px}
\label{table:speedup}
\end{table*}

\begin{figure*}
\begin{minipage}{\textwidth}
        \centering
        \begin{subfigure}[b]{0.33\textwidth}
          \includegraphics[width=\linewidth]{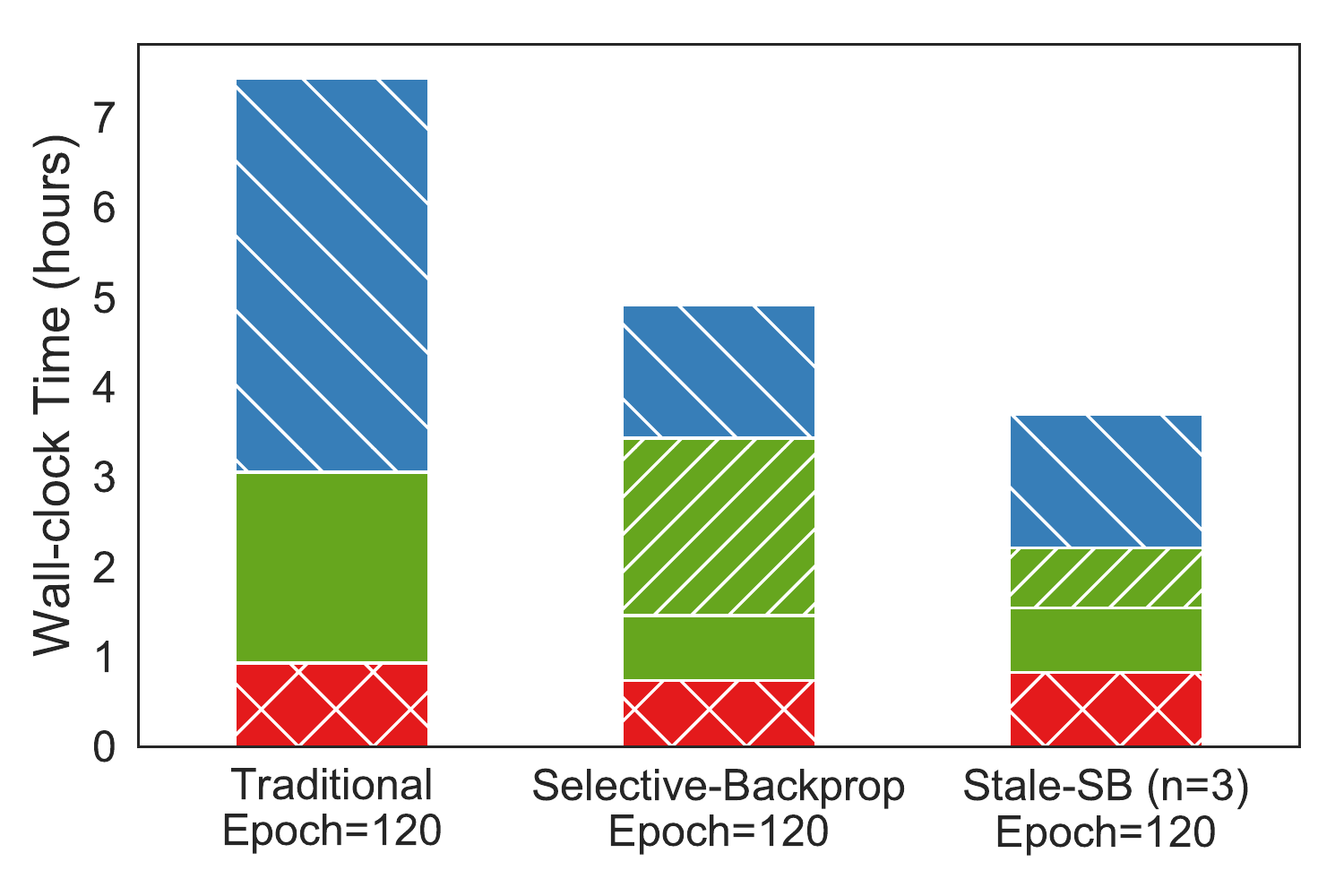}
          \caption{CIFAR10}
          \label{fig:stacked-cifar10}
        \end{subfigure}%
        \begin{subfigure}[b]{0.33\textwidth}
          \includegraphics[width=\linewidth]{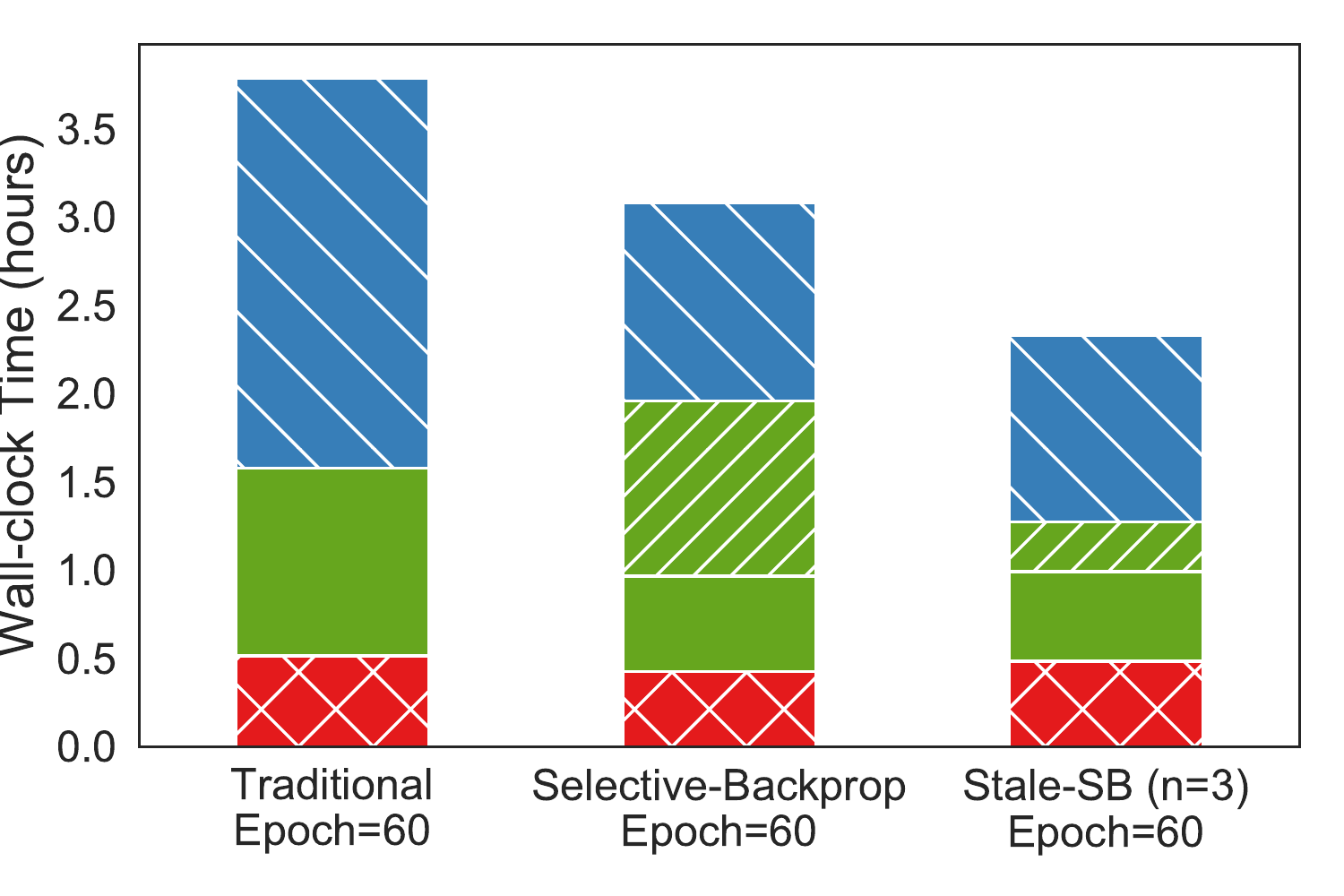}
          \caption{CIFAR100}
          \label{fig:stacked-cifar100}
        \end{subfigure}%
        \begin{subfigure}[b]{0.33\textwidth}
          \includegraphics[width=\linewidth]{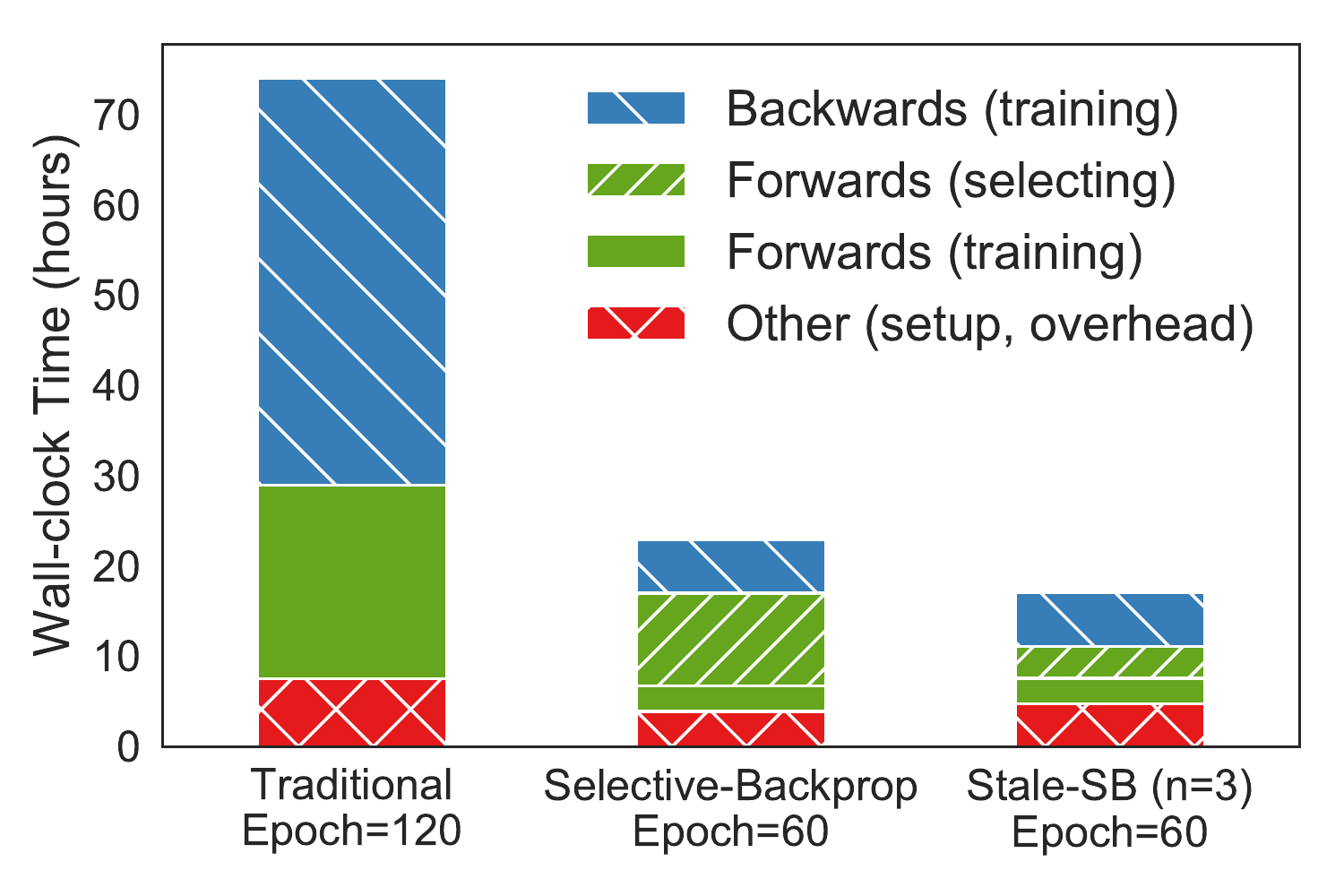}
          \caption{SVHN}
          \label{fig:stacked-svhn}
        \end{subfigure}%
        \caption{\SB{} reduces time spent in the backward pass in order to speed up time
        to the target error rate (in this case, 1.4x of \Scan{}'s final error rate). \Stale{} further accelerates
        training by reducing the time spent performing selection passes.}
        \label{fig:stacked}
\end{minipage}
% \squeeze{}
% \squeeze{}
\end{figure*}

\subsection{Reducing selection times further speeds training}
\label{sec:eval:speedup2}

In Figure~\ref{fig:stacked}, we see that \SB{} reduces total time to target error
rates compared to \Scan{} by reducing the time spent in the backward pass.
In Figure~\ref{fig:stacked-cifar10} and Figure~\ref{fig:stacked-cifar100}, both
\Scan{} and \SB{} take the same number of epochs to reach the target error rate. However,
\SB{} performs more overall forward passes. As described in Section~\ref{sec:implementation}, this
is because we perform one selection forward pass for each candidate example, plus one
training forward pass for each selected example. The ``Other'' bar shown for each
run includes per-run overheads (e.g., loading the dataset into memory and the network onto the GPU) and per-epoch overheads (e.g., evaluating test accuracy).

\begin{figure*}
\begin{minipage}{\textwidth}
        \centering
        \begin{subfigure}[b]{0.33\textwidth}
          \includegraphics[width=\linewidth]{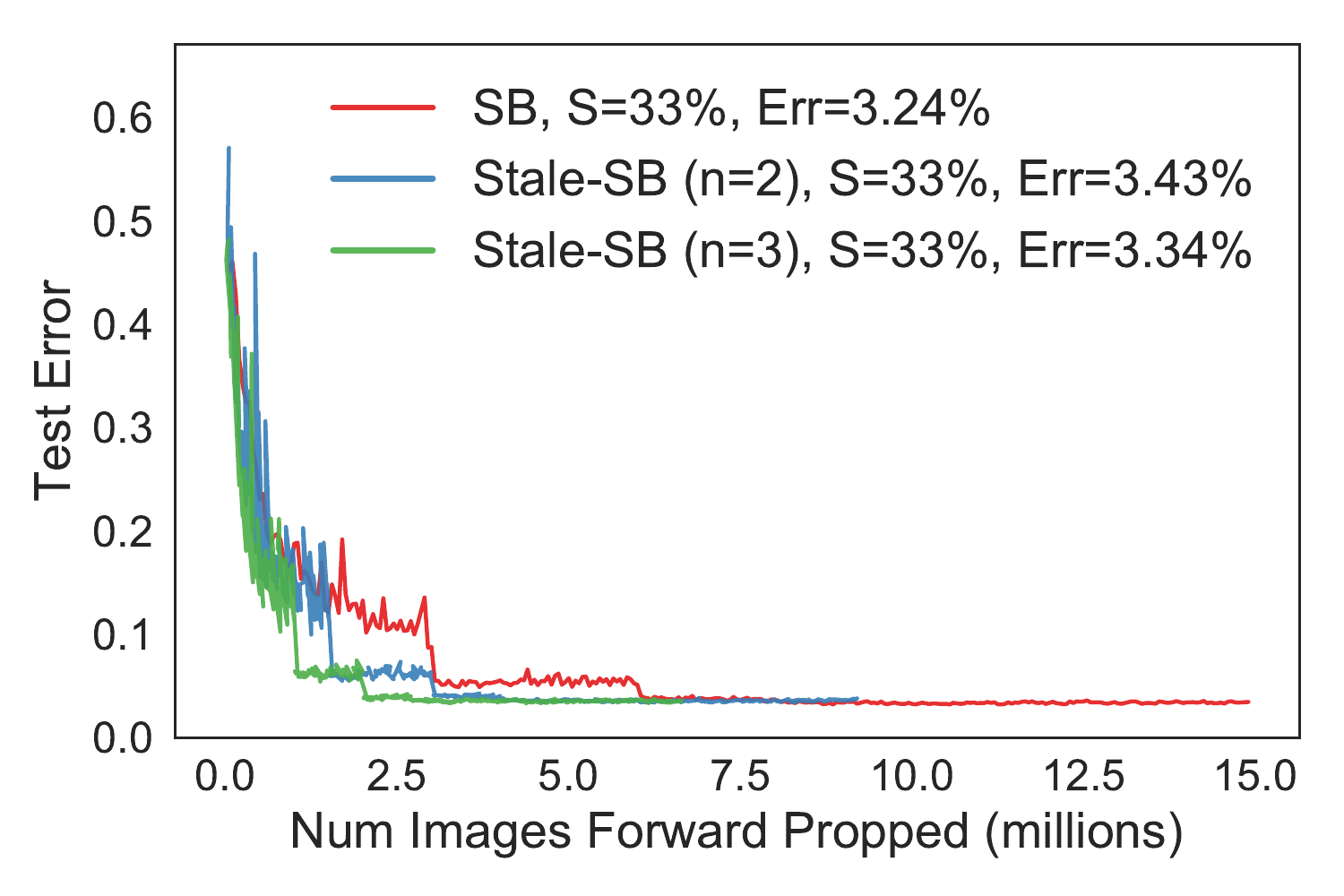}
          \caption{CIFAR10}
          \label{fig:staleness-cifar10}
        \end{subfigure}%
        \begin{subfigure}[b]{0.33\textwidth}
          \includegraphics[width=\linewidth]{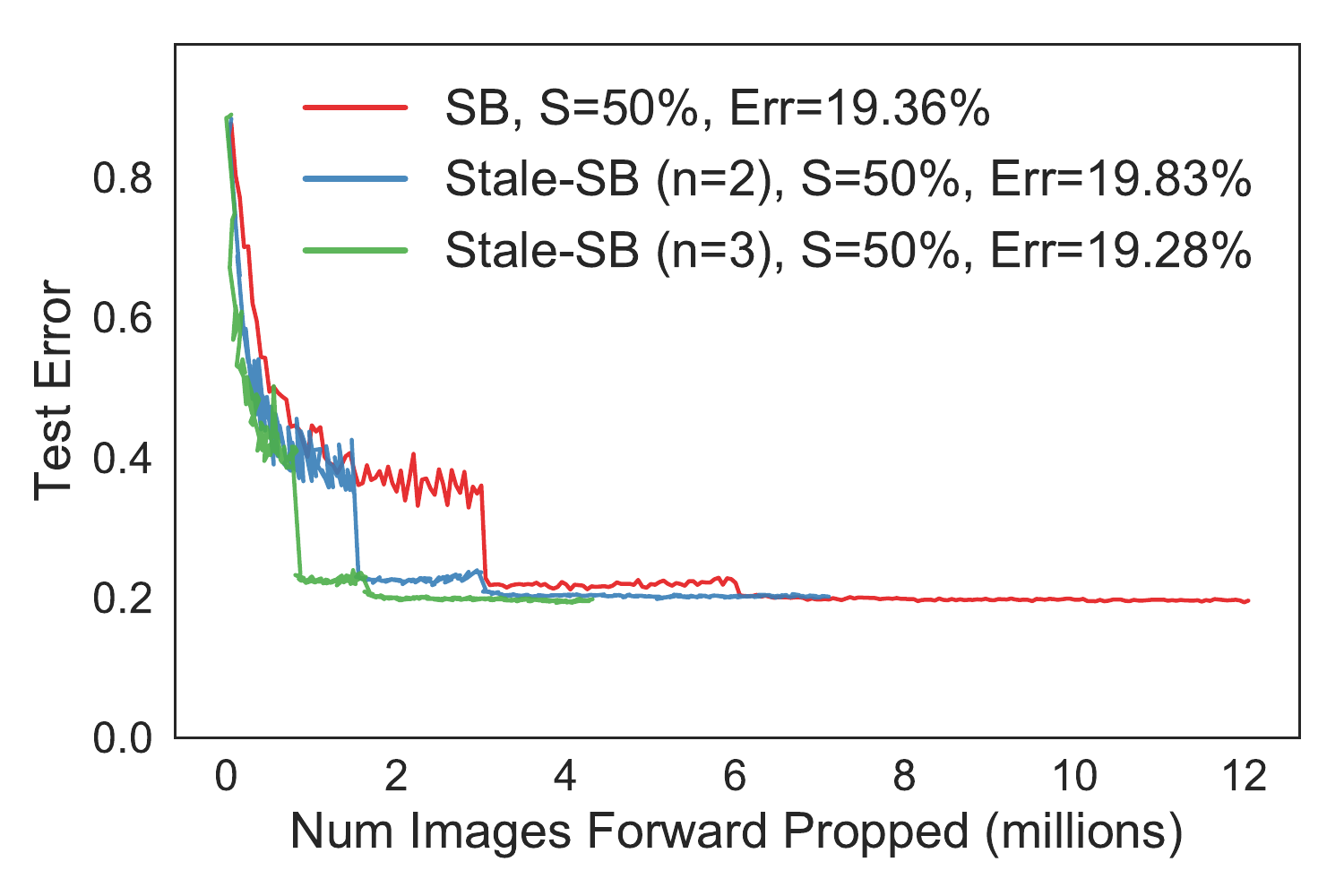}
          \caption{CIFAR100}
          \label{fig:staleness-cifar100}
        \end{subfigure}%
        \begin{subfigure}[b]{0.33\textwidth}
          \includegraphics[width=\linewidth]{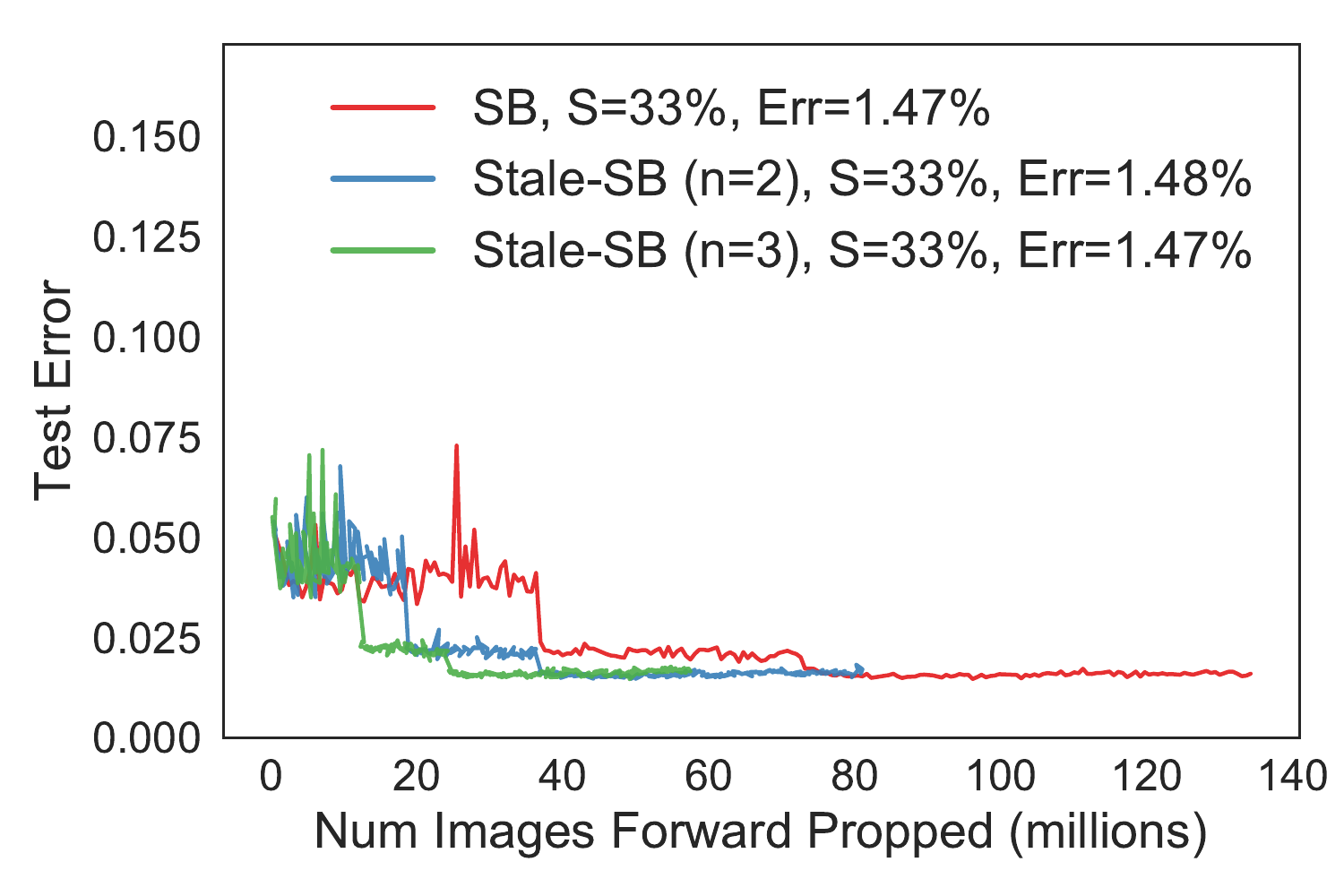}
          \caption{SVHN}
          \label{fig:staleness-svhn}
        \end{subfigure}%
        \caption{Increasing loss staleness reduces number of forward passes with little loss in accuracy.}
        \label{fig:staleness}
\end{minipage}
% \squeeze{}
% \squeeze{}
\end{figure*}

\textbf{Using stale losses to reduce selection passes.} After \SB{}'s reduction
of backward passes performed, over half of the remaining training time is
spent on forward passes. We evaluate \Stale{}, which uses the losses of forward
passes from previous epochs to perform selection. With \Stale{}, we run fewer
selection passes, running them only every $n=2$ or $n=3$ epochs.
That is, if $n=2$, an example that incurs a high loss has a high chance of
being trained on in the next two epochs, instead of just one.
In Figure~\ref{fig:stacked}, we see that \Stale{} with $n=3$ reduces the time spent performing
selection passes by two-thirds, thereby further reducing the total wall-clock time. 
In Figure~\ref{fig:staleness}, we see that the reduced number of total forward
passes in \Stale{} has little effect on final error. \Stale{}'s ability to reduce
selection passes while maintaining test accuracy leads to the end-to-end speedups shown in 
Table~\ref{table:speedup}. On average, \Stale{} with $n=3$ reaches target error rates
\StaleOverSB{} faster than \SB{}. With $n=3$, we believe \Stale{} captures most of the benefits of reducing
selection passes, though we have not yet experimented with values of $n>3$.

\begin{comment}
% putting this experience in design
\textbf{Running inference asynchronously.} In our experiments, we run \SB{} on a
single GPU where example selection is performed serially with training. For
each training iteration, we select examples for each backward pass by running an
average of three forward passes (when selectivity is 33\%). After the batch is
populated, we perform a forward and backward pass for training.
The trend towards ML accelerators that can do forward passes at low cost~\cite{jouppi17}
is especially beneficial to \SB{} since the dominant cost in the selection process is
due to a forward pass. One could thus run the selection process on an inference
accelerator in parallel with the backward passes on a GPU. The primary overhead associated
with pipelining \SB{} is syncing the trained model between devices. Our preliminary results
suggest that syncing the model one time per epoch reduces training accuracy by \textbf{XXXXX\%},
\textbf{XXXXX\%}, and \textbf{XXXXX\%} for CIFAR10, CIFAR100 and SVHN, respectively.
\end{comment}

\subsection{\System{} sensitivity analysis}
\label{sec:eval:sweep}

\textbf{\SB{} is robust to modest amounts of label error.} One potential
downside of \SB{} is that it could increase susceptibility to noisy labels.
However, we show that on SVHN, a dataset known to include label
error~\cite{papernot18}, \SB{} still converges faster than \Scan{} to almost
all target error rates. We also evaluate \SB{} on CIFAR10 with manually
corrupted labels. Following the UniformFlip approach in \cite{ren18}, we
randomly flip 1\% (500 examples), 10\% (5000 examples) and 20\% (10000 examples).
%of CIFAR10.

\begin{figure*}
\begin{minipage}{\textwidth}
        %\begin{subfigure}[b]{0.33\textwidth}
        %  \includegraphics[width=\linewidth]{{figs/sysml20/labelerror/label-error-0.001}.pdf}
        %  \caption{Test error, 0.1\% shuffled}
        %  \label{fig:label-error-1}
        %\end{subfigure}%
        \begin{subfigure}[b]{0.33\textwidth}
          \includegraphics[width=\linewidth]{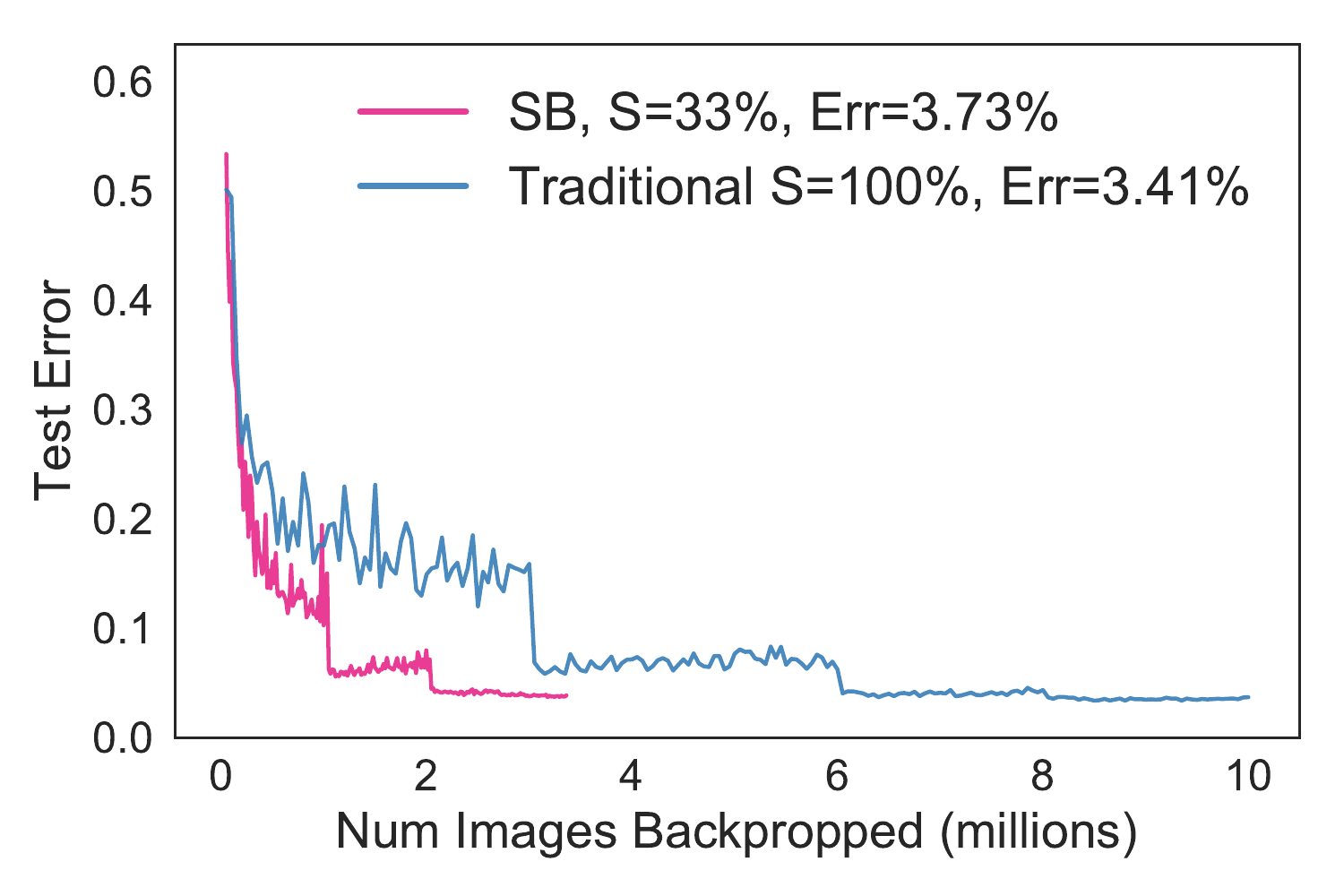}
          \caption{Test error, 1\% shuffled}
          \label{fig:label-error-1}
        \end{subfigure}%
        \begin{subfigure}[b]{0.33\textwidth}
          \includegraphics[width=\linewidth]{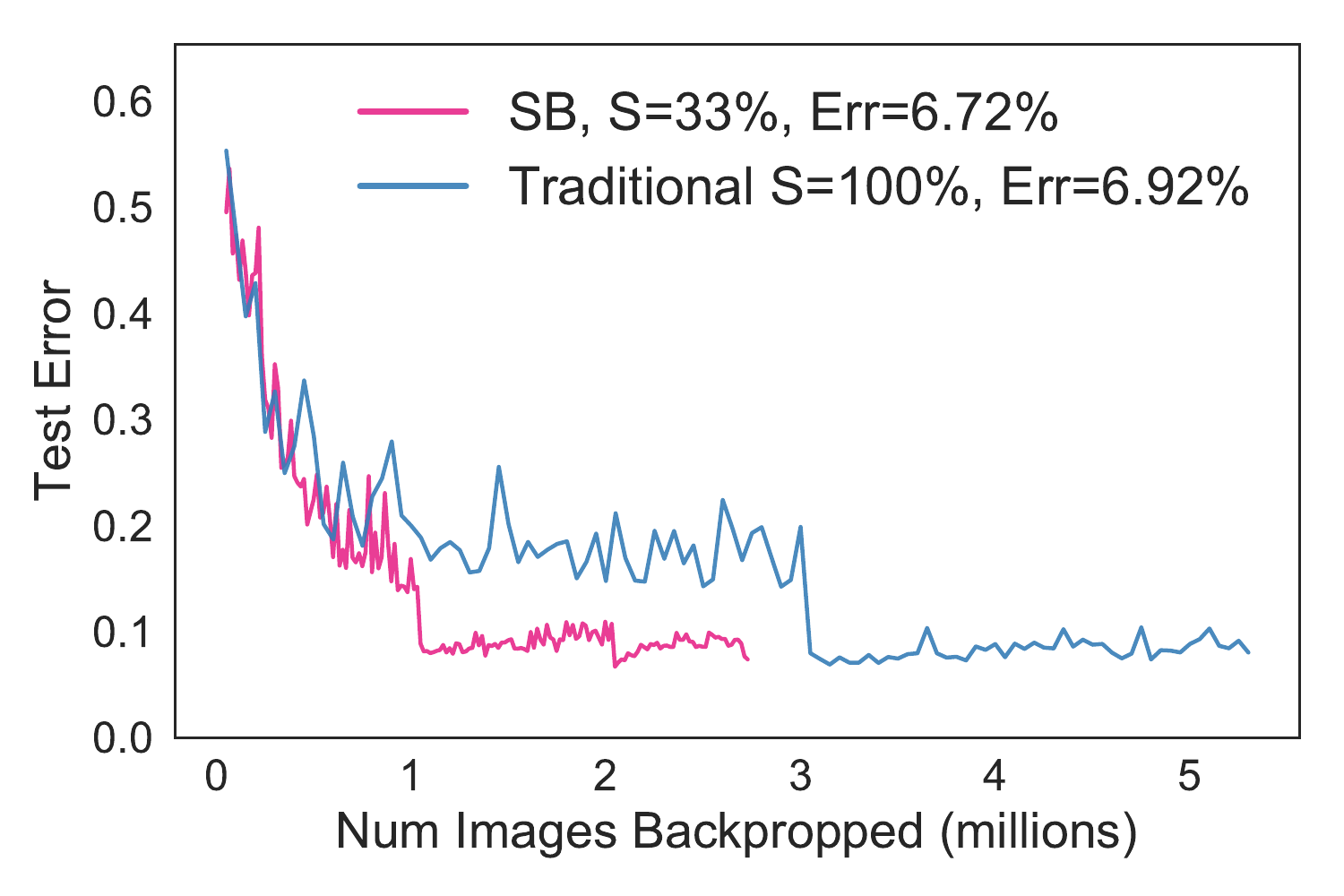}
          \caption{Test error, 10\% shuffled}
          \label{fig:label-error-2}
        \end{subfigure}%
        \begin{subfigure}[b]{0.33\textwidth}
          \includegraphics[width=\linewidth]{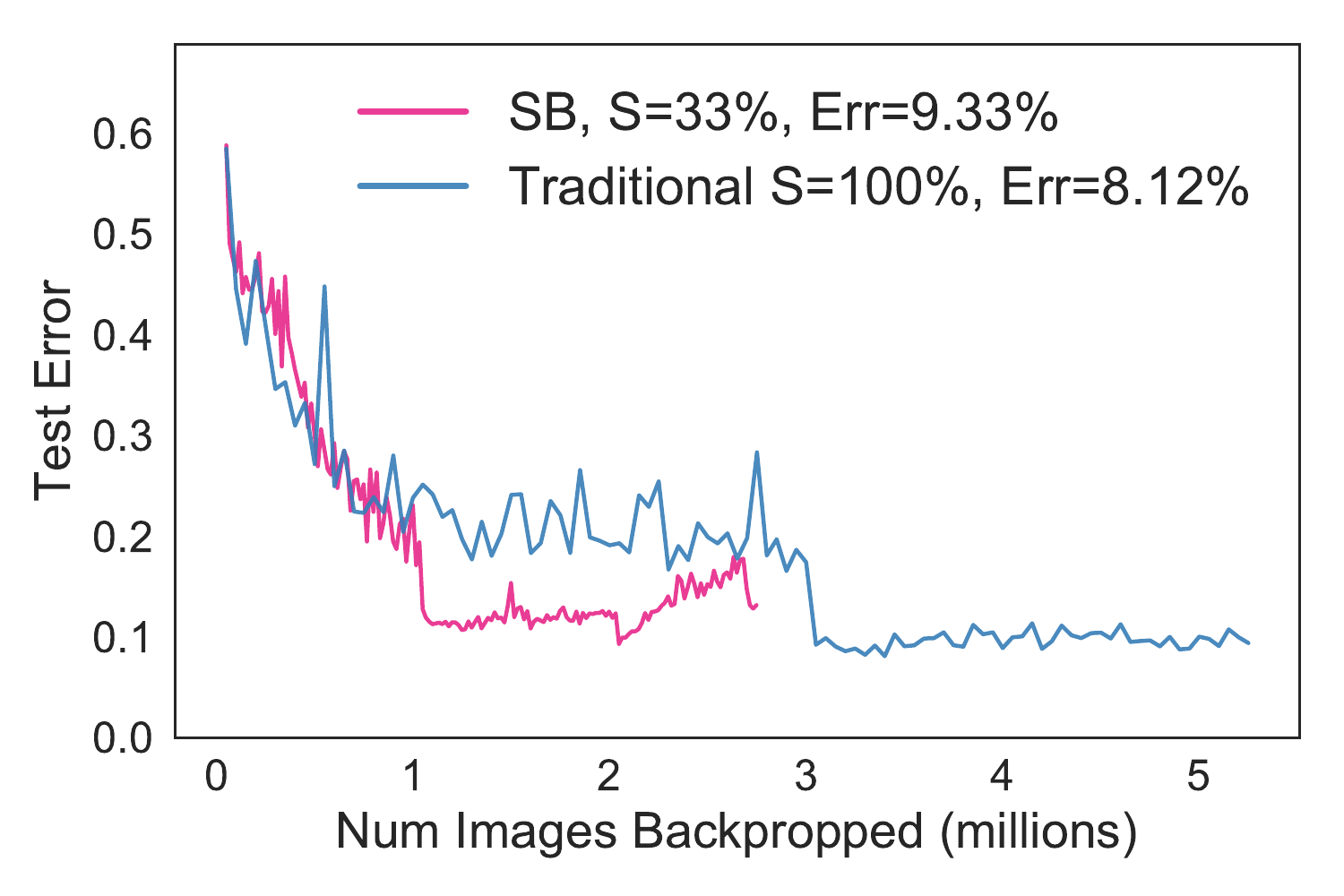}
          \caption{Test error, 20\% shuffled}
          \label{fig:label-error-3}
        \end{subfigure}%
        \caption{\SB{} reaches
        similar test error rates compared to \Scan{} with 1\% and 10\% shuffled labels.}
        \label{fig:cifar10-labelerror}
\end{minipage}
\end{figure*}

Fig~\ref{fig:cifar10-labelerror} shows that \SB{} accelerates
training for all three settings.  With 1\% and 10\% of examples corrupted,
\SB{} reaches a comparable final test accuracy. 
%However, w
With 20\% corruption,
\SB{} overfits to the incorrect labels and increases the final test error. So, while
\SB{} is robust to modest amounts of label error, it is most effective on
relatively clean, validated datasets.

\begin{figure*}
\begin{minipage}{\textwidth}
        \centering
        \begin{subfigure}[b]{0.33\textwidth}
          \includegraphics[width=\linewidth]{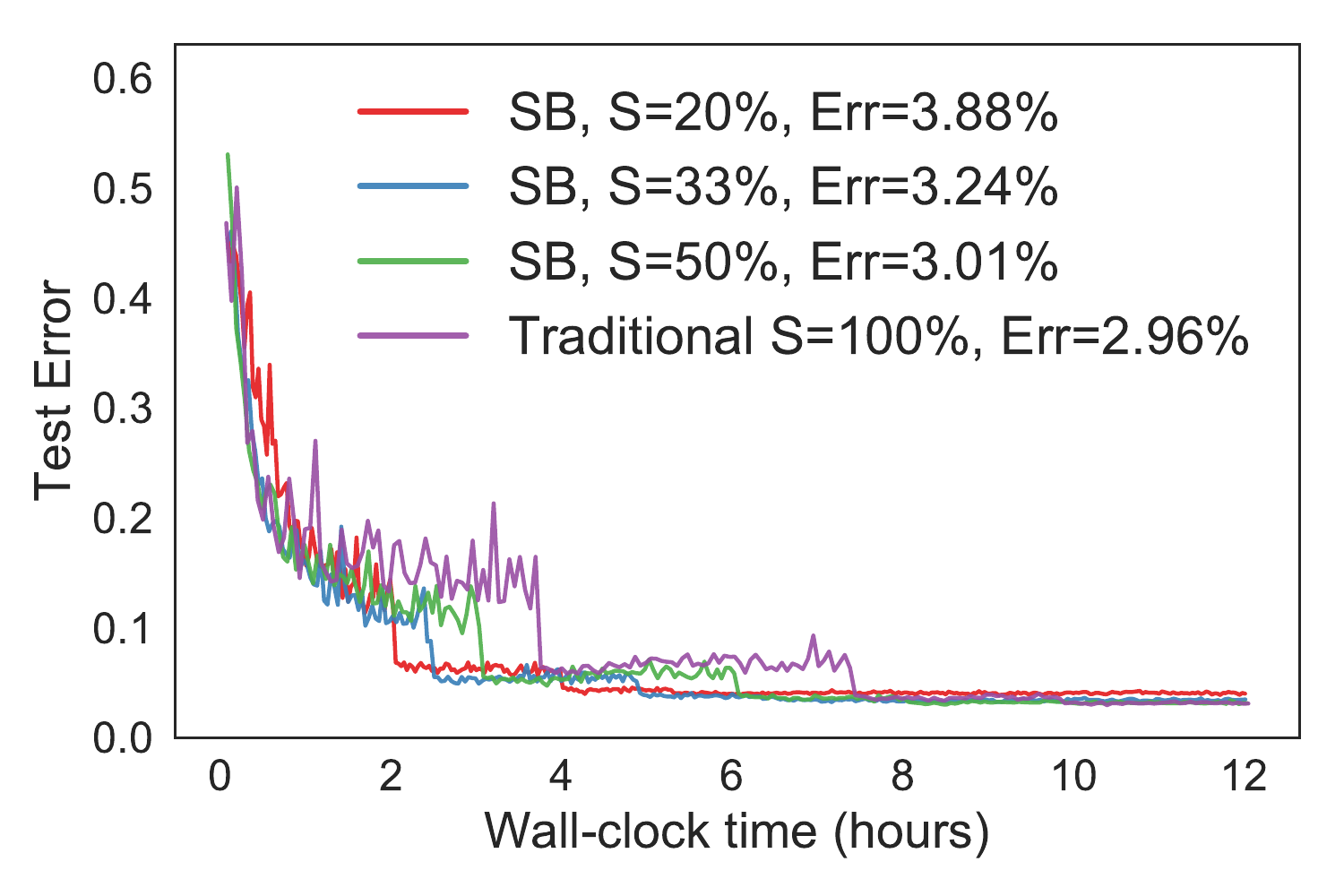}
          \caption{CIFAR10}
          \label{fig:selectivity-cifar10}
        \end{subfigure}%
        \begin{subfigure}[b]{0.33\textwidth}
          \includegraphics[width=\linewidth]{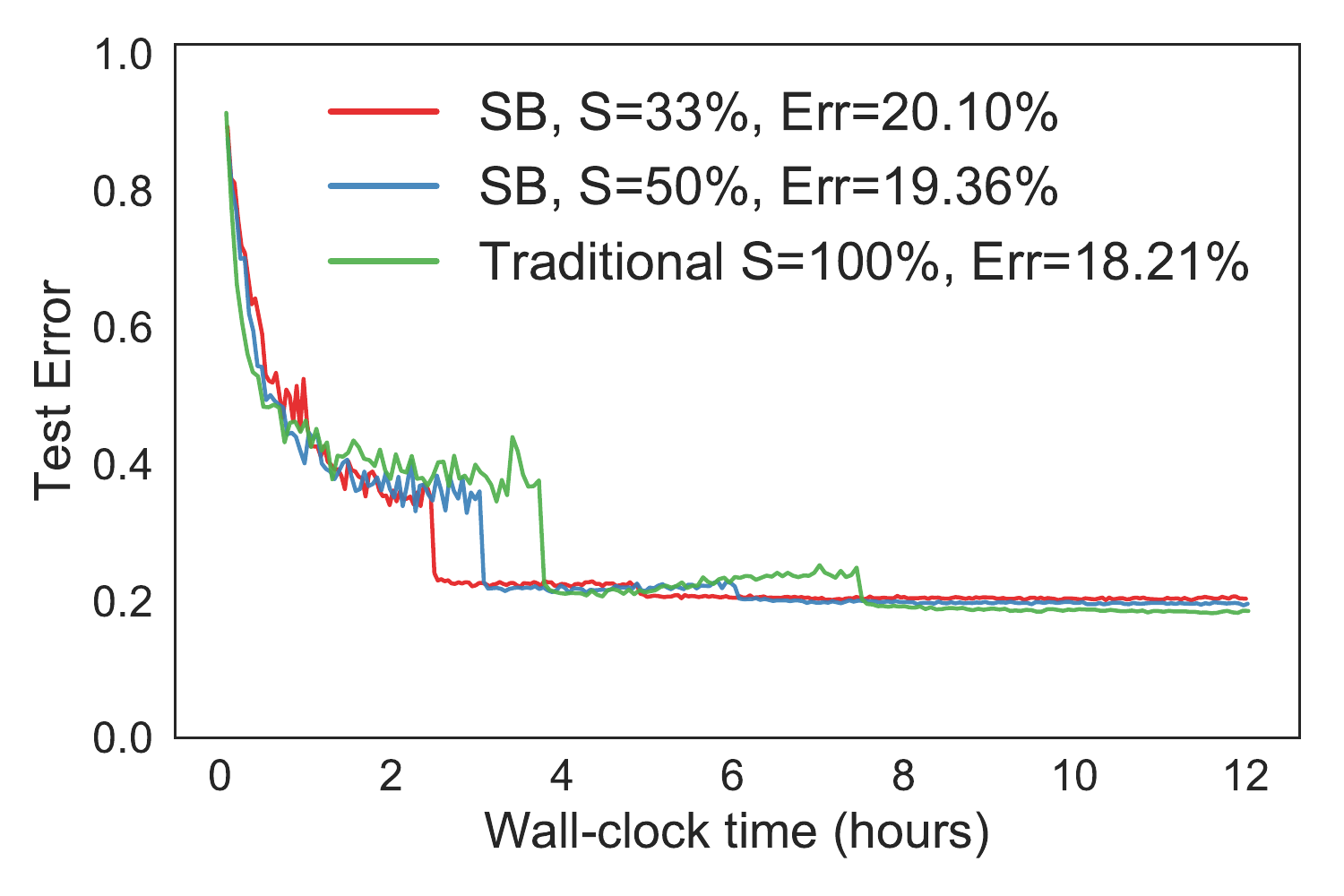}
          \caption{CIFAR100}
          \label{fig:selectivity-cifar100}
        \end{subfigure}%
        \begin{subfigure}[b]{0.33\textwidth}
          \includegraphics[width=\linewidth]{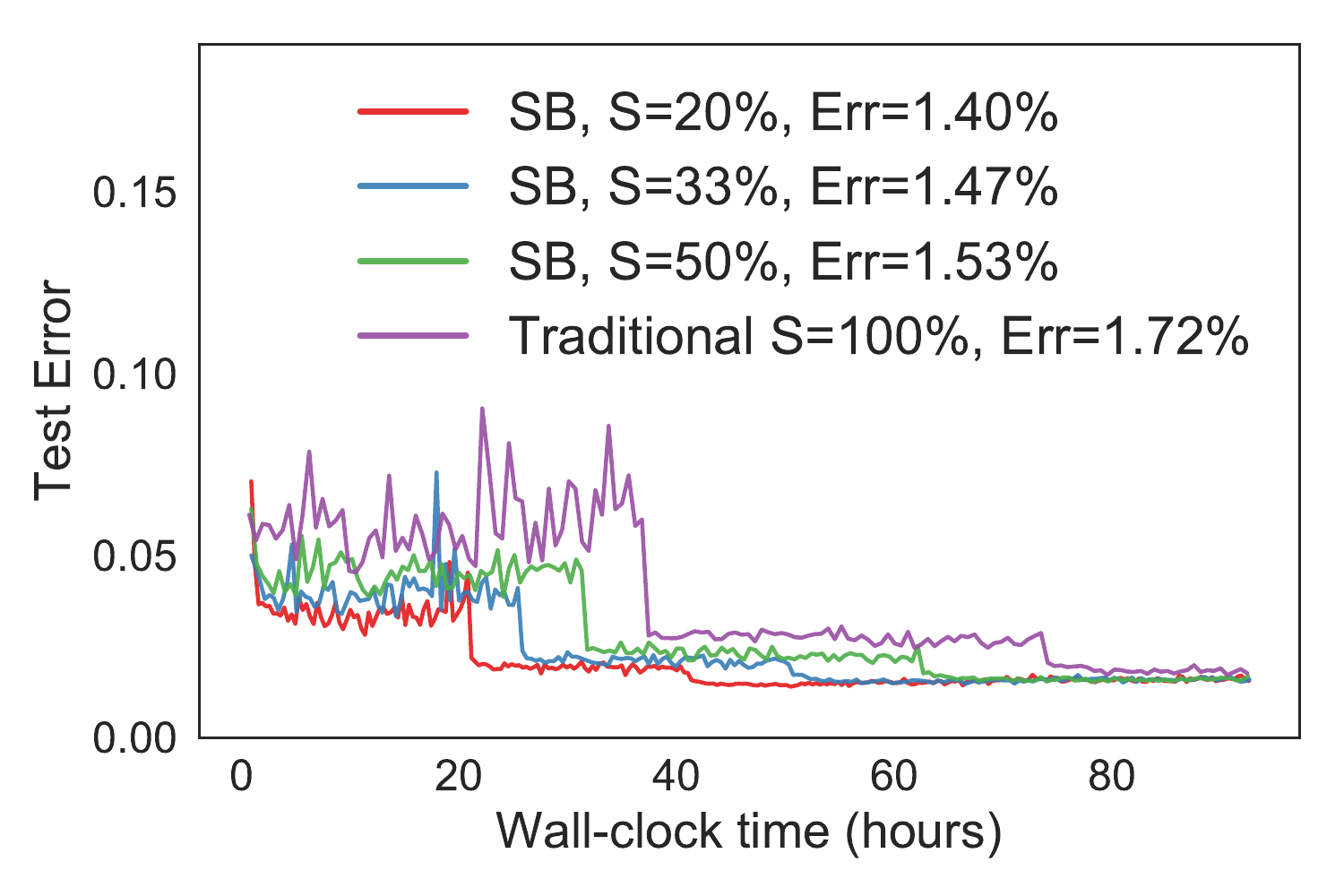}
          \caption{SVHN}
          \label{fig:selectivity-svhn}
        \end{subfigure}%
        \caption{\SB{} accelerates training for a range of selectivities. Higher selectivity gives faster training but can increase error.}
        \label{fig:selectivity}
\end{minipage}
% \squeeze{}
% \squeeze{}
\end{figure*}

\textbf{Higher selectivity accelerates training, but increases final
error.} Tuning $\beta$ in Equation~\ref{eq:prob} changes \SB{}'s selectivity.
In Figure~\ref{fig:selectivity}, we see that increasing \SB{}'s selectivity,
focusing more on harder examples, increases the speed of learning but
can cause result in higher final error. For CIFAR10, \SB{} reaches within 0.92\% of
\Scan{}'s final error rate with 20\% selectivity. For CIFAR100, it
reaches within 2.54\% of the final error rate with 25\% selectivity.
As with other hyperparameters, the best selectivity depends on the target
error and dataset. 
Overall, we observe that \SB{} speeds up training
with a range (20--65\%) for selectivity.

\textbf{\SB{} using additional learning rate schedules.} We train \SB{} using the
provided learning rate schedule in ~\cite{devries2017} which reproduces state-of-the-art
accuracies on CIFAR10, CIFAR100, and SVHN for Wide Resnet with Cutout. We also train using
a static learning rate schedule to adjust for confounding factors, as well as an accelerated
learning rate schedule. In both cases, we see the same trends as with the initial learning
rate. We include the configurations in Section~\ref{sec:eval:pareto} and the training curves
in the appendix.

\subsection{Putting it all together}
\label{sec:eval:pareto}

\begin{figure*}
\begin{minipage}{\textwidth}
        \centering
        \begin{subfigure}[b]{0.33\textwidth}
          \includegraphics[width=\linewidth]{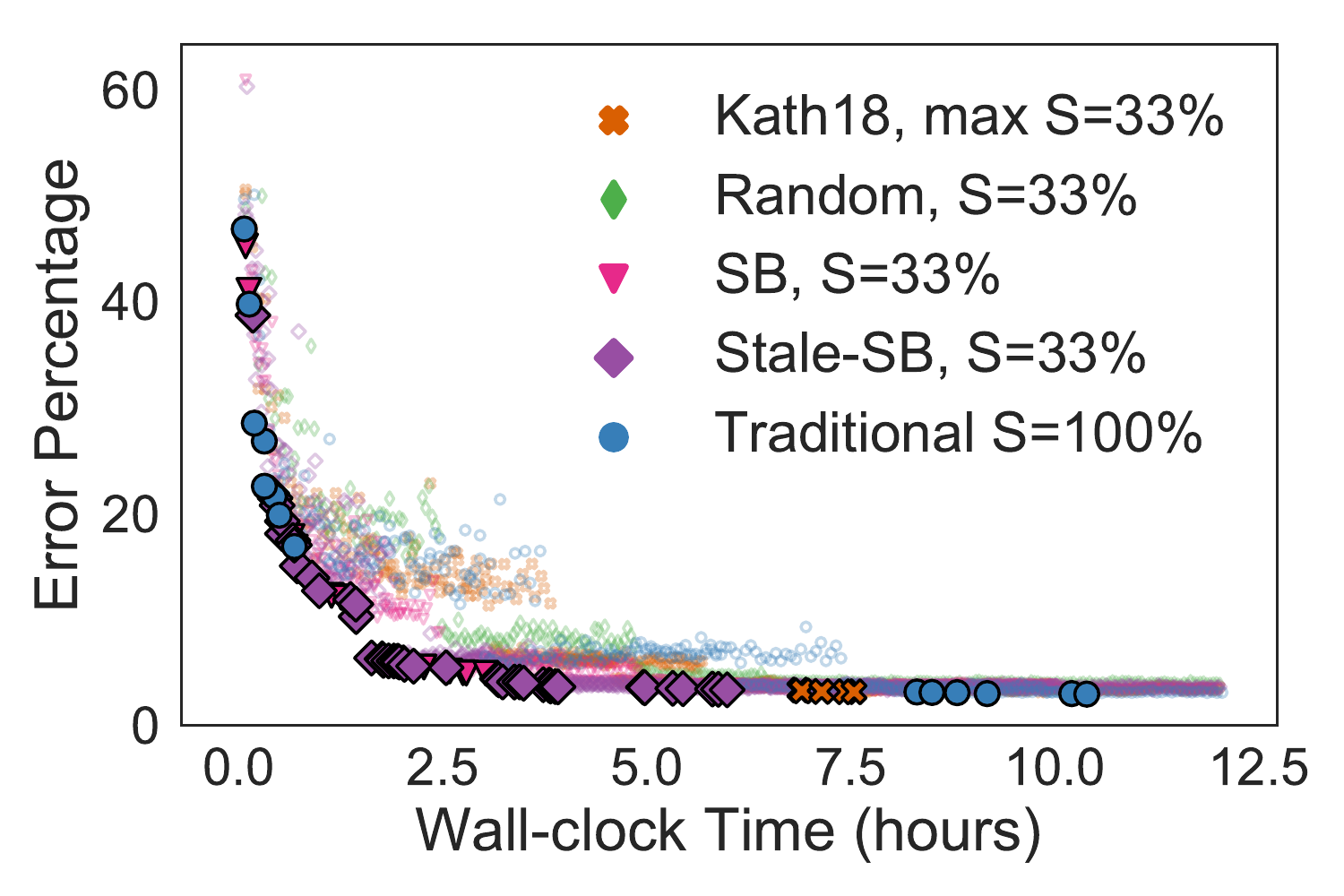}
          \caption{CIFAR10}
          \label{fig:pareto-cifar10}
        \end{subfigure}%
        \begin{subfigure}[b]{0.33\textwidth}
          \includegraphics[width=\linewidth]{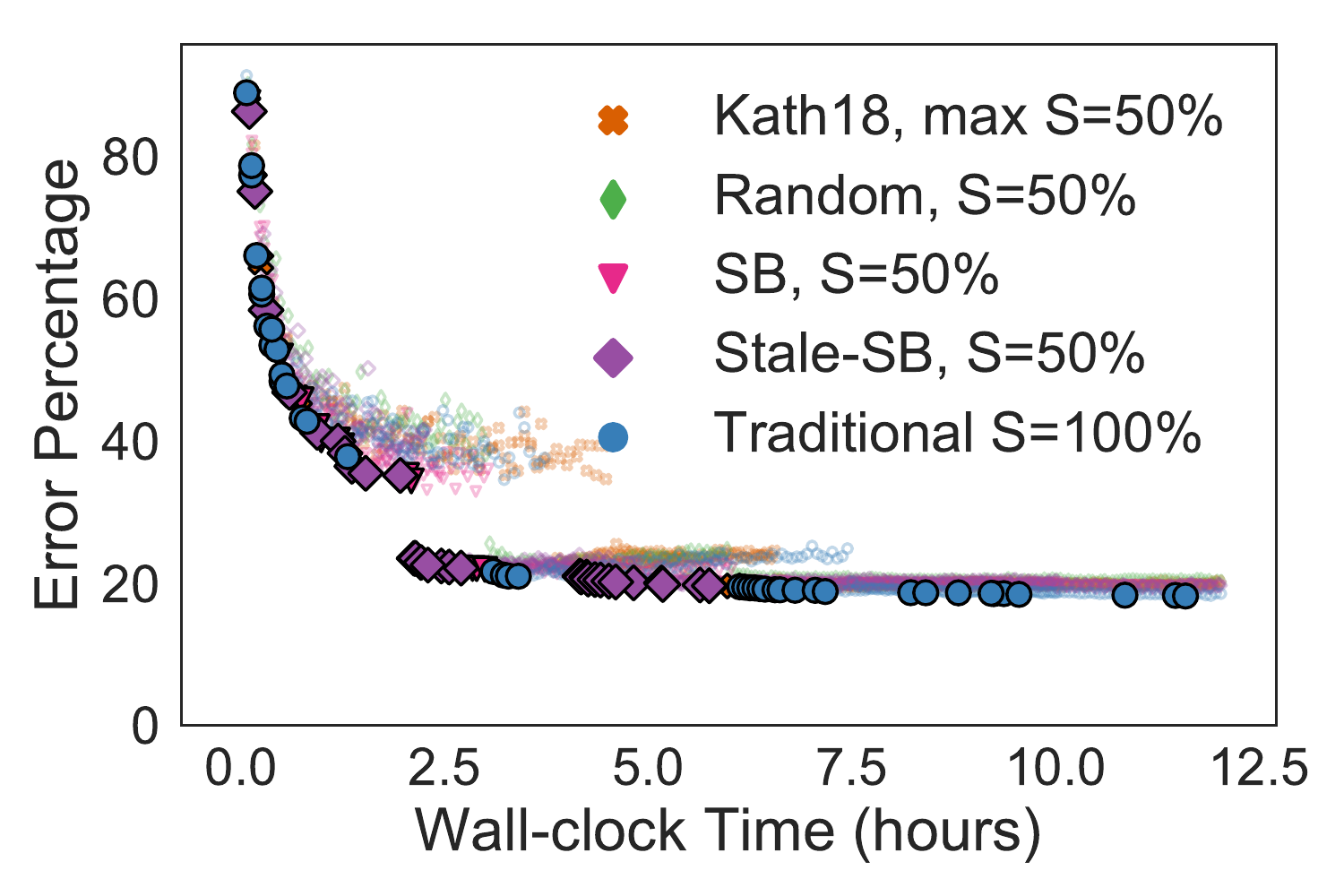}
          \caption{CIFAR100}
          \label{fig:pareto-cifar100}
        \end{subfigure}%
        \begin{subfigure}[b]{0.33\textwidth}
          \includegraphics[width=\linewidth]{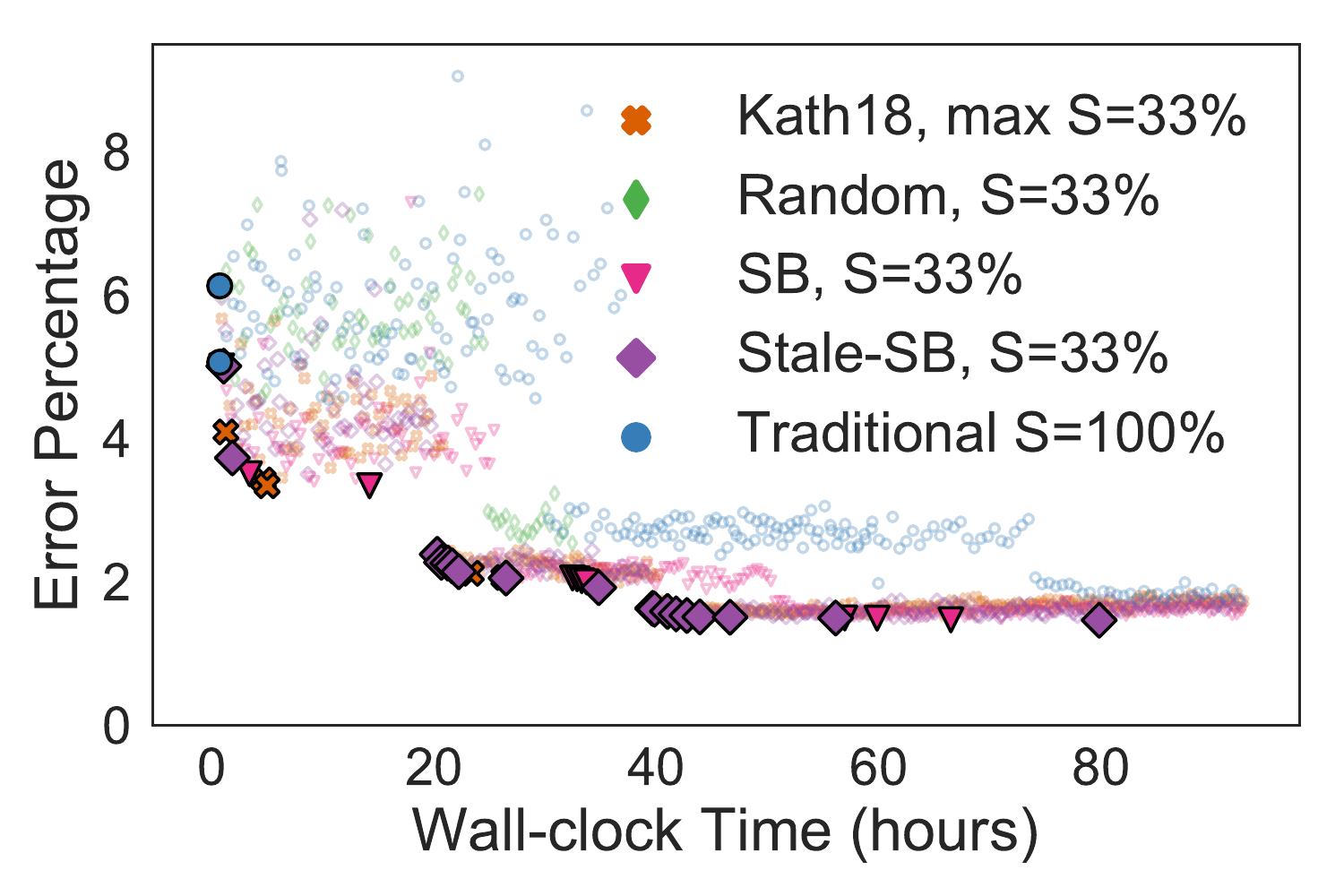}
          \caption{SVHN.}
          \label{fig:pareto-svhn}
        \end{subfigure}%
        \caption{Pareto-optimal points for training time vs error trade-off are opaque and filled. \SB{} and \Stale{} offer the majority of Pareto-optimal options for trading off training time and accuracy.}
        \label{fig:pareto}
\end{minipage}
\end{figure*}

The optimal training setup to reach a certain target error rate depends 
on a variety of factors. In the previous sections, we compared \Scan{},
\SB{}, \Stale{}, and \Kath{} using a variety of different configurations.
In Figure~\ref{fig:pareto}, we plot the wall-clock
time needed to reach a range of target error rates for all four strategies, each
trained with two learning rate schedules and run with different selectivities.
A small subset of configurations make up the Pareto frontier, which represent
the best strategy for a given target error rate. Points on the Pareto frontier
are colored in bold whereas suboptimal points are shown with transparency. 

\textbf{\SB{} provides the majority of Pareto-optimal configurations.}
As an approximate signal for robustness of our strategy, we calculate the
fraction of Pareto points provided by \SB{} and \Stale{}, \Kath{} and \Scan{}. 
For a majority of training time budgets, \SB{} gives the lowest error
rates. For CIFAR10, CIFAR100, and SVHN, \SB{} and its optimized variant \Stale{}
account for 72\%, 47\% and 80\% of the Pareto-optimal choices, respectively. The
exception is cases with very large training time budgets, where \Scan{} reaches lower
final error rates than \SB{}. Overall, \Scan{} accounts for 10\%, 43\% and 6\% of Pareto points
in CIFAR10, CIFAR100 and SVHN, respectively. As shown in Table~\ref{table:speedup},
\SB{} is also faster than \Kath{}, a state-of-the-art importance sampling technique for 
speeding up training, while achieving the same final error rate. \Kath{} provides
10\%, 8\% and 14\% of Pareto-optimal points.

% !TeX root = ./selective_backprop.tex

%\section{Practicality}
%\label{sec:practicality}

\textbf{Practicality.} \System{} reduces training iterations and wall-clock time
needed to achieve a target error rate with little programmer effort.
We evaluated \System{} with a diverse set of network architectures and datasets.
In each case, we did not retune initial hyperparameters from canonical setups.
Most of these training setups included traditional accuracy-boosting techniques,
including data augmentation, cutout, dropout, and batch normalization.  \System{} still
improved training atop these existing optimizations. \System{} is also mathematically
lightweight and simple to add to code.

\begin{comment}
\squeeze{}
{\small
\begin{verbatim}
  sb = SB(cnn,
          cnn_optimizer,
          batch_size,
          num_classes)
  for epoch in range(args.epochs):
    sb.trainer.train(train_loader)
\end{verbatim}
}
\squeeze{}
\end{comment}

% !TeX root = ./selective_backprop.tex

\section{Conclusion}

\System{} accelerates training of neural networks. The key idea is to skip the
backward pass for training examples for which the forward-pass loss function
indicates little value. \System{} lets the current state of the network dictate
selectivity for each example, and is lightweight, scalable, and effective.
Experiments on several datasets and networks show that \System{} converges to
target error rates up to \MaxSpeedup{} faster than with standard SGD and
\SBOverKath{} faster than the state-of-the-art sampling approach introduced in
\cite{katharopoulos18}. Determining selectivity with stale loss information
further accelerates training by \StaleOverSB{}.  \System{} is also simple to
add to code. An open source implementations of \System{} is available at \sourceURL{}.

\clearpage

\bibliography{selective_backprop}
\bibliographystyle{sysml2019}

\onecolumn
\appendix
\section{Supplementary Materials}

\subsection{\System{} code}

Our implementation of \System{} can be found at \sourceURL{}.  The specific
training set up used in this paper is available at
\url{http://bit.ly/SBCutoutAnon}, which we modified to use our implementation
as a submodule.

\subsection{Variance of relative losses over time}

% \citeauthor{hinton07} \cite{hinton07}
\citet{hinton07} warns that when sampling, easy examples are likely
to develop large gradients while being ignored. 
We show this effect on CIFAR10. % in the supplementary material. %in Fig.~\ref{fig:cifar10-forgetting}: 
In typical training, the probability of selection has a natural variance, 
but increases once sampling is introduced. One benefit of using a look-once approach for
determining example importance is that \System{}
always uses the up-to-date state of the network, 
which is in constant flux during training.

\begin{figure*}[h]
%\squeeze{}
\hspace*{\fill}%
\begin{subfigure}[t]{0.4\linewidth}
% \begin{minipage}[t]{0.33\linewidth}
    \includegraphics[width=\linewidth]{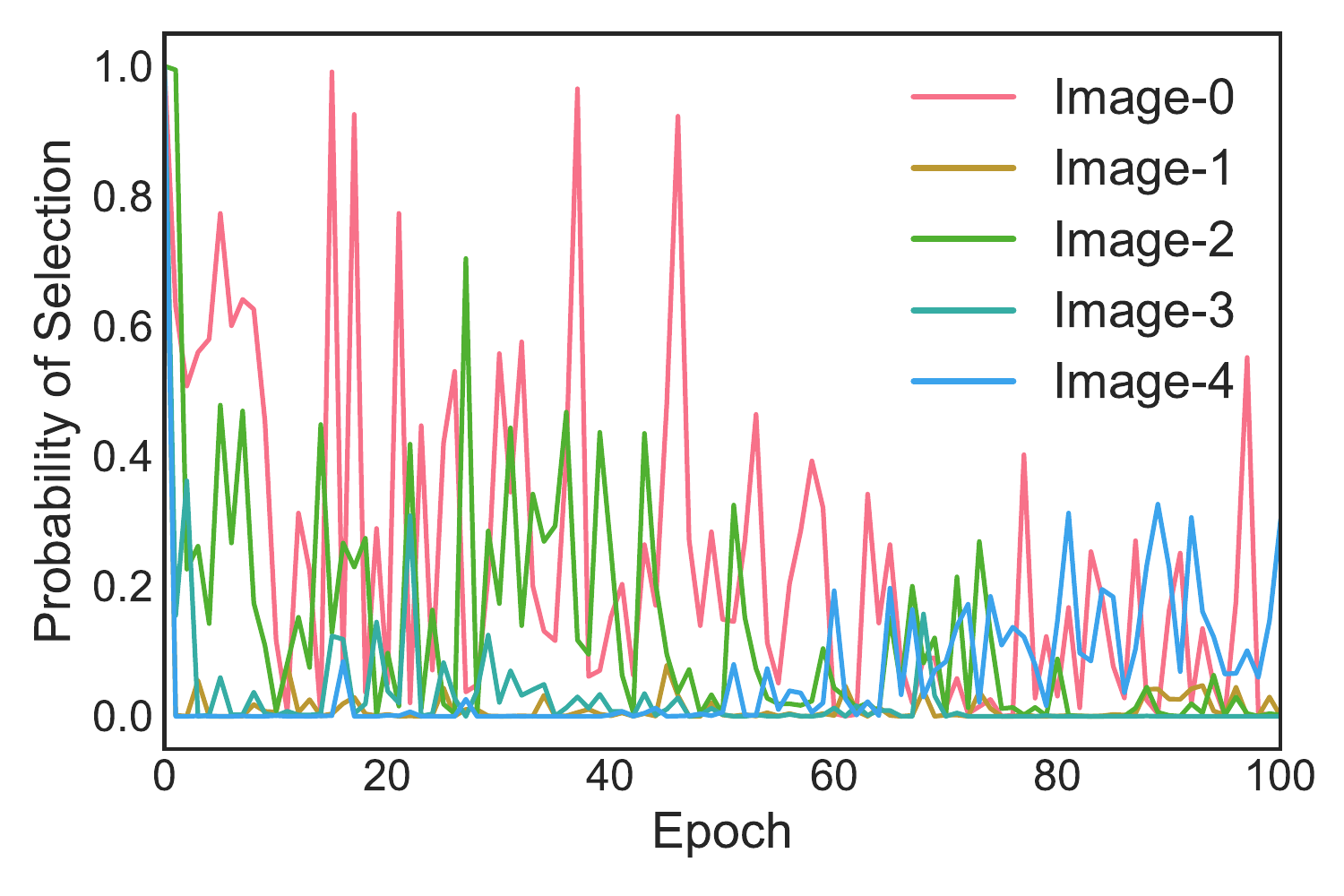}
    \vspace{-2em}
    \caption{Without \SB{} sampling.}
    \label{fig:forgetting-baseline}
\end{subfigure}%
% \end{minipage}%
    \hfill%
\begin{subfigure}[t]{0.4\linewidth}
% \begin{minipage}[t]{0.33\linewidth}
    \includegraphics[width=\linewidth]{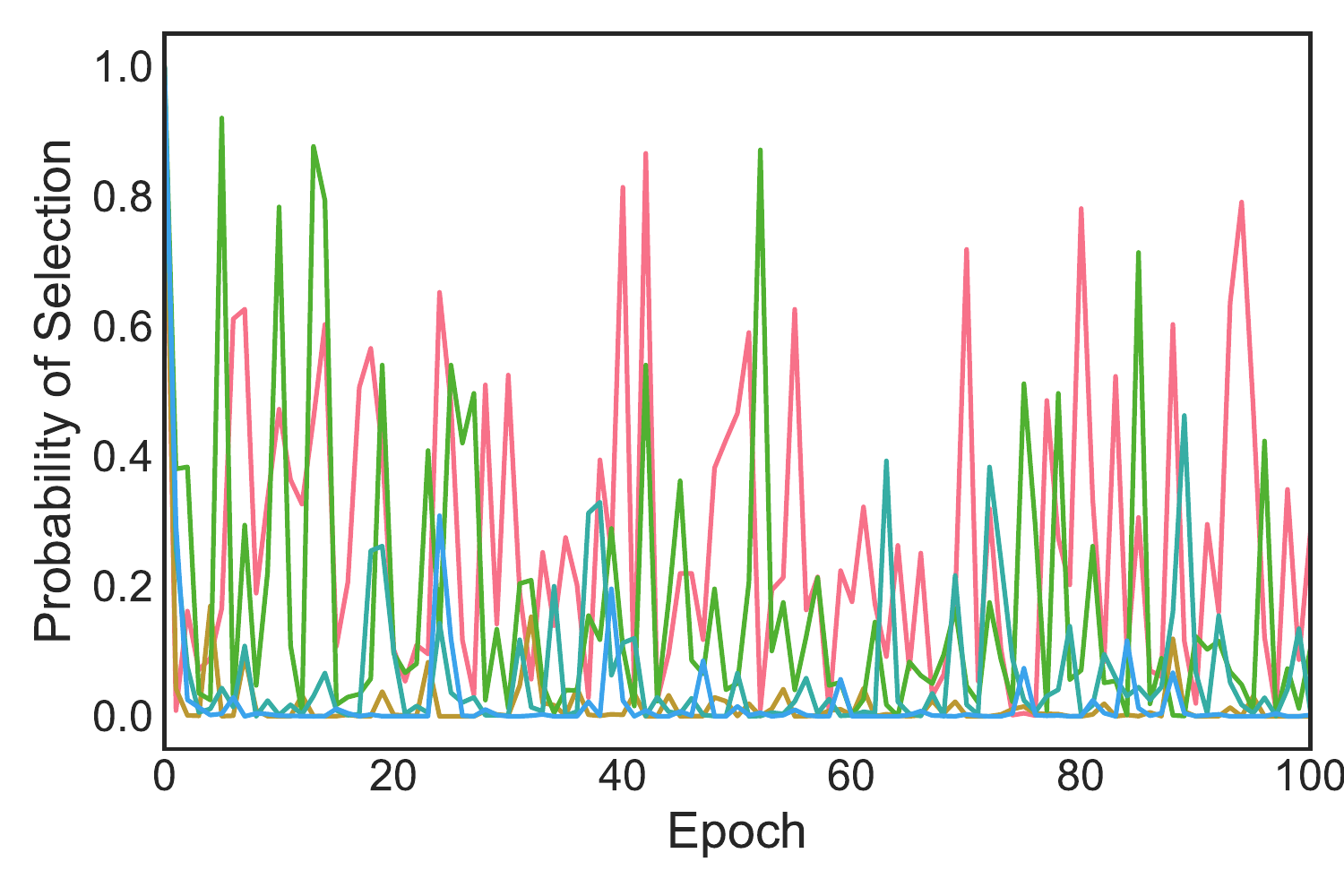}
    \vspace{-2em}
    \caption{With \SB{} sampling.}
    \label{fig:forgetting-sb}
\end{subfigure}%
% \end{minipage}
\hspace*{\fill}%
\caption{Select probabilities of five examples when training
        MobilenetV2 on CIFAR10. Each line represents one image. Likelihood of
      selection fluctuates more when sampling is introduced.}
\label{fig:cifar10-forgetting}
%\squeeze{}
\end{figure*}

\subsection{Sensitivity analyses}

\textbf{Network architecture.} We train CIFAR10 using three architectures,
ResNet18, DenseNet, and MobileNetV2 \cite{he16, huang17, sandler18} We train
CIFAR10 using the training setup from pytorch-cifar~\cite{pytorch-cifar}. The
learning rate is set at 0.1.

\begin{figure*}[h]
%\squeeze{}
\hspace*{\fill}%
\begin{subfigure}[t]{0.33\linewidth}
    \includegraphics[width=\linewidth]{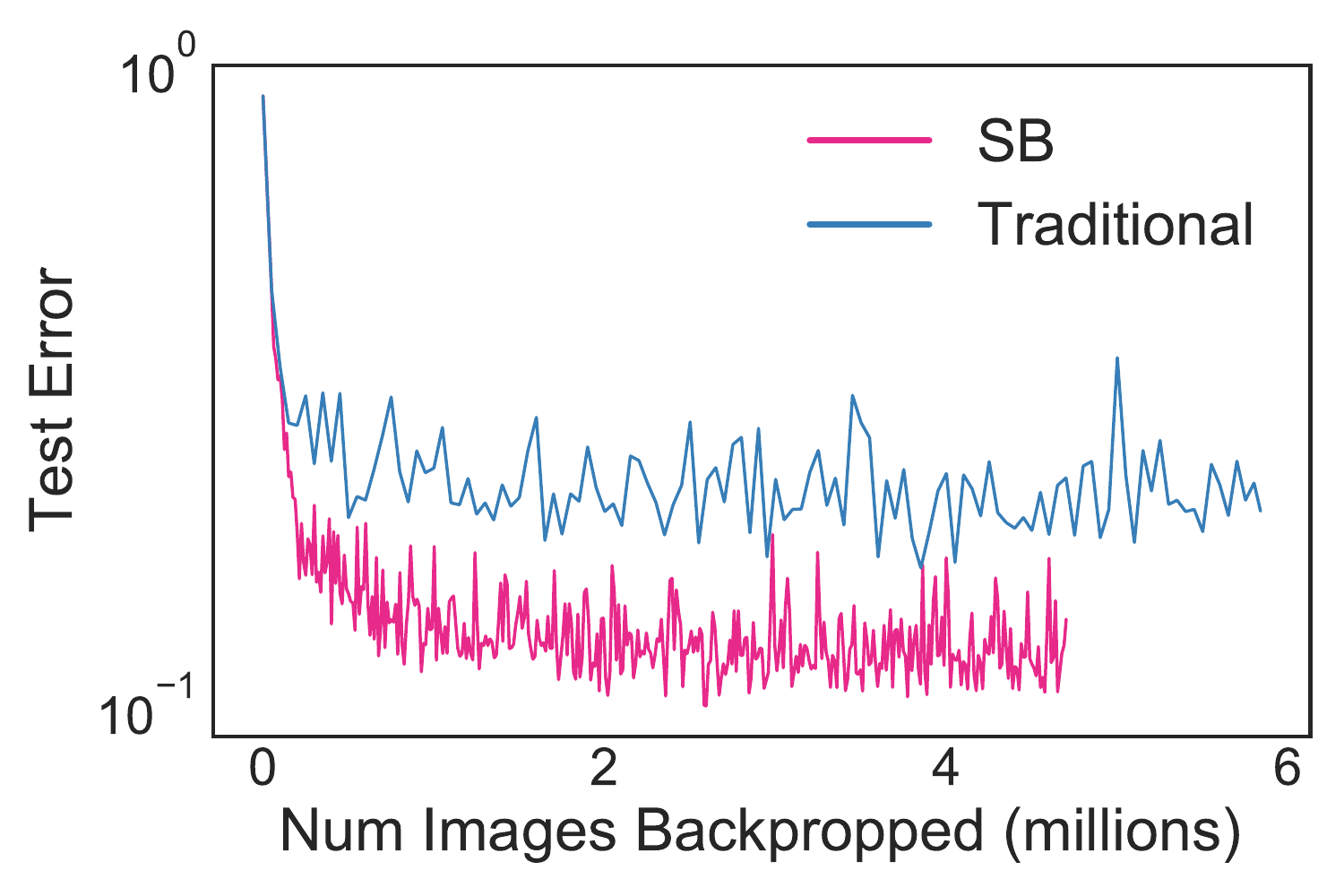}
    \vspace{-2em}
    \caption{DenseNet.}
    \label{fig:cifar10-densenet}
\end{subfigure}%
    \hfill%
\begin{subfigure}[t]{0.33\linewidth}
    \includegraphics[width=\linewidth]{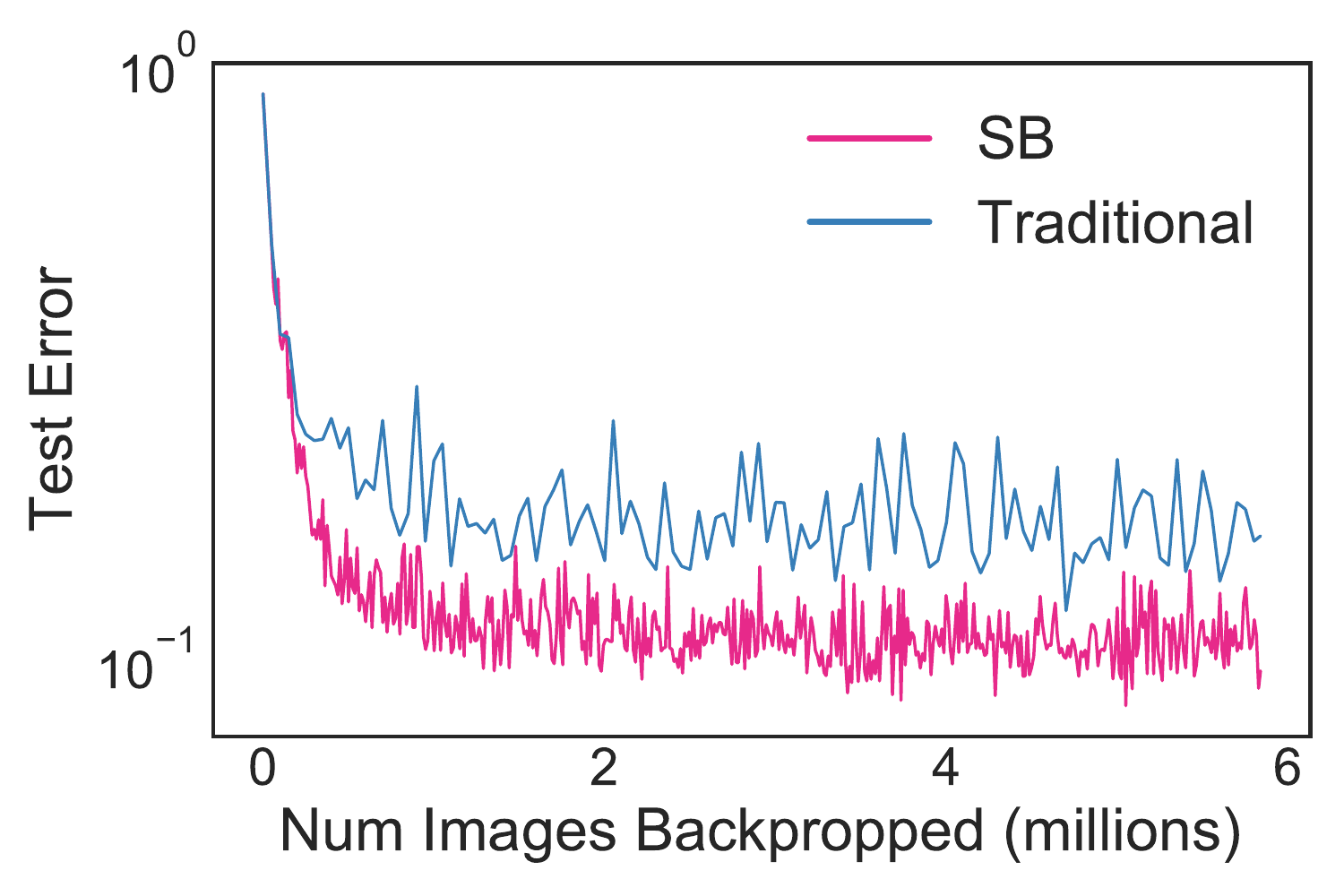}
    \vspace{-2em}
    \caption{ResNet18.}
    \label{fig:cifar10-resnet}
\end{subfigure}%
\hspace*{\fill}%
\begin{subfigure}[t]{0.33\linewidth}
    \includegraphics[width=\linewidth]{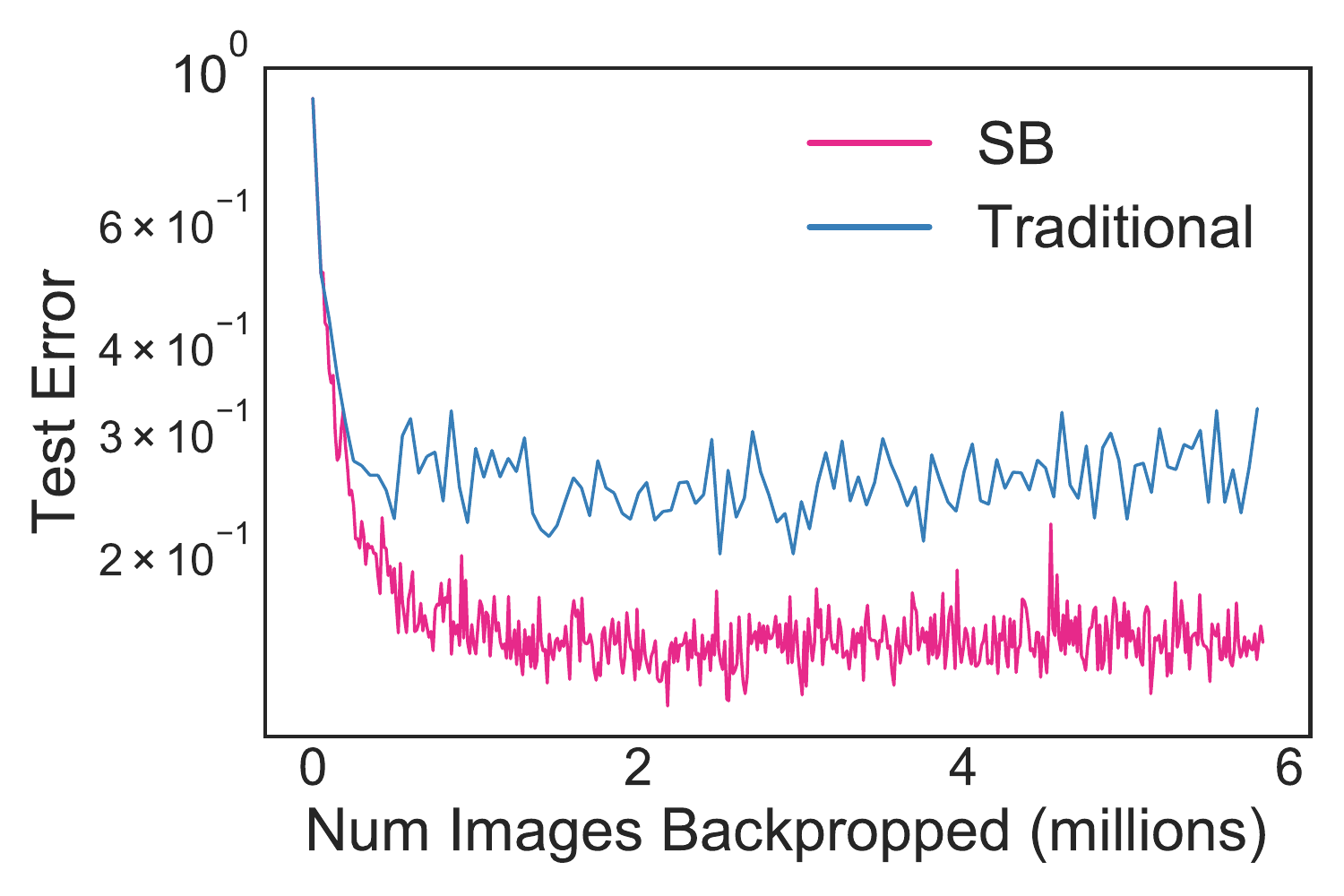}
    \vspace{-2em}
    \caption{MobilenetV2.}
    \label{fig:cifar10-mobilenetv2}
\end{subfigure}%
\hspace*{\fill}%
\caption{Training on CIFAR10 using three different network architectures.}
\label{fig:cifar10-nets}
%\squeeze{}
\end{figure*}

\textbf{Additional learning rate schedule.} We train CIFAR10, CIFAR100 and SVHN
on an additional, accelerated learning rate schedule and observe the same
trends as in Section~\ref{sec:eval}.  For CIFAR10, we start with $lr=0.1$ and
decay by 5x at 48, 96, and 128 epochs.  For CIFAR100, we start with $lr=0.1$
and decay by 10x at 48 and 96 epochs. For SVHN we initialize the learning rate
to 0.1 and decay by 10x at epochs 60 and 120.

\label{appendeix:lr2}

\begin{figure*}[h]
\begin{minipage}{\textwidth}
        \centering
        \begin{subfigure}[b]{0.33\textwidth}
          \includegraphics[width=\linewidth]{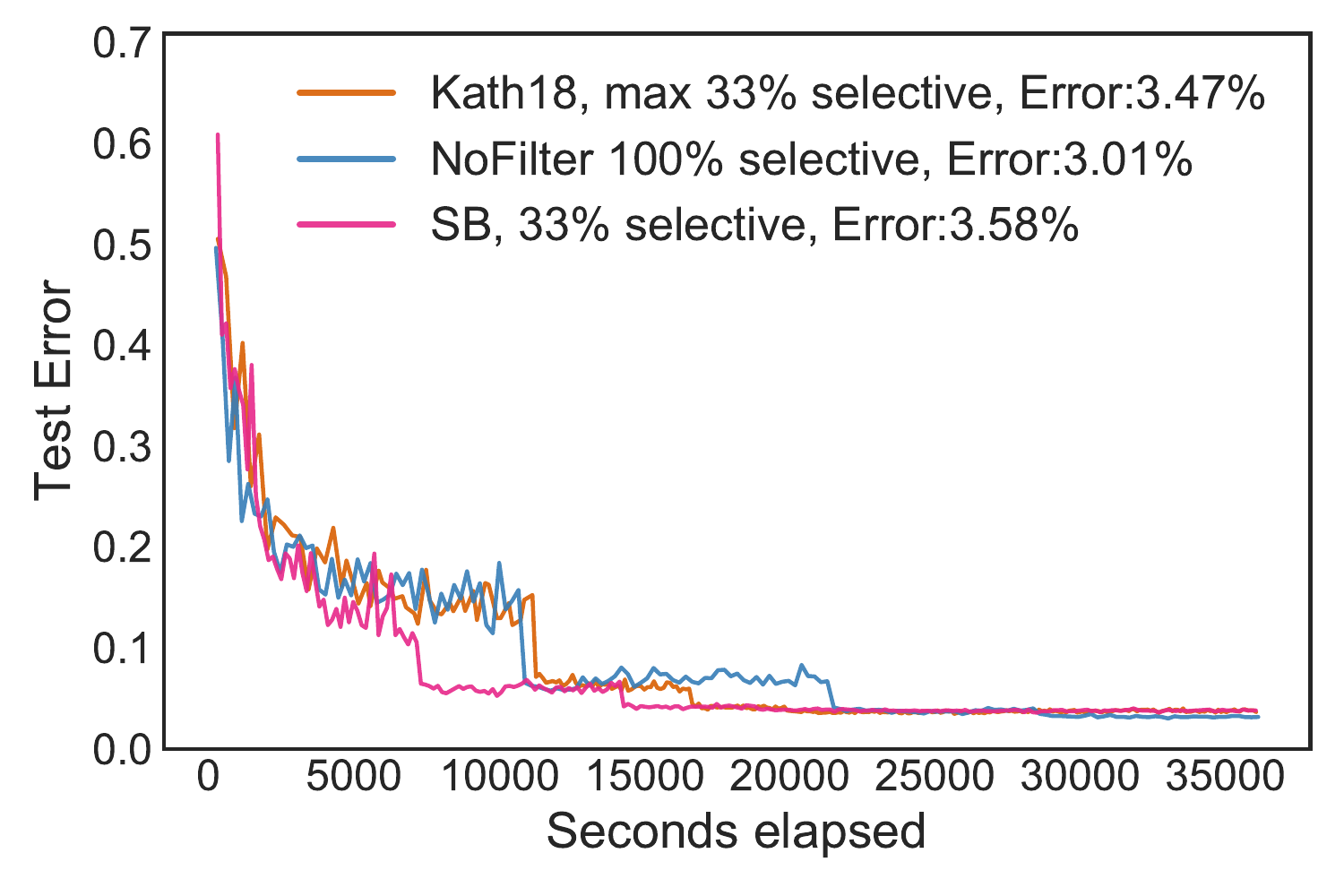}
          \caption{CIFAR10}
          \label{fig:strategy-seconds-cifar10-lr3}
        \end{subfigure}%
        \begin{subfigure}[b]{0.33\textwidth}
          \includegraphics[width=\linewidth]{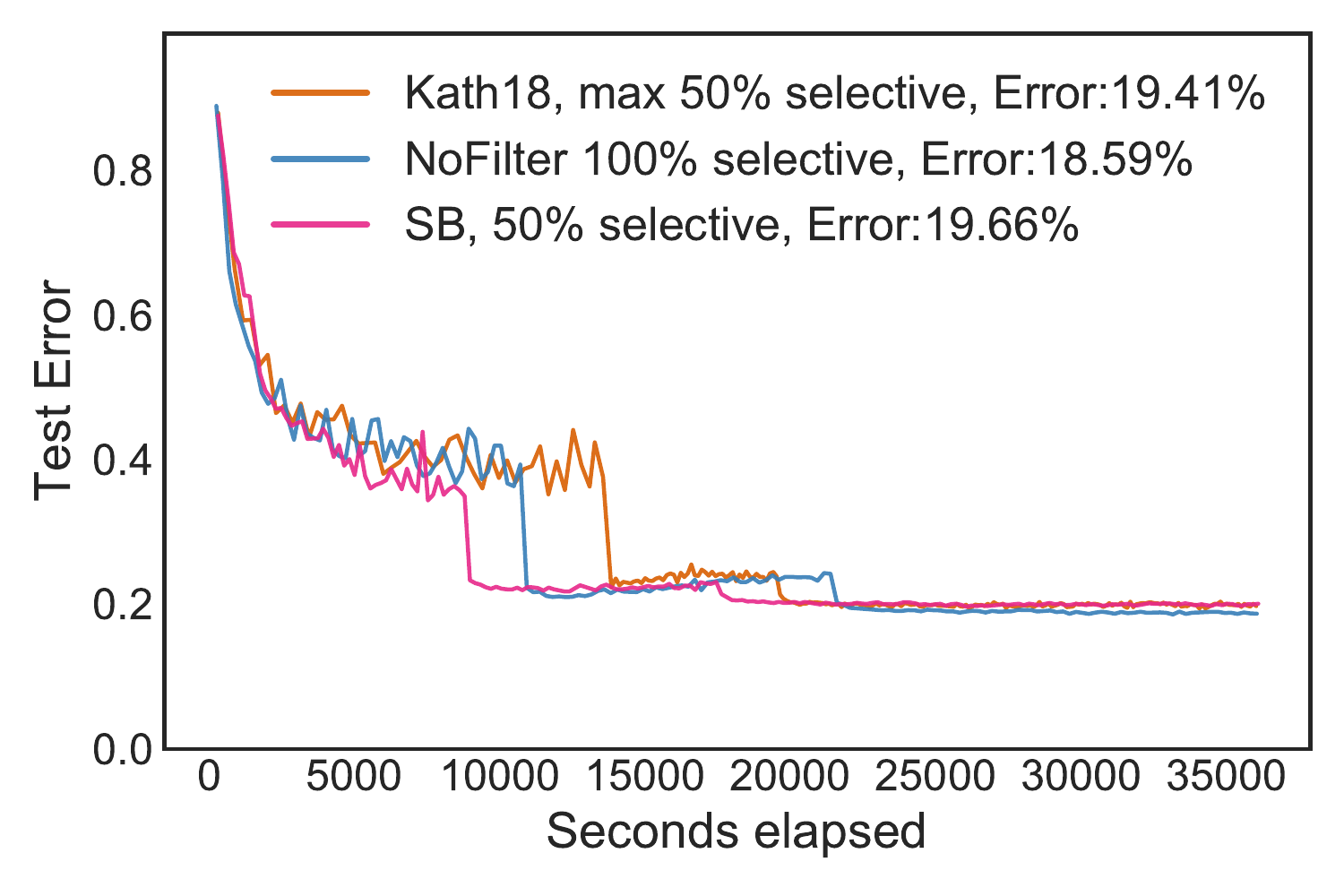}
          \caption{CIFAR100}
          \label{fig:strategy-seconds-cifar100-lr3}
        \end{subfigure}%
        \begin{subfigure}[b]{0.33\textwidth}
          \includegraphics[width=\linewidth]{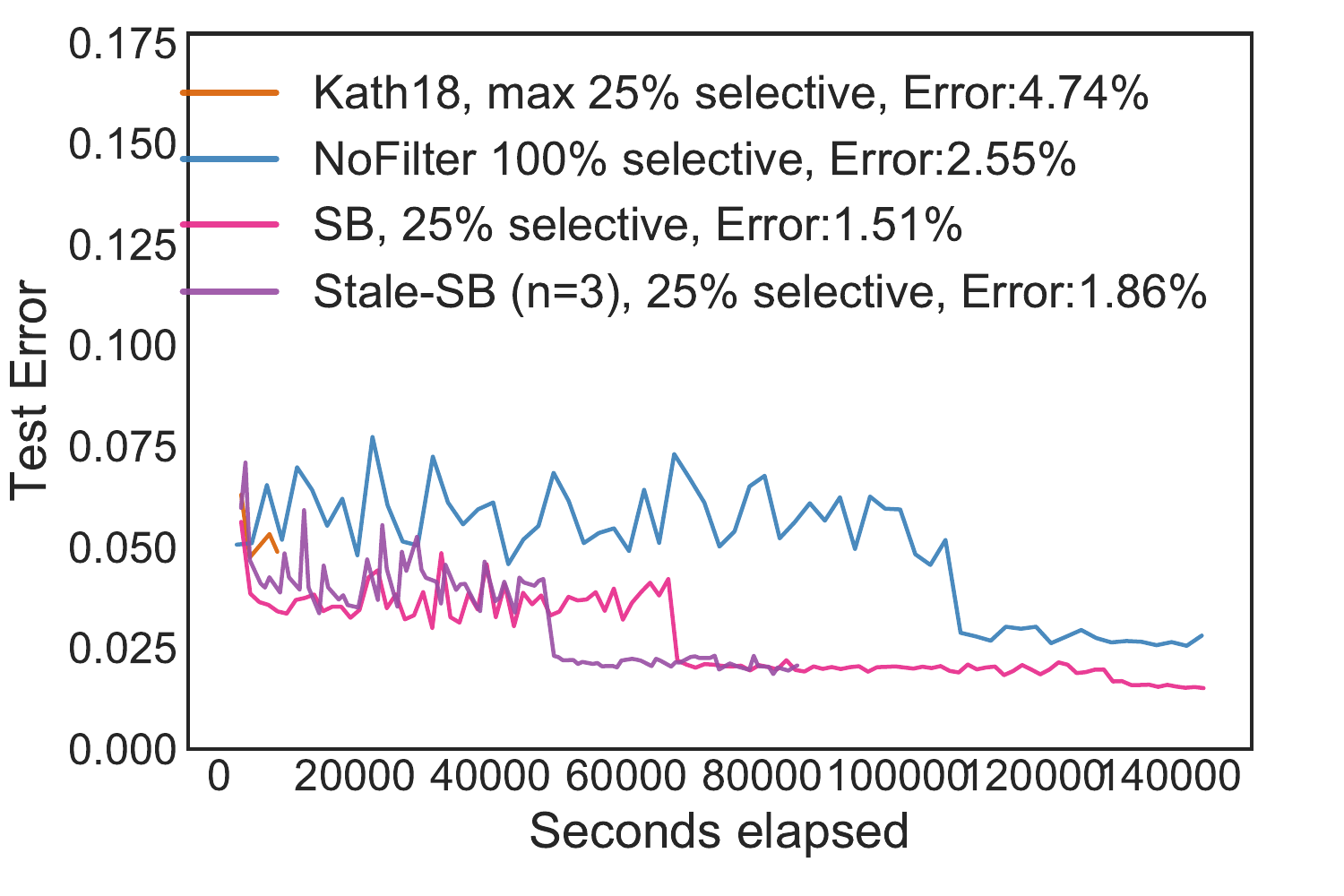}
          \caption{SVHN}
          \label{fig:strategy-seconds-svhn-lr3}
        \end{subfigure}%
        \caption{\SB{} reduces wall-clock time to target error with an accelerated learning rate schedule.}
        \label{fig:strategy-seconds-lr3}
\end{minipage}
% \squeeze{}
% \squeeze{}
\end{figure*}

\subsection{Asymmetry between cost of backward and forward pass}

\begin{figure*}[h]
\begin{minipage}{\textwidth}
        \centering
        \begin{subfigure}[b]{0.3\textwidth}
          \includegraphics[width=\linewidth]{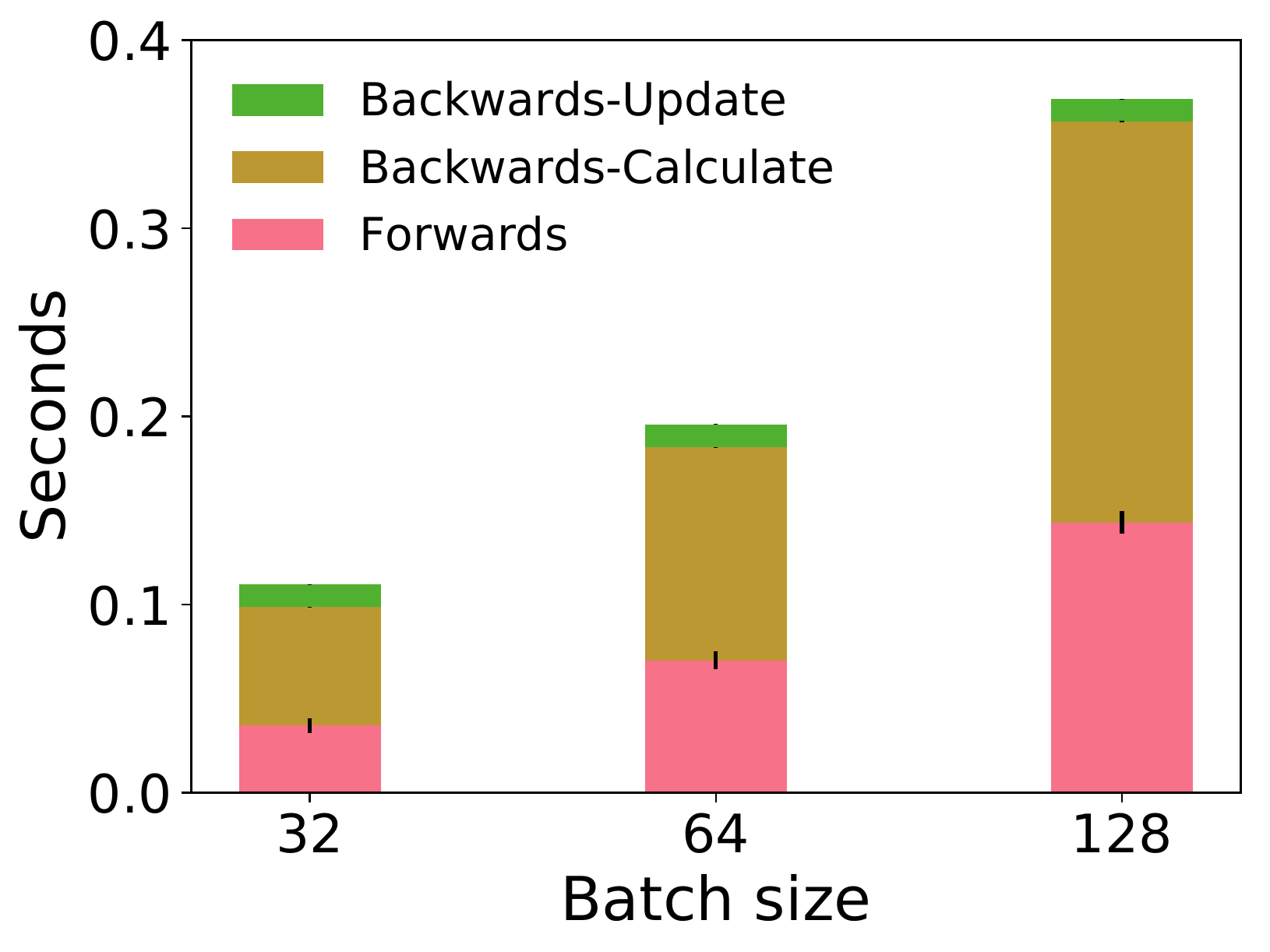}
          \caption{NVIDIA Tesla K20}
          \label{fig:asymmetry-k20}
        \end{subfigure}%
        \begin{subfigure}[b]{0.3\textwidth}
          \includegraphics[width=\linewidth]{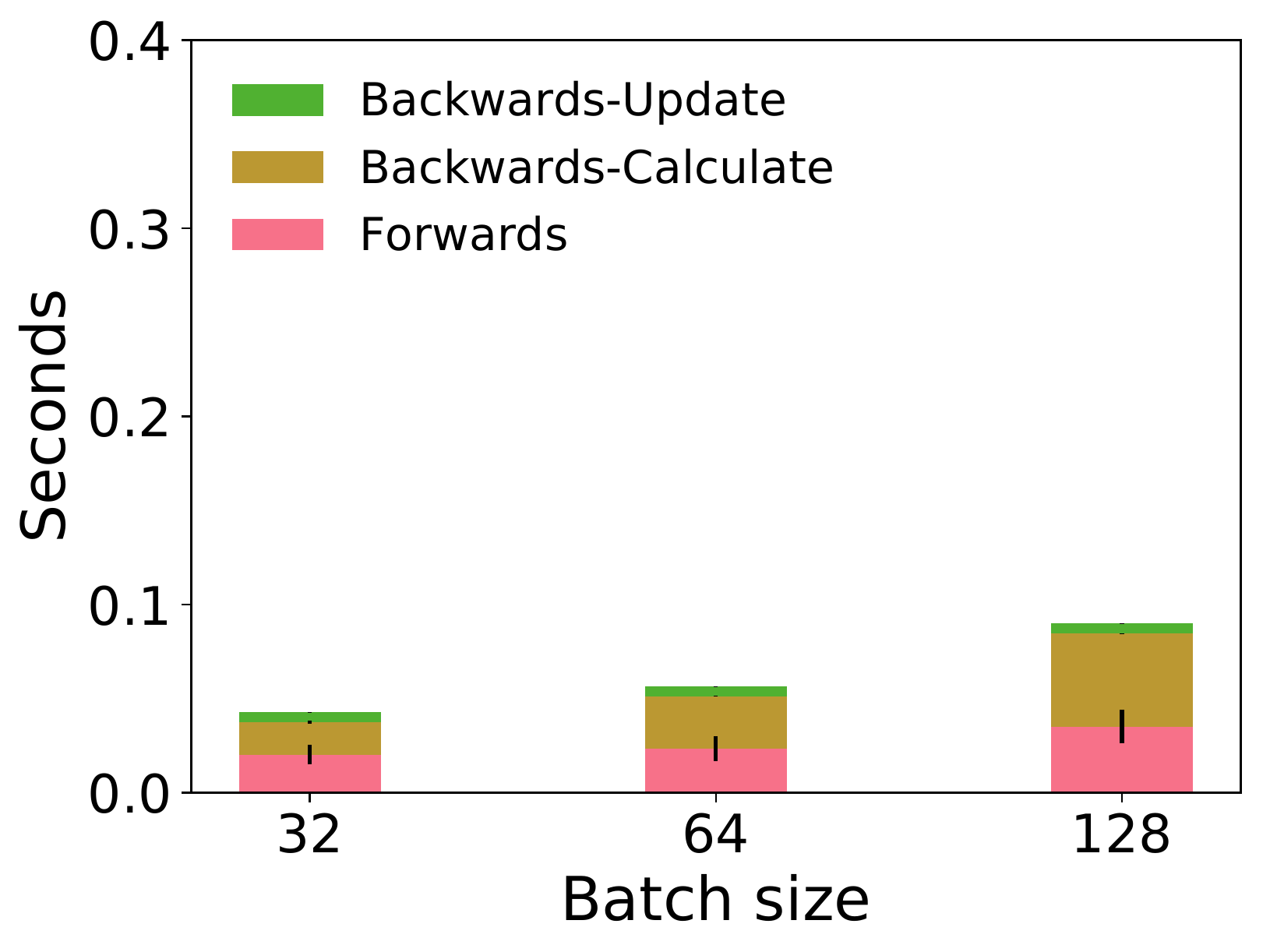}
          \caption{NVIDIA GeForce GTX-1070}
          \label{fig:asymmetry-gtx}
        \end{subfigure}%
        \begin{subfigure}[b]{0.3\textwidth}
          \includegraphics[width=\linewidth]{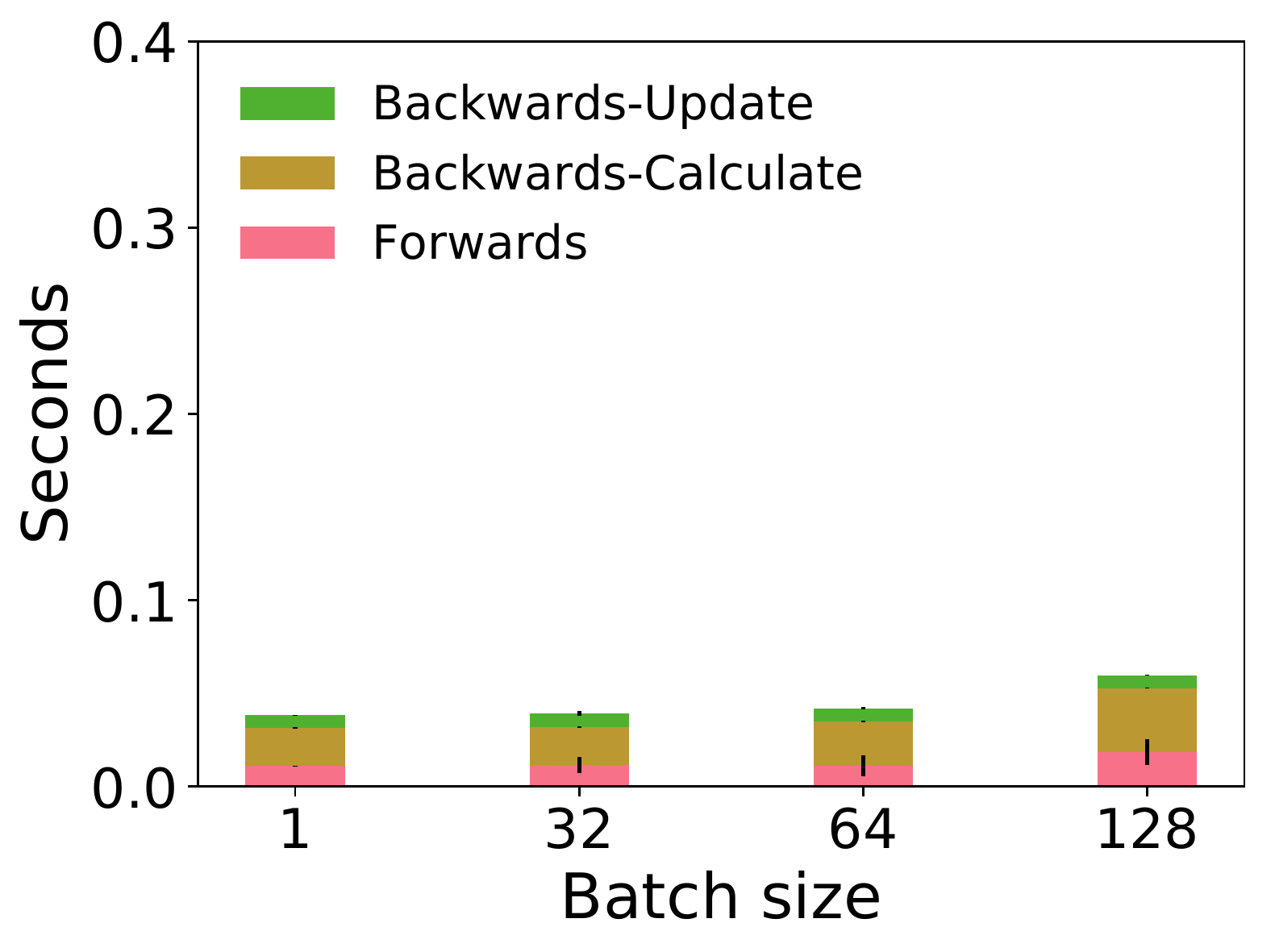}
          \caption{NVIDIA TitanV}
          \label{fig:asymmetry-titanv}
        \end{subfigure}%
        \caption{Breakdown of processing time per batch while training MobileNetV2 with \Scan{}.}
        \label{fig:asymmetry}
\end{minipage}
%\squeeze{}
\vspace{-10px}
\end{figure*}

In Figure~\ref{fig:asymmetry}, we confirm that on both a variety of modern
GPUs, the backward pass takes up to 2.5x as long as the forward pass. In
Section~\ref{sec:eval}, we run experiments on the K20 and the TitanV with a
batch size of 128.

\end{document}